\pdfoutput=1

\documentclass[11pt]{article}

\usepackage[final]{acl}

\usepackage{times}
\usepackage{latexsym}

\usepackage[T1]{fontenc}

\usepackage{changes}
\usepackage[utf8]{inputenc}

\usepackage{microtype}

\usepackage{inconsolata}

\usepackage{graphicx}

\usepackage{bm} 
\usepackage{bbm}
\usepackage{tcolorbox}
\usepackage{amsmath}
\usepackage{algorithm}
\usepackage{algorithmic}
\usepackage{setspace}
\usepackage{enumitem}

\usepackage{booktabs}
\usepackage{tikz}
\usepackage{xcolor}
\usepackage{subcaption} 
\usepackage{enumitem}
\usepackage{amssymb}

\newcommand*{\affmark}[1][*]{\textsuperscript{#1}}
\newcommand*{\email}[1]{\texttt{#1}}

\title{Spiral of Silence in Large Language Model Agents}

\author{
Mingze Zhong\affmark[1], 
Meng Fang\affmark[2], 
Zijing Shi\affmark[1], 
Yuxuan Huang\affmark[2], \\
\textbf{Shunfeng Zheng\affmark[1]}, 
\textbf{Yali Du\affmark[3]}, 
\textbf{Ling Chen\affmark[1]}, 
\textbf{Jun Wang\affmark[4]} \\
\affmark[1]AAII, University of Technology Sydney, NSW, Australia \\
\email{\{Mingze.Zhong, Zijing.Shi, Shunfeng.Zheng\}@student.uts.edu.au} \\
\email{Ling.Chen@uts.edu.au} \\
\affmark[2] University of Liverpool, Liverpool, UK \\
\email{\{Meng.Fang, Yuxuan.Huang\}@liverpool.ac.uk} \\
\affmark[3] King’s College London, London, UK \\
\email{yali.du@kcl.ac.uk} \\
\affmark[4] University College London, London, UK \\
\email{jun.wang@ucl.ac.uk}
}

\begin{document}

\maketitle

\begin{abstract}
%The Spiral of Silence (SoS) theory posits that individuals holding minority opinions often remain silent for fear of social isolation, enabling majority views to dominate public discourse. However, when agents are large language models (LLMs) instead of humans, this classical, emotion-driven explanation no longer holds, as these models neither possess emotions nor experience social anxiety. Thus, a fundamental question arises: Can purely statistical language generation mechanisms produce SoS dynamics in collectives of LLM agents? 
%This study propose an evaluation framework comprising four controlled experimental conditions that systematically vary the presence of ``Persona'' and ``History''  signals. Opinion dynamics are quantified using trend analysis through the Mann–Kendall test and Spearman’s rank correlation, combined with concentration metrics such as kurtosis and interquartile range. Experiments on three open-source and one closed-source model reveal that: (i) both signals together produce strong majority dominance, replicating SoS dynamics; (ii) persona signals alone foster diversity but no consensus, while history signals alone create strong anchoring; and (iii) without social signals, most models exhibit a default bias. 
%These findings challenge emotion-based accounts of SoS and highlight the need to understand and mitigate opinion convergence in LLM-agent systems, linking computational sociology with responsible AI design.
The Spiral of Silence (SoS) theory holds that individuals with minority views often refrain from speaking out for fear of social isolation, enabling majority positions to dominate public discourse. When the ``agents'' are large language models (LLMs), however, the classical psychological explanation is not directly applicable, since SoS was developed for human societies. This raises a central question: can SoS-like dynamics nevertheless emerge from purely statistical language generation in LLM collectives? We propose an evaluation framework for examining SoS in LLM agents. Specifically, we consider four controlled conditions that systematically vary the availability of ``History'' and ``Persona'' signals. Opinion dynamics are assessed using trend tests such as Mann–Kendall and Spearman’s rank, along with concentration measures including kurtosis and interquartile range. Experiments across open-source and closed-source models show that history and persona together produce strong majority dominance and replicate SoS patterns; history signals alone induce strong anchoring; and persona signals alone foster diverse but uncorrelated opinions, indicating that without historical anchoring, SoS dynamics cannot emerge. The work bridges computational sociology and responsible AI design, highlighting the need to monitor and mitigate emergent conformity in LLM-agent systems.

\end{abstract}

% !TEX root = acl_latex.tex
\section{Introduction}

In human societies, public opinion is shaped by complex social dynamics. The \textit{Spiral of Silence} (SoS) theory ~\cite{noelle1974spiral} posits that individuals tend to withhold opinions they perceive as unpopular due to fear of social isolation. This creates a self-reinforcing cycle in which minority views gradually disappear from public discourse, making dominant opinions appear increasingly widespread.

%Insights from social psychology provide a valuable foundation for understanding how emergent behaviors arise in populations of artificial agents, particularly those built on large language models (LLMs)~\cite{ai2023gpt,touvron2023llama}. With their increasing autonomy and  generalization capabilities, LLM-based agents are now widely deployed in open-ended multi-agent environments, where they collaborate or compete to accomplish diverse tasks.
%Although these models do not possess human emotions or social drives, recent studies have shown that they can adapt their behavior in response to social-like structures in their input, such as displaying sycophantic tendencies toward users \cite{sharma2023towards} or reproducing group-level biases in multi-agent interactions \cite{ashery2024dynamics}. 
%This raises a central research question: can populations of LLM agents spontaneously exhibit self-reinforcing opinion dynamics such as SoS? Investigating this possibility is essential for understanding the collective behaviors that may emerge in LLM-driven systems and for ensuring their safe, predictable deployment in real-world multi-agent settings.
A pressing question is whether analogous dynamics can arise in populations of artificial agents. Large language models (LLMs) are rapidly deployed in multi-agent environments where they collaborate, negotiate, or compete~\cite{ai2023gpt,touvron2023llama}. Unlike humans, LLMs do not experience emotions or social anxieties, yet recent studies show they adapt to social-like cues—displaying sycophantic tendencies toward users~\cite{sharma2023towards} and reproducing group-level biases~\cite{ashery2024dynamics}. If SoS-like effects can emerge in such settings, this would challenge traditional emotion-based explanations and raise critical concerns about bias amplification, conformity, and opinion manipulation in LLM-driven systems.

To investigate this possibility, we design an evaluation framework for detecting SoS dynamics in LLM agents. We adopt a controlled movie-rating task, which provides a quantifiable environment for measuring opinion formation. Two key signals are introduced: (i) \textit{History}, the average rating of preceding agents, serving as a dynamic proxy for the collective opinion climate;  (ii) \textit{Persona}, a role assigned to each agent that encodes predispositions and ensures diversity of initial preferences.
Crossing these signals allows us to disentangle the influence of collective anchoring and individual predisposition, and to test whether their interaction gives rise to SoS-like convergence.

We hypothesize that SoS dynamics will be most evident when they drive agents to align with perceived majorities. To measure this, we combine concentration statistics such as interquartile range~\cite{clark2005interquartile} and kurtosis~\cite{balanda1988kurtosis} with trend diagnostics including the Mann–Kendall test~\cite{mann1945nonparametric} and Spearman’s rank correlation~\cite{spearman1904proof}.
We evaluate a range of LLM families, including the open-source \texttt{Qwen}~\cite{bai2023qwen}, \texttt{DeepSeek}~\cite{liu2024deepseek}, and \texttt{Mistral}, as well as the closed-source \texttt{GPT-4o-mini}~\cite{hurst2024gpt}, enabling both cross-family comparisons and within-family scaling analyses. Our findings show that in the absence of social signals, agents default to positive movie ratings; persona signals promote opinion heterogeneity; history signals exert strong anchoring effects; and SoS dynamics emerge most clearly when both signals are present.

%Based on this set-up, we hypothesize that if an SoS effect is present among LLM agents, they will increasingly align with the perceived majority when both historical ratings and varying initial predispositions are present. To quantify these dynamics, we employ two categories of metrics: concentration statistics such as the Interquartile Range ~\cite{clark2005interquartile} and Kurtosis ~\cite{balanda1988kurtosis}, and trend diagnostics, including the Mann-Kendall test ($S$) ~\cite{mann1945nonparametric} and Spearman's $\rho$ \cite{spearman1904proof}. We evaluated open-source LLM families: \texttt{Qwen}  ~\cite{bai2023qwen} series, \texttt{DeepSeek} ~\cite{liu2024deepseek}, \texttt{Mistral} and \texttt{Llama-3} ~\cite{grattafiori2024llama} and closed-source \texttt{GPT-4o-mini} ~\cite{hurst2024gpt}, facilitating both cross-family comparisons and within-family scaling analyses. The results show that in the absence of social signals, LLM agents default to positive movie ratings. Introducing a persona encourages opinion heterogeneity, while historical average ratings exert an anchoring influence. The SoS effect is significantly more likely to emerge when both signals are provided.

Our contributions are as follows:
\begin{itemize}[noitemsep, topsep=0pt, leftmargin=*]
\item We propose a systematic framework for testing SoS in LLM agents, isolating the roles of collective influence through History and individual predisposition through Persona.\footnote{Data and code available at: \url{https://github.com/aialt/SoS-LLMs}}

\item We provide the first evidence that SoS-like dynamics can arise from learned language generation mechanisms, yielding new insights into the nature of conformity.

\item We identify behavioral regularities such as positivity bias and anchoring effects, and discuss their implications for the design and governance of multi-agent AI systems.
\end{itemize}
%We introduce an innovative experimental framework to empirically test and quantify the Spiral of Silence in LLM agents, successfully isolating the effects of individual predisposition (Persona) and collective influence (History).

%$\bullet$ We provide the first evidence that SoS-like dynamics can emerge from statistical generation mechanisms, offering a new perspective on conformity.
%We provide the first empirical evidence that SoS dynamics can emerge from purely statistical mechanisms without emotional drivers, offering a new perspective on traditional social psychological explanations.

%$\bullet$ We identify key behavioral regularities such as positivity bias and anchoring effects, and highlight their implications for the governance of multi-agent AI systems.
%Through extensive experiments on a variety of mainstream models, we systematically analyze the conditions for the convergence of opinion, identifying key behavioral patterns such as ``positivity bias'' and ``anchoring effects''.

%$\bullet$ We highlight the critical implications of these findings for the design and governance of multi-agent AI systems, particularly concerning risks like bias amplification and opinion manipulation.

%\input{preliminaries}

% !TEX root = acl_latex.tex
\section{Methodology}

This section describes the task simulation, presents the evaluation framework, and specifies the metrics and criteria for measuring opinion dynamics and detecting SoS effects in LLM agents.

%\subsection{Task Simulation}
\subsection{Problem Setup}
%\zj{
We construct a multi-agent environment to simulate an online rating system. In this task, a population of $N$ agent sequentially rate the same movie on an integer scale. We formalize the rating space as an $M$-level cardinal metric:
$\mathcal{M} \triangleq \{1, \ldots, M\}$.
At step $k$, the rating given by agent $i$ to movie $j$ is denoted by $r_{j,k} \in \mathcal{M}$. The set of historical ratings for movie $j$ up to the $k$-th rating is defined as $\mathcal{H}_{j,k} \triangleq \{ r_{j,1}, \ldots, r_{j,k} \}$.
%}

%\zj{
As agents rate sequentially, each observes prior ratings giving its own. The ``collective opinion climate'', representing an agent’s perception of public opinion, is defined at step $k+1$ as the average of all preceding ratings available to the $(k+1)$-th agent:
$\mathcal{F}(\mathcal{H}_{j,k}) = \frac{1}{k} \sum_{l=1}^{k} r_{j,l}$.

\subsection{LLM Agent Design}

We design LLM agents by varying the presence of \textit{History} and \textit{Persona} signals to examine their effects on the emergence of SoS.  

\begin{itemize}[noitemsep, topsep=0pt, leftmargin=*]
\item \textbf{History:} We operationalize the collective opinion climate as the average rating of all preceding agents, provided as input to the next agent. This signal is endogenous and dynamic: each new rating updates the climate and is passed forward, creating a feedback loop. Such recursive updating allows a slight majority to amplify its influence under the SoS effect, distinguishing this process from a static anchoring effect.  

\item \textbf{Persona:} To introduce heterogeneous predispositions, each agent is assigned a unique persona. Personas are specified through rich textual descriptions covering attributes such as occupation, interests, and background.
\end{itemize}

This $2 \times 2$ design yields four controlled scenarios:  

\begin{itemize}[noitemsep, topsep=0pt, leftmargin=*]
\item \textbf{History + Persona}: The agent is assigned a persona and observes the historical average rating of preceding agents. This condition captures how an identity-driven agent behaves under the influence of a collective opinion climate.  

\item \textbf{History only}: The agent observes only the historical average rating, without a persona. This isolates the effect of perceived public opinion on a generic agent.  

\item \textbf{Persona only}: The agent receives only a persona description, providing a fixed identity and predispositions but no historical signal. This condition examines how internal preferences, shaped by persona, influence rating behavior.  

\item \textbf{No History, No Persona}: The agent receives only rating instructions and movie information, without persona context or historical ratings. This baseline captures a generic agent’s behavior in the absence of external signals.  
\end{itemize}

%The corresponding prompt templates are provided in Appendix~\ref{sec:prompt}.
The corresponding prompt templates are provided in Appendix~\ref{sec:prompt}, and an example persona is given in Appendix~\ref{sec:persona}.

\subsection{Quantifying SoS Dynamics}

To evaluate whether LLM agents exhibit Spiral of Silence behavior, we focus on two defining patterns: (i) a dynamic self-reinforcing process whereby the majority opinion becomes increasingly dominant over time, and (ii) a final outcome of high consensus, where minority opinions have effectively faded. To capture these aspects, we employ opinion trend metrics and rating concentration metrics.

% Our goal is to empirically detect the Spiral of Silence, a phenomenon characterized by two distinct yet interconnected features: (1) a dynamic self-reinforcing process whereby the majority opinion becomes increasingly dominant over time, and (2) a final outcome of high consensus, where minority opinions have effectively faded. To capture both aspects, we employ two complementary categories of quantitative metrics. For each movie, we obtain a sequence of $T$=80  integer ratings for this analysis.

\subsubsection{Opinion Trend Metrics}

%\zj{
We introduce the Majority-Conforming Opinion (MCO) sequence, which captures the dominant opinion trend by aggregating cumulative proportions of positive and negative ratings. Formally, for movie $j$ at $k$-th rating, we compute the cumulative proportions of positive and negative ratings as $\text{pos}_{j,k} = \frac{1}{k}\sum_{t\le k}\mathbf{1}\!\bigl[r_{i,j,t}\ge 6\bigr]$ and $\text{neg}_{j,k} = \frac{1}{k}\sum_{t\le k}\mathbf{1}\!\bigl[r_{i,j,t}\le 5\bigr]$, where $\mathbf{1}[\cdot]$ is the indicator function. The MCO sequence is then defined as $\text{MCO}_{j,k} = \max\bigl\{\text{pos}_{j,k},\,\text{neg}_{j,k}\bigr\}$. On this basis, two metrics are proposed to quantify opinion trends.
%}
% To capture the dynamic reinforcement process of the majority opinion, we first define the cumulative proportions of positive and negative ratings up to round $k$, $\text{pos}_{j,k} = \frac{1}{k}\sum_{t\le k}\mathbf{1}\!\bigl[r_{i,j,t}\ge 6\bigr]$ and $\text{neg}_{j,k} = \frac{1}{k}\sum_{t\le k}\mathbf{1}\!\bigl[r_{i,j,t}\le 5\bigr]$, where the summation iterates over each round of ratings $t$ from 1 to $k$, where $\mathbf{1}[\cdot]$ is the indicator. The dynamic ``Majority-Conforming Opinion'' (MCO) sequence is then defined as $\text{MCO}_{j,k} = \max\bigl\{\text{pos}_{j,k},\,\text{neg}_{j,k}\bigr\}$. \noindent To reduce the impact of early stage fluctuations, all trend statistics are computed starting from round m=11.
\begin{itemize}[noitemsep, topsep=0pt, leftmargin=*]
\item \textbf{Mann--Kendall statistic ($S$)},
%\zj{
which is well-suited for detecting monotonic trends in time series without assuming linearity. We apply the Mann--Kendall trend test to the MCO sequence. The $S$ statistic is defined as
$S_{j}=\sum_{k=m}^{T-1}\sum_{t=k+1}^{T}\!\operatorname{sgn} \bigl(\text{MCO}_{j,t}-\text{MCO}_{j,k}\bigr)$,
where $m$ denotes the starting round for analysis and $T$ is the final round of ratings. A significantly positive $S$ shows that the majority opinion strengthens monotonically over time. This provides direct evidence for the self-reinforcing dynamic that characterizes the SoS.
%}

\item \textbf{Spearman's Rank Correlation ($\rho$)}, 
%\zj{
which is used to quantify the strength of a monotonic trend. 
To complement this, we compute Spearman's rank correlation between the MCO sequence and the time steps. 
It is calculated as
$\rho_j = 1 - \frac{6 \sum d_k^2}{ n (n^2 - 1) }$, where $d_k$ is the difference between the rank of the $k$-th time step and the rank of its corresponding $\text{MCO}_{j,k}$ value, and $n$ is the number of observations (i.e., $n=T-m+1$).
A value of $\rho$ close to 1 indicates that the majority opinion exhibits a strong monotonic increase.
%}
% \textbf{ ($S$): } Used to detect a monotonic trend in the MCO sequence. In our work, we use it to verify a key component of SoS: whether the majority opinion systematically strengthens over time. A significantly positive value $S$ indicates a monotonic increase in the majority support, providing direct evidence of the ``spiral'' effect. Its statistic is calculated as 
% $
% S_{j}=\sum_{k=m}^{T-1}\sum_{t=k+1}^{T}\!\operatorname{sgn} \bigl(\text{MCO}_{j,t}-\text{MCO}_{j,k}\bigr).
% $

% \noindent $\bullet$ \textbf{Spearman's Rank Correlation ($\rho$): } Quantifies the rank-based correlation between the MCO sequence and the time steps. It specifically measures the strength of the monotonic relationship between the rank of the time steps ($k=m,\cdots,T$) and the rank of the MCO sequence values. We use it to assess the strength and consistency of the majority opinion's growth trend. A value of $\rho$ close to 1 implies a nearly perfect reinforcement of the majority opinion. It is calculated as 
% $
% \rho_j = 1 - \frac{6 \sum d_k^2}{ n (n^2 - 1) },
% $
% \noindent where $d_k$ is the difference between the rank of the $k$-th time step and the rank of its corresponding $\text{MCO}_{j,k}$ value, and $n$ is the number of observations (i.e., $T-m+1$).

\end{itemize}

\subsubsection{Rating Concentration Metrics}
%\zj{
To assess whether the ratings of LLM agents converge toward a strong consensus, we evaluate the dispersion of the final set of $L$ ratings using the following metrics:
%}

\begin{itemize}[noitemsep, topsep=0pt, leftmargin=*]
\item \textbf{Kurtosis}, which measures the ``peakedness'' of the rating distribution and indicates whether the ratings are sharply concentrated around a single value. We compute it over the final $L$ ratings as 
$
\text{Kurt}_{L}(j)=\frac{1}{L}\sum^{T}_{k=T-L+1}
    \left(\frac{r_{i,j,k}-\mu_j}{\sigma_j}\right)^{4}-3,
$
where $\mu_j$ and $\sigma_j$ are the mean and standard deviation of the same $L$ ratings. 
A positive Kurtosis value indicates a distribution more sharply peaked than the normal distribution, showing that ratings are concentrated around a single value. This pattern aligns with the consensus formation expected under the SoS.

\item  \textbf{Interquartile Range (IQR)}, which provide a robust measure of the central spread of recent ratings. We use it to assess whether late-stage ratings are tightly clustered, with a smaller IQR indicating stronger opinion concentration. It is calculated as 
$
IQR_L(j) = Q^{(L)}_3(j) - Q^{(L)}_1(j),
$
where $Q^{(L)}_1(j)$ and $Q^{(L)}_3(j)$ denote the 25th and 75th percentiles of the last $L$ ratings.

\end{itemize}

% To quantify the final outcome of the spiral process (i.e., a strong consensus), we measure the dispersion of the final $L$=30 ratings using the following metrics:
% \noindent $\bullet$ \textbf{Kurtosis: } Describes the ``peakedness'' of the rating distribution. We use it to detect if the distribution is sharply peaked around a single value. Positive kurtosis indicates a more peaked distribution than normal, which means ratings are highly concentrated, which is consistent with the unification of opinions resulting from SoS. It is calculated as :
% $
% \text{Kurt}_{L}(j)=\frac{1}{L}\sum^{T}_{k=T-L+1}
%     \left(\frac{r_{i,j,k}-\mu_j}{\sigma_j}\right)^{4}-3,
% $
% where $\mu_j$ and $\sigma_j$ are the mean and standard deviation of the same $L$ ratings.

% \noindent $\bullet$ \textbf{Interquartile Range (IQR): }  A robust measure of the central spread of recent ratings. We use it to assess whether late-stage ratings are tightly clustered; a smaller IQR indicates high opinion concentration, consistent with the state after minority opinions are silenced. It is calculated as 
% $
% IQR_L(j) = Q_3 -Q_1,
% $
% where $Q_1$ and $Q_3$ are the 25th and 75th percentiles (the first and third quartiles) of the final $L$ ratings, respectively.

\begin{figure*}[t]
    \centering
    \begin{subfigure}[t]{0.24\textwidth}
        \centering
        \includegraphics[width=\linewidth]{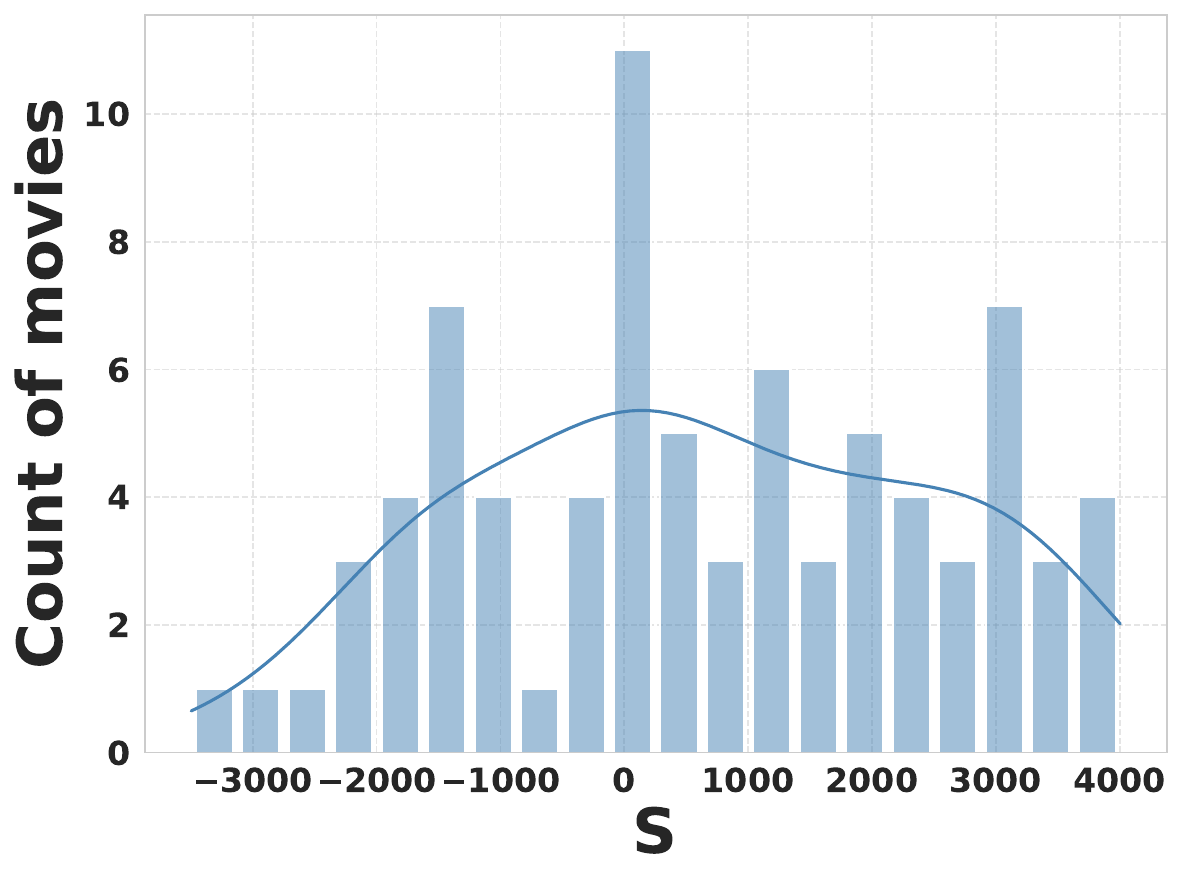}
        \caption{\small Mann–Kendall Distribution}
        \label{fig:Distributions of GPT-4o-mini wpwh a}
    \end{subfigure}
    \hfill 
    \begin{subfigure}[t]{0.24\textwidth}
        \centering
        \includegraphics[width=\linewidth]{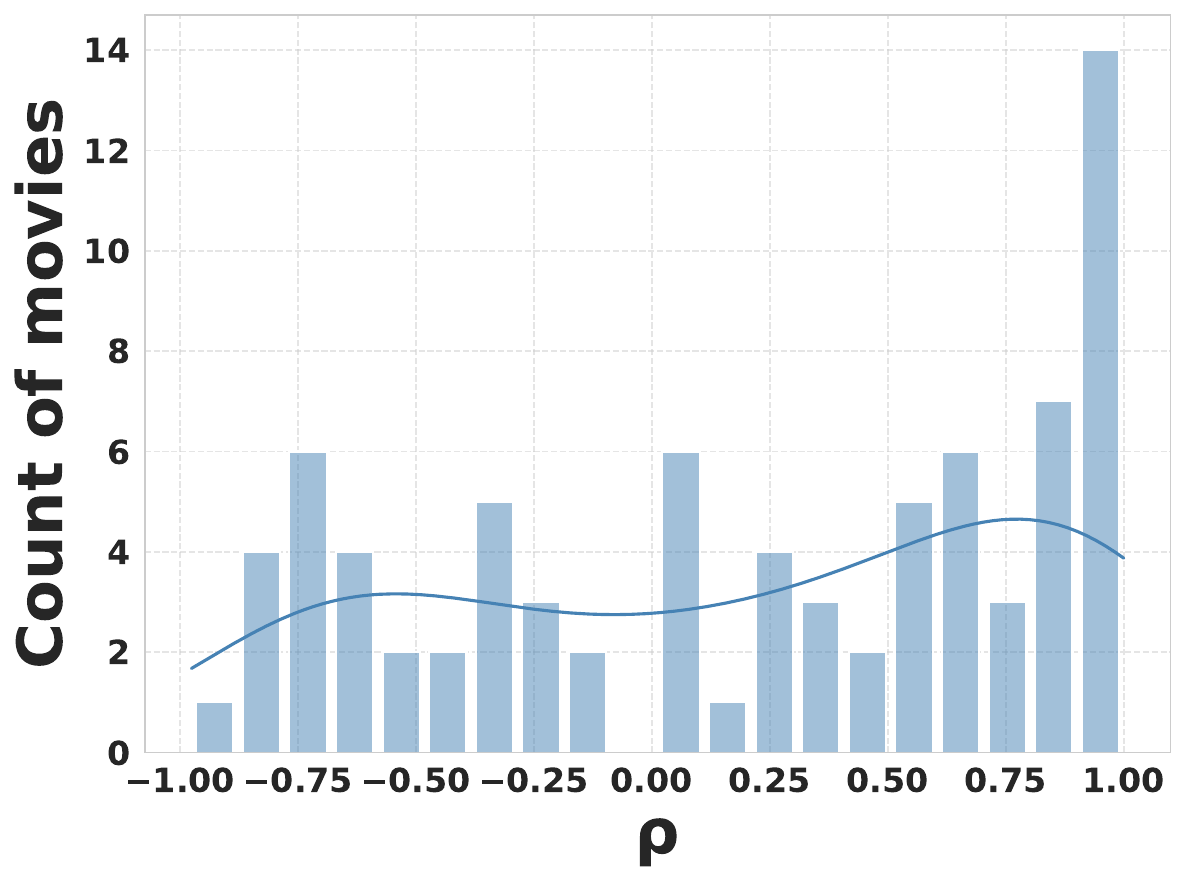}
        \caption{\small Spearman Distribution}
        \label{fig:Distributions of GPT-4o-mini wpwh b}
    \end{subfigure}
    \hfill 
    \begin{subfigure}[t]{0.24\textwidth}
        \centering
        \includegraphics[width=\linewidth]{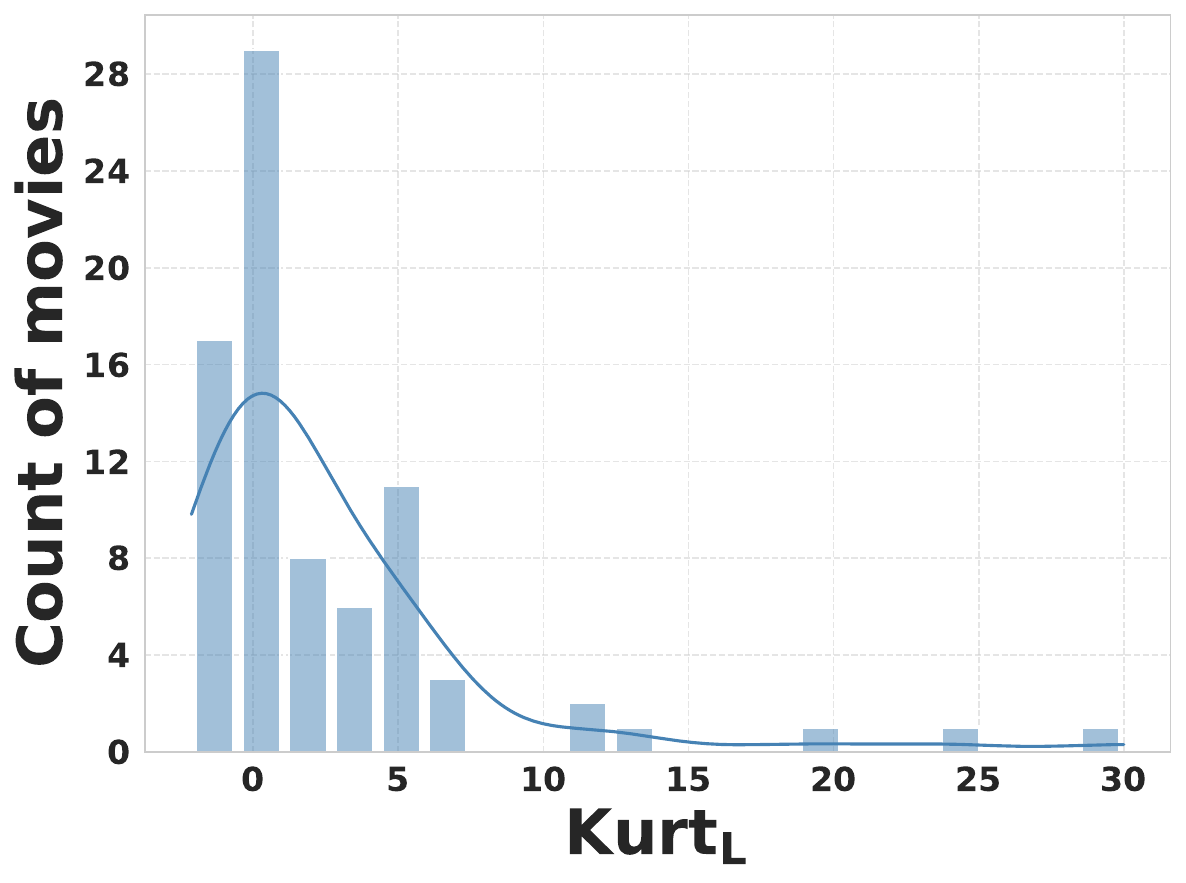}
        \caption{\small Kurtosis Distribution}
        \label{fig:Distributions of GPT-4o-mini wpwh c}
    \end{subfigure}
    \hfill 
    \begin{subfigure}[t]{0.24\textwidth}
        \centering
        \includegraphics[width=\linewidth]{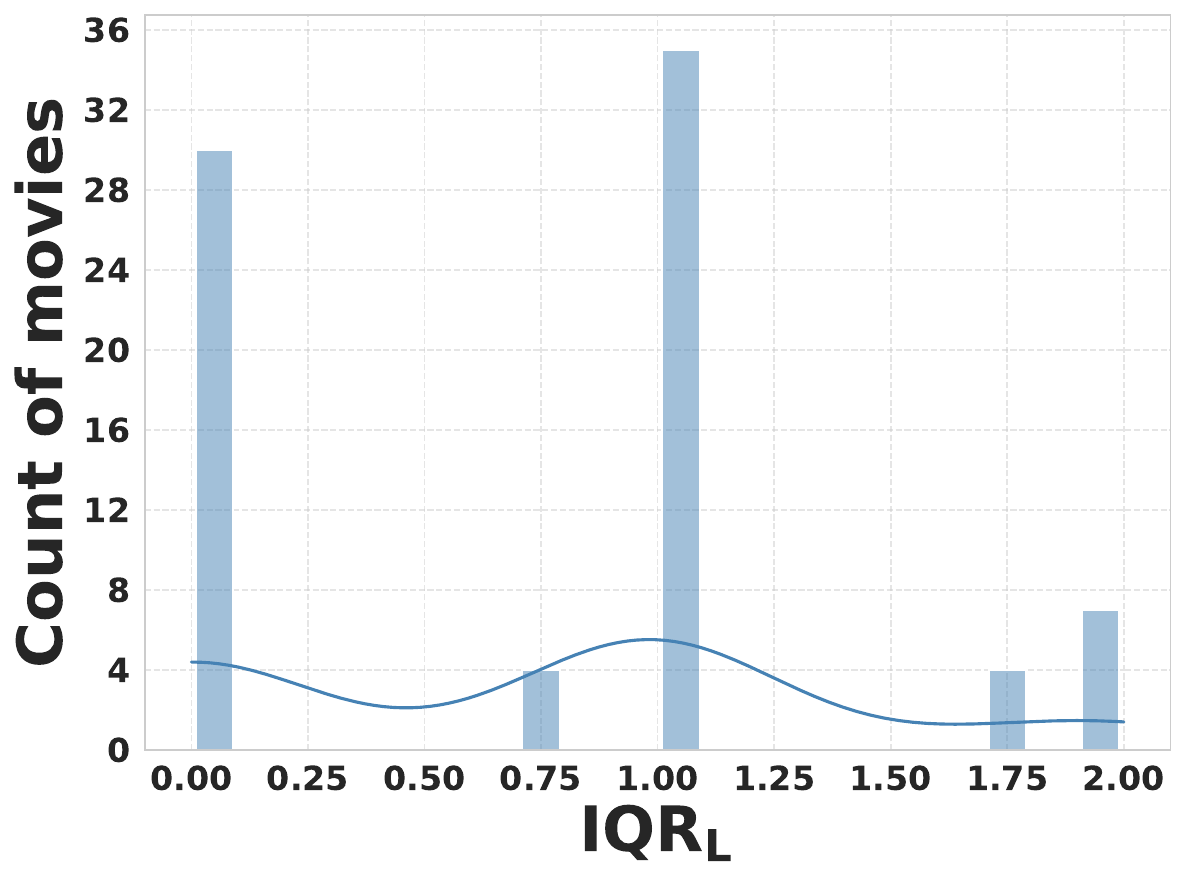}
        \caption{\small IQR Distribution}
        \label{fig:Distributions of GPT-4o-mini wpwh d}
    \end{subfigure}
    %\caption{Distributions of metrics for all movie rating sequences on GPT-4o-mini under history + persona scenario (w/ History \& w/ Persona).}
    \caption{Results of SoS metrics for GPT-4o-mini in Scenario I (History + Persona).}
    \label{fig:Distributions of GPT-4o-mini wpwh}
\end{figure*}

\begin{figure*}[t]
    \centering
    \begin{subfigure}[t]{0.24\textwidth}
        \centering
        \includegraphics[width=\linewidth]{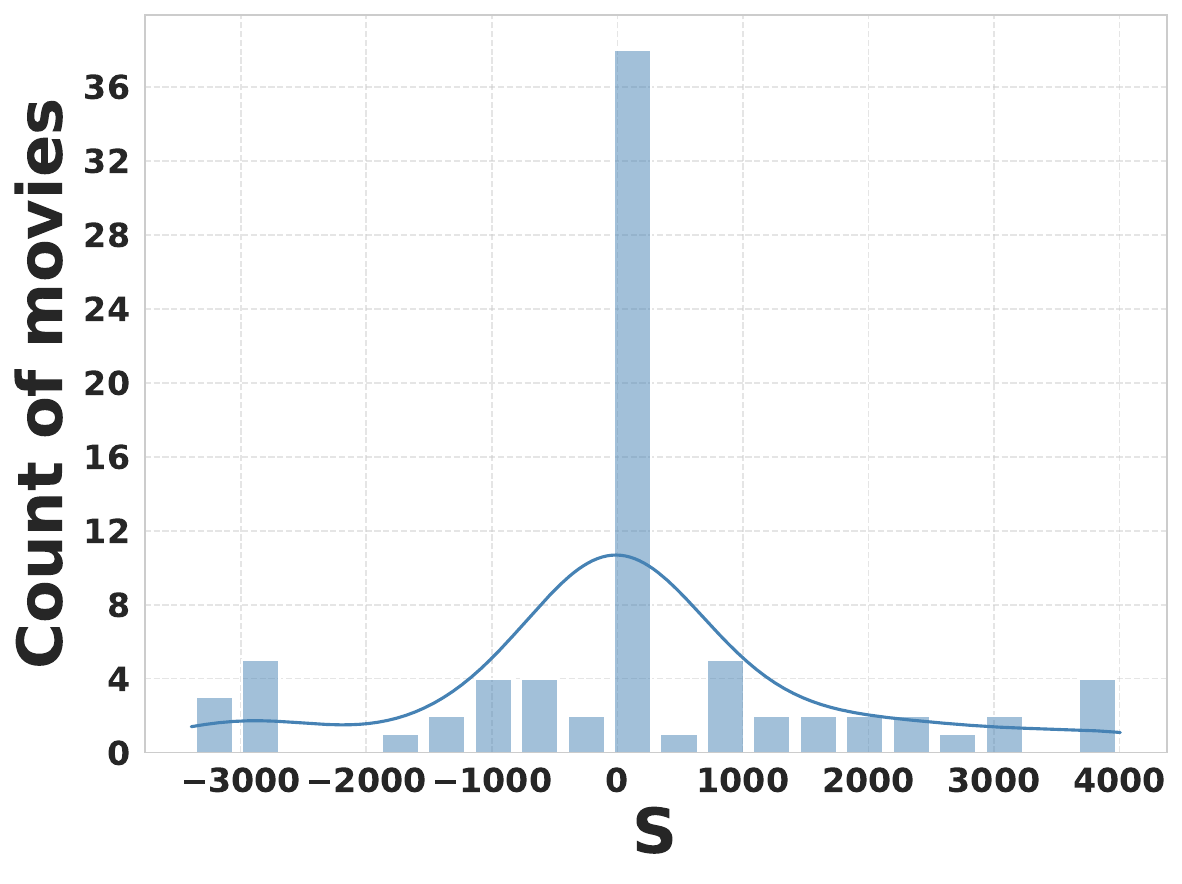}
        \caption{Mann–Kendall Distribution}
        \label{fig:Distributions of Ministral-8B-Instruct-2410 wpwh a}
    \end{subfigure}
    \hfill 
    \begin{subfigure}[t]{0.24\textwidth}
        \centering
        \includegraphics[width=\linewidth]{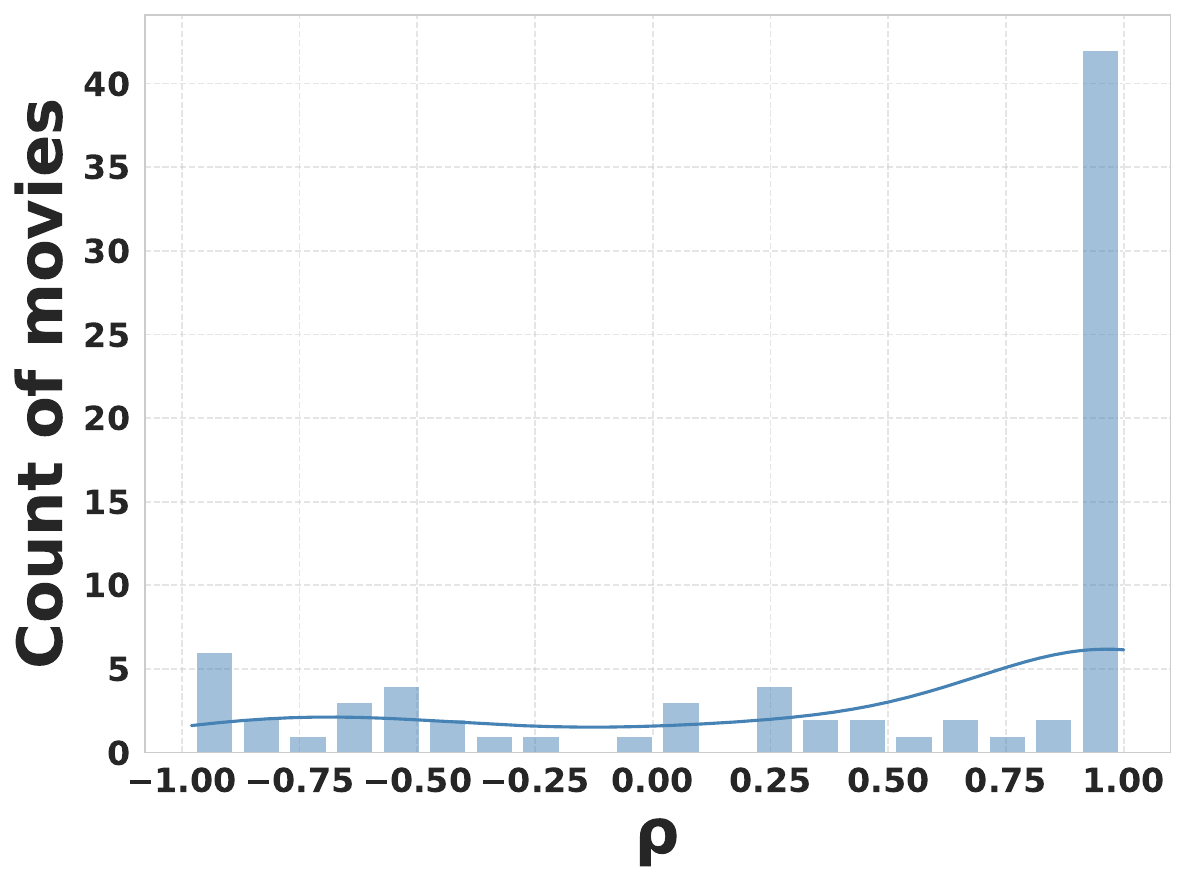}
        \caption{Spearman Distribution}
        \label{fig:Distributions of Ministral-8B-Instruct-2410 wpwh b}
    \end{subfigure}
    \hfill 
    \begin{subfigure}[t]{0.24\textwidth}
        \centering
        \includegraphics[width=\linewidth]{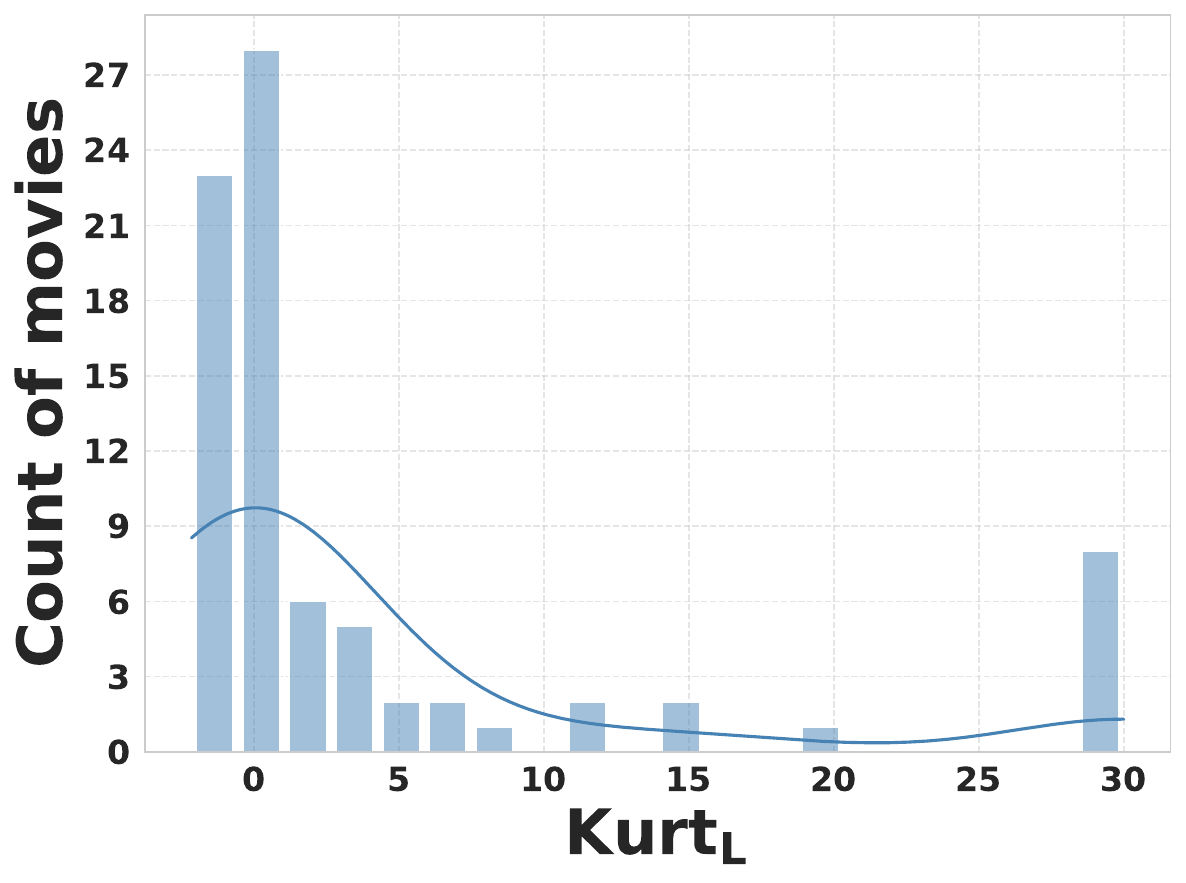}
        \caption{Kurtosis Distribution}
        \label{fig:Distributions of Ministral-8B-Instruct-2410 wpwh c}
    \end{subfigure}
    \hfill 
    \begin{subfigure}[t]{0.24\textwidth}
        \centering
        \includegraphics[width=\linewidth]{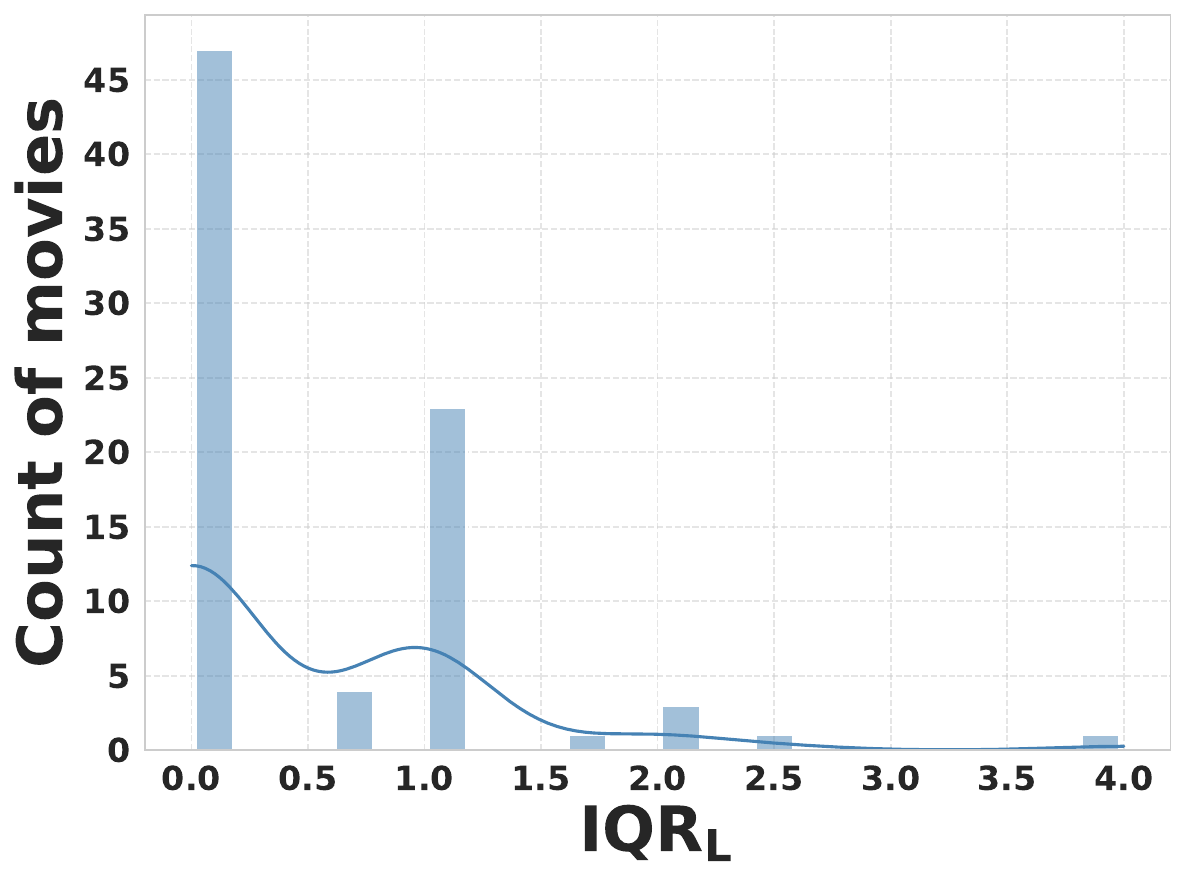}
        \caption{IQR Distribution}
        \label{fig:Distributions of Ministral-8B-Instruct-2410 wpwh d}
    \end{subfigure}
    %\caption{Distributions of metrics for all movie rating sequences on Mistral-8b-instruct under history + persona scenario (w/ History \& w/ Persona).}
    \caption{Results of SoS metrics for Mistral-8b-instruct in Scenario I (History + Persona).}
    \label{fig:Distributions of Ministral-8B-Instruct-2410 wpwh}
\end{figure*}

\section{Experiments}

\subsection{Experimental Setup}

\paragraph{Models.} 
We evaluate a range of backend LLMs, including one closed-source model, 
GPT-4o-mini, and six open-source models: 
DeepSeek-V2-Lite-Chat,
Mistral-8B-Instruct, and Qwen-2.5 series (1.5B, 3B, 7B). 
% This ensures that our findings are generalizable across different models and scales. 

%\mf{We use a diverse set of large language models (LLMs), comprising both a leading closed-source model, GPT-4o-mini}

\paragraph{Dataset.}
We construct our dataset from two main sources to support the evaluation framework.
\begin{itemize}[noitemsep, topsep=0pt, leftmargin=*]
    \item \textbf{Movies:} We scrape movie data from IMDb covering films released after January 12, 2025, including titles, genres, overviews, and IMDb average scores. 
    This cutoff date is selected to ensure that the movies fall outside the models’ training data, thereby preventing data contamination.
    \item \textbf{Personas:} 
    %\mf{
    We randomly sample
    %} 
    100 distinct profiles from the \texttt{elite\_persona} subset of the PersonaHub dataset~\cite{ge2024scaling}.
    %\mf{
    This subset is chosen for its coherent and information-rich descriptions, which support persona-based agent modeling.
    %}
    %This subset was chosen for its coherent and information-rich descriptions, which better support persona-based agent modeling. An example of a generated persona description is shown in Figure~\ref{fig:persona}.
\end{itemize}

%\begin{figure}[!t]
%\centering
%\begin{quote}
%\itshape
%A computer enthusiast who is interested in optimizing the performance of their system, particularly the CPU, GPU, and RAM. They are looking for software tools that can help them monitor and control the performance of their system, and they are willing to invest time in learning how to use these tools effectively. They are not necessarily looking for a professional-grade software tool, but rather a user-friendly and easy-to-use software that can provide comprehensive information about their system's performance and help them optimize it. They are also interested in software tools that can help them monitor the stability of their system after overclocking, as they want to avoid damaging their system.
%\end{quote}
%\caption{An Persona Example.}
%\label{fig:persona}
%\end{figure}

\paragraph{Implementation Details.} 
%\mf{
In our experiments, we use a rating scale of $M=10$ and fix the agent population size at $N=100$. For each movie, ratings are collected sequentially in a randomized order of agents. In the ``w/ History'' scenario, the prompt for the $n$-th agent includes the average rating of the preceding $n-1$ agents.
%}
%\zj{
The generation temperature is fixed at 0.1 in all experiments, and each agent’s final rating is the average of three independent model runs. 
For evaluation, ratings $\geq 6$ are considered positive and ratings $\leq 5$ negative when computing option trend metrics. For rating concentration metrics, we use $L=30$ final ratings, and we prepend $m=10$ randomly generated warm-up ratings to each sequence to mitigate early-stage volatility.
%}

\subsection{Results}

% \mz{To ensure the generalizability of our findings, our analysis focuses on two representative models: GPT-4o-mini, a closed-source model, and Mistral-8B-Instruct, an open-source model. We evaluate and compare their behavior across all four experimental scenarios.}

%\zj{
This section present detailed results for two representative models, the closed-source GPT-4o-mini and the open-source Mistral-8B-Instruct, across all four experimental scnarios. For the complete set of results covering all other models, please refer to Appendix \ref{sec:appendix-other-model-metrics}.
%}

\begin{figure*}[t]
    \centering
    \begin{subfigure}[t]{0.24\textwidth}
        \centering
        \includegraphics[width=\linewidth]{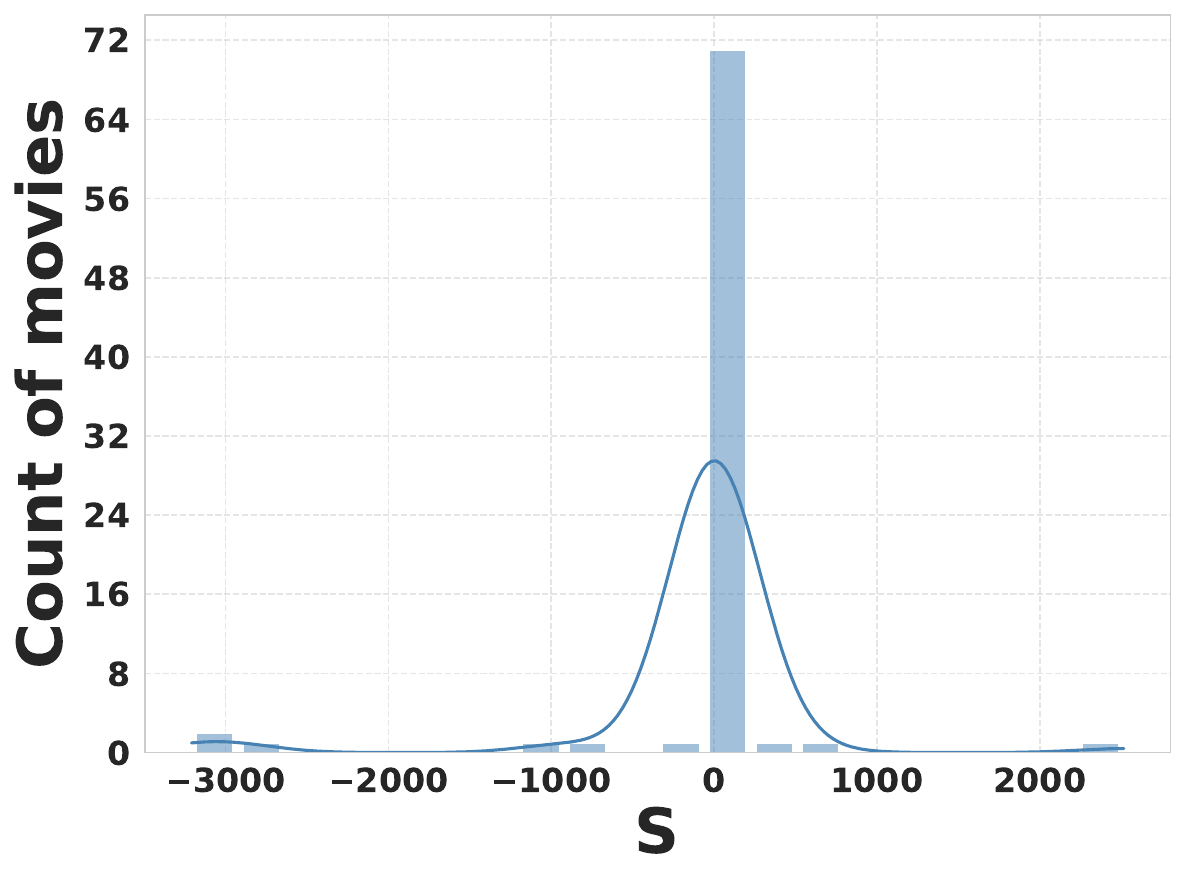}
        \caption{\small Mann–Kendall Distribution}
        \label{fig:Distributions of GPT-4o-mini opwh a}
    \end{subfigure}
    \hfill 
    \begin{subfigure}[t]{0.24\textwidth}
        \centering
        \includegraphics[width=\linewidth]{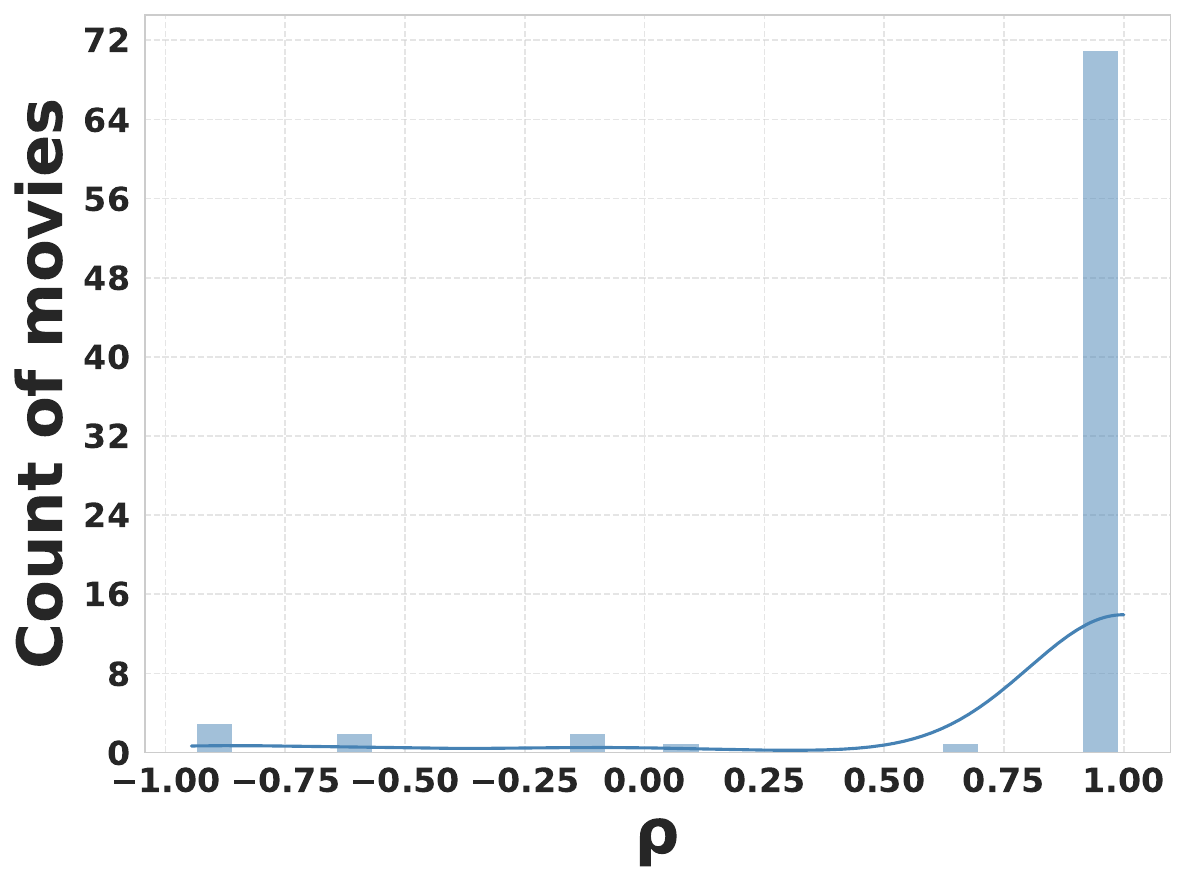}
        \caption{\small Spearman Distribution}
        \label{fig:Distributions of GPT-4o-mini opwh b}
    \end{subfigure}
    \hfill 
    \begin{subfigure}[t]{0.24\textwidth}
        \centering
        \includegraphics[width=\linewidth]{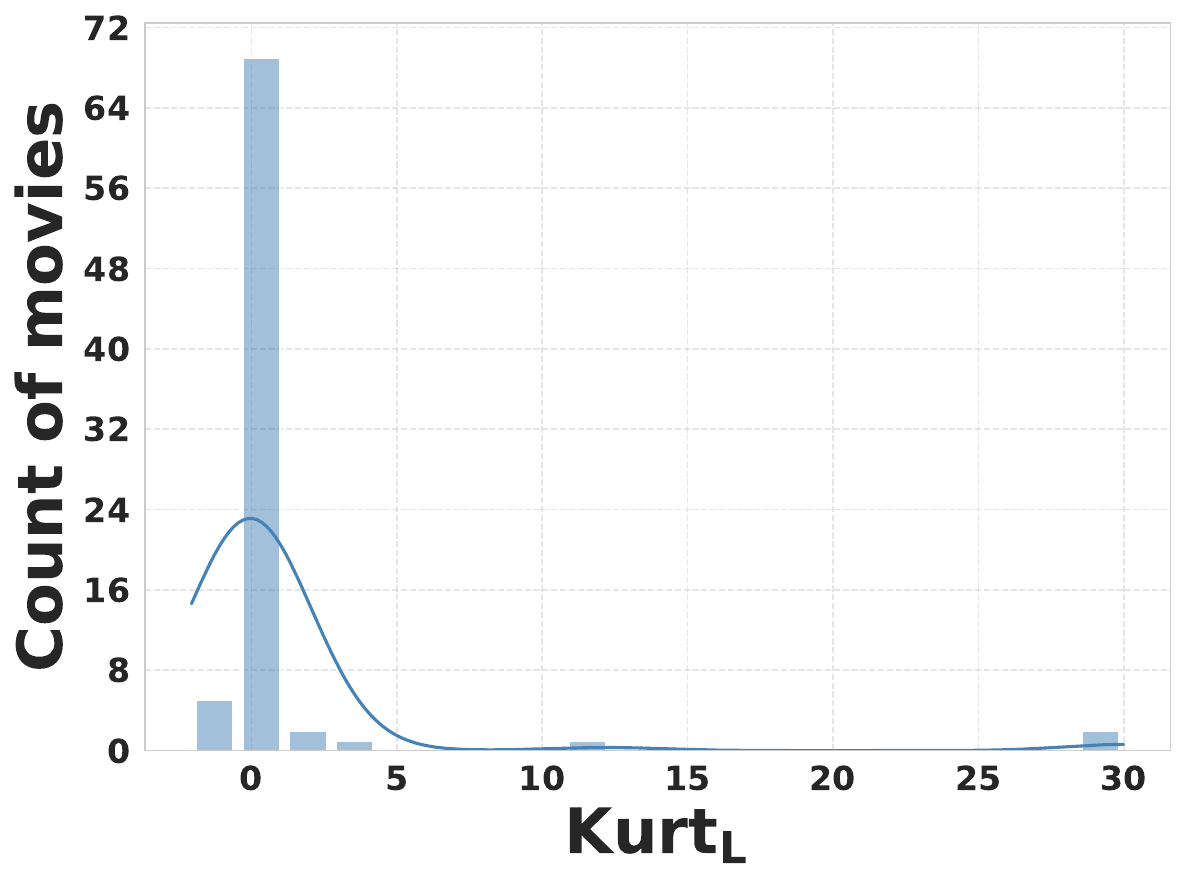}
        \caption{\small Kurtosis Distribution}
        \label{fig:Distributions of GPT-4o-mini opwh c}
    \end{subfigure}
    \hfill 
    \begin{subfigure}[t]{0.24\textwidth}
        \centering
        \includegraphics[width=\linewidth]{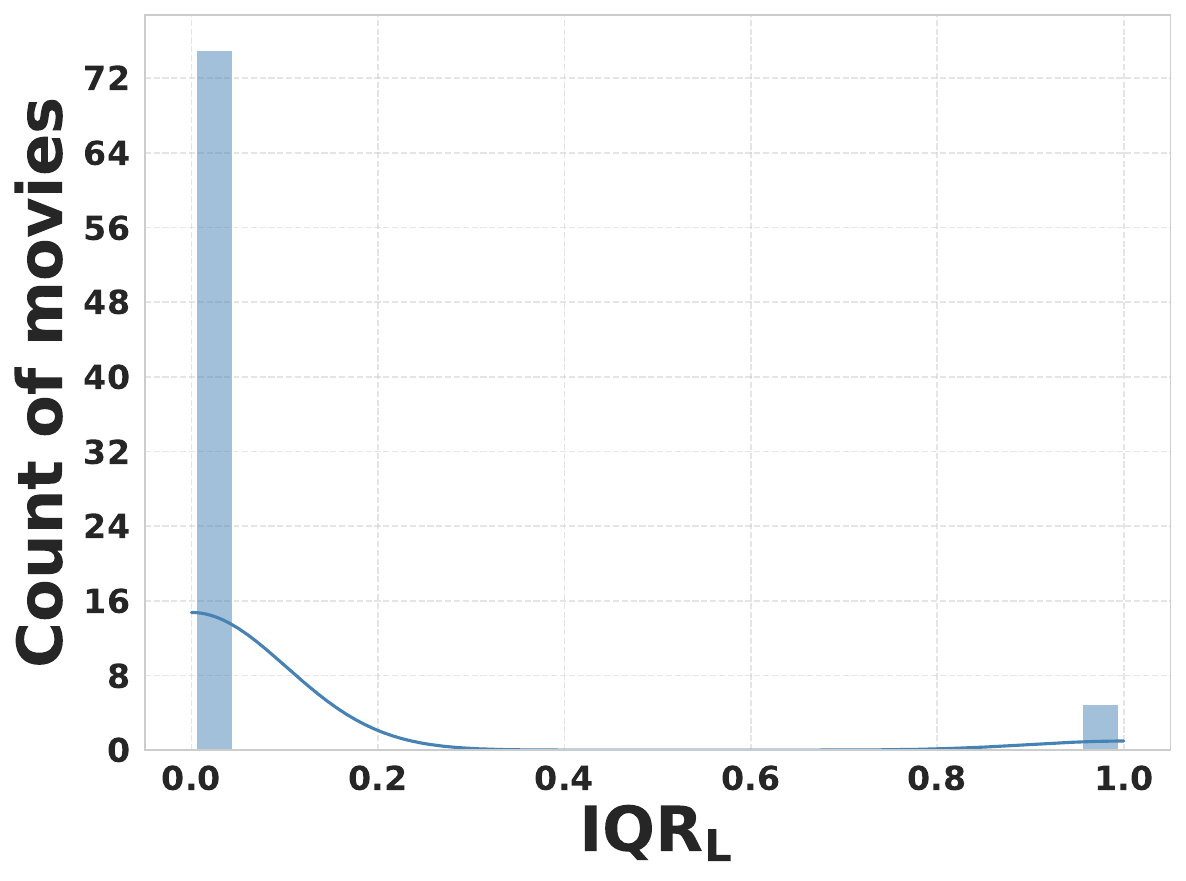}
        \caption{\small IQR Distribution}
        \label{fig:Distributions of GPT-4o-mini opwh d}
    \end{subfigure}
    %\caption{Distributions of metrics for all movie rating sequences on GPT-4o-mini under only history scenario (w/ History \& w/o Persona).}
    \caption{Results of SoS metrics for GPT-4o-mini in Scenario II (History Only).}
    \label{fig:Distributions of GPT-4o-mini opwh}
\end{figure*}

% \begin{figure*}[t]
% \centering
% \includegraphics[width=\linewidth]{figure/mistral-8b/Mistral-8B-Instruct-2410_ratings_no_persona_with_history_distribution.pdf}
% \caption{Distributions of Metrics for All Movie Rating Sequences on Mistral-8B-Instruct-2410 with ``w/ History \& w/o Persona''.}
% \label{fig:Distributions of Mistral-8B-Instruct-2410 opwh}
% \end{figure*}

\begin{figure*}[t]
    \centering
    \begin{subfigure}[t]{0.24\textwidth}
        \centering
        \includegraphics[width=\linewidth]{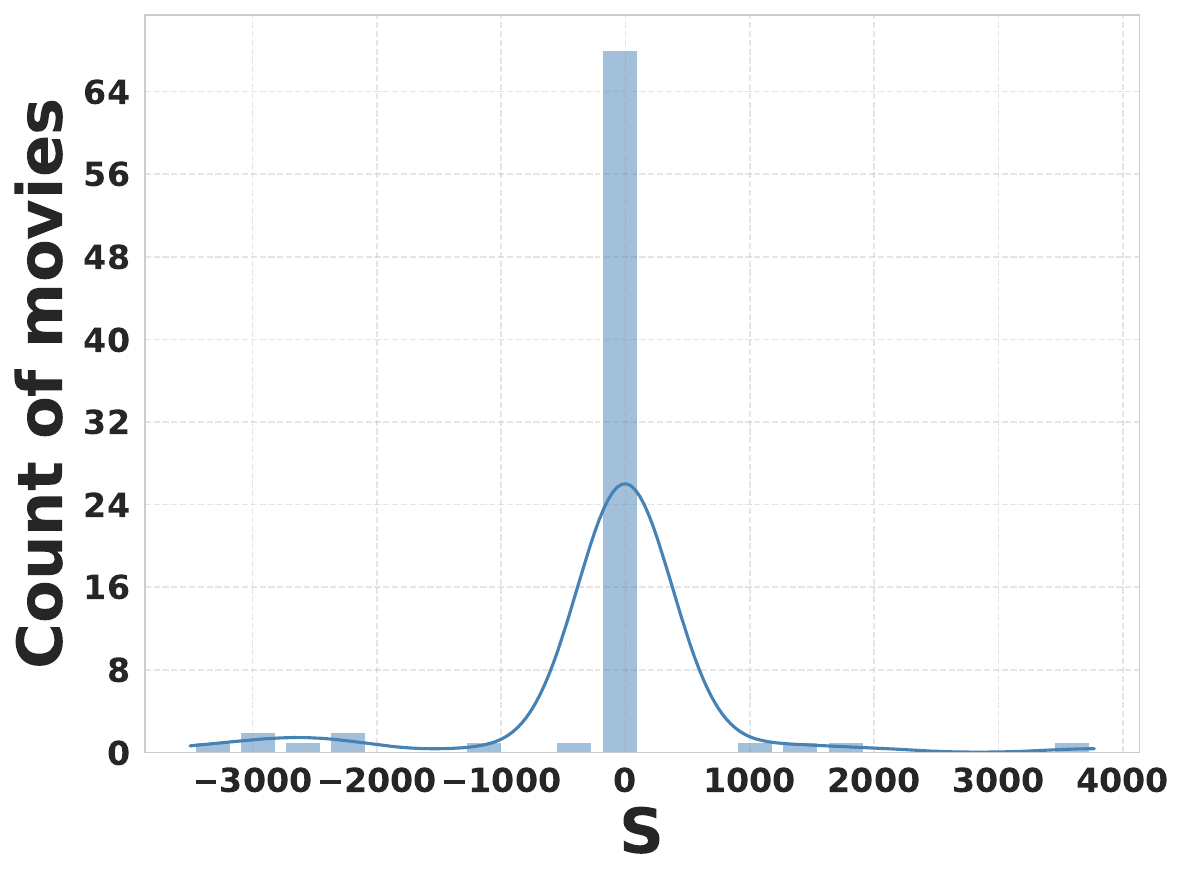}
        \caption{\small Mann–Kendall Distribution}
        \label{fig:Distributions of Mistral-8B-Instruct-2410 opwh a}
    \end{subfigure}
    \hfill 
    \begin{subfigure}[t]{0.24\textwidth}
        \centering
        \includegraphics[width=\linewidth]{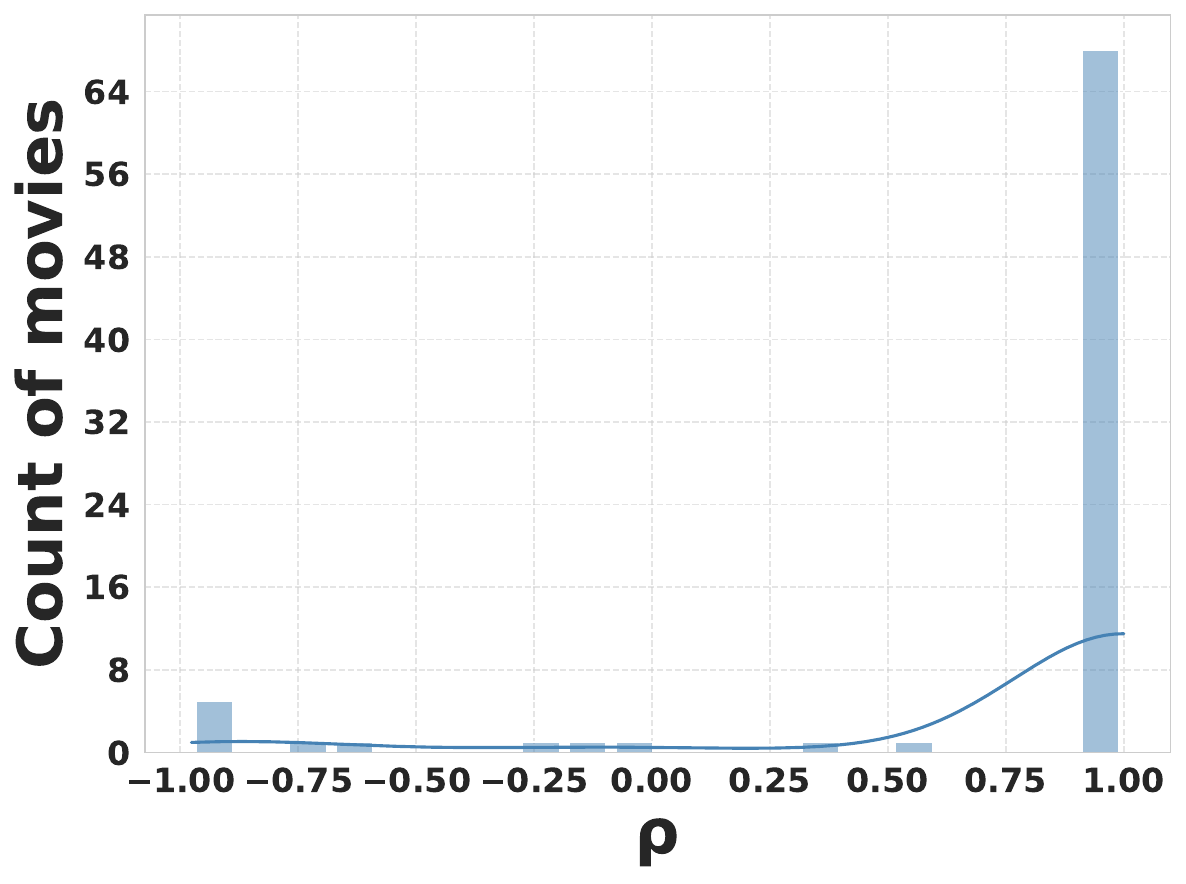}
        \caption{\small Spearman Distribution}
        \label{fig:Distributions of Mistral-8B-Instruct-2410 opwh b}
    \end{subfigure}
    \hfill 
    \begin{subfigure}[t]{0.24\textwidth}
        \centering
        \includegraphics[width=\linewidth]{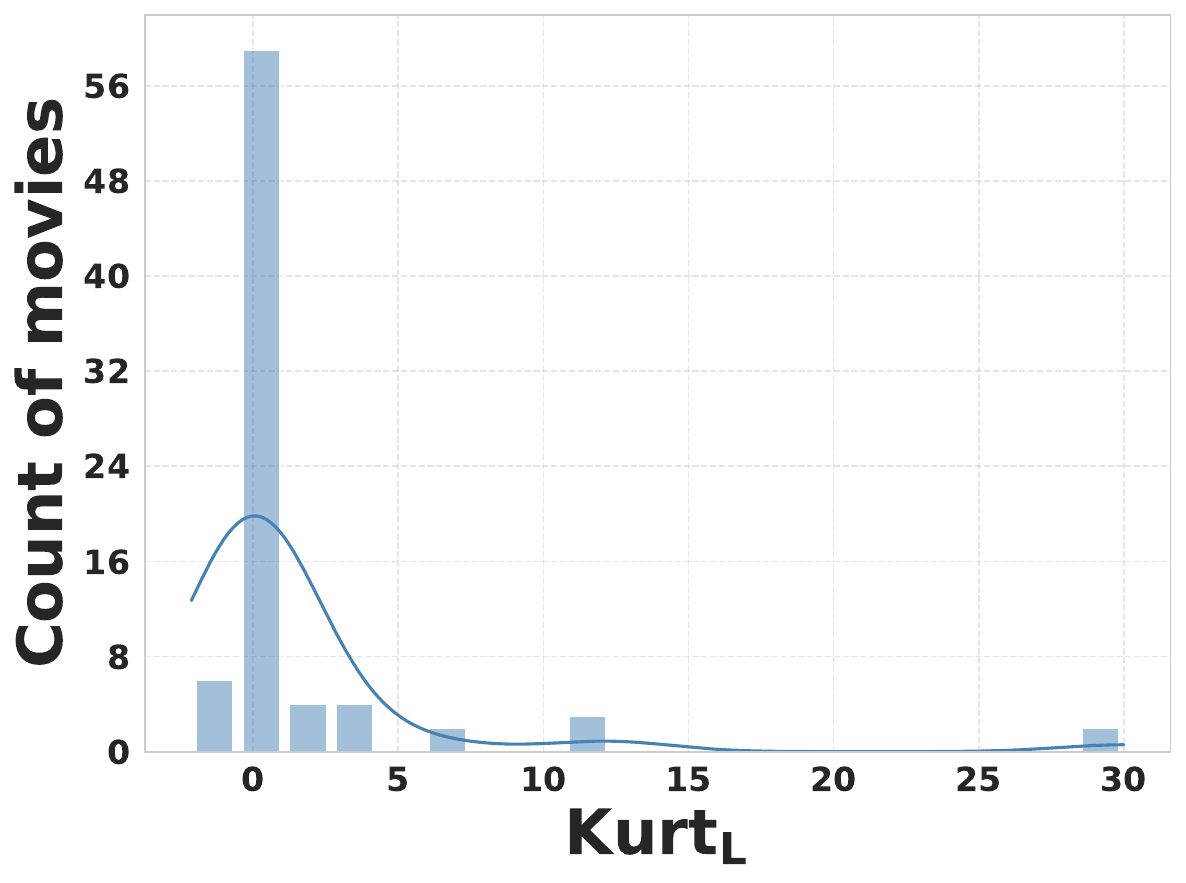}
        \caption{\small Kurtosis Distribution}
        \label{fig:Distributions of Mistral-8B-Instruct-2410 opwh c}
    \end{subfigure}
    \hfill 
    \begin{subfigure}[t]{0.24\textwidth}
        \centering
        \includegraphics[width=\linewidth]{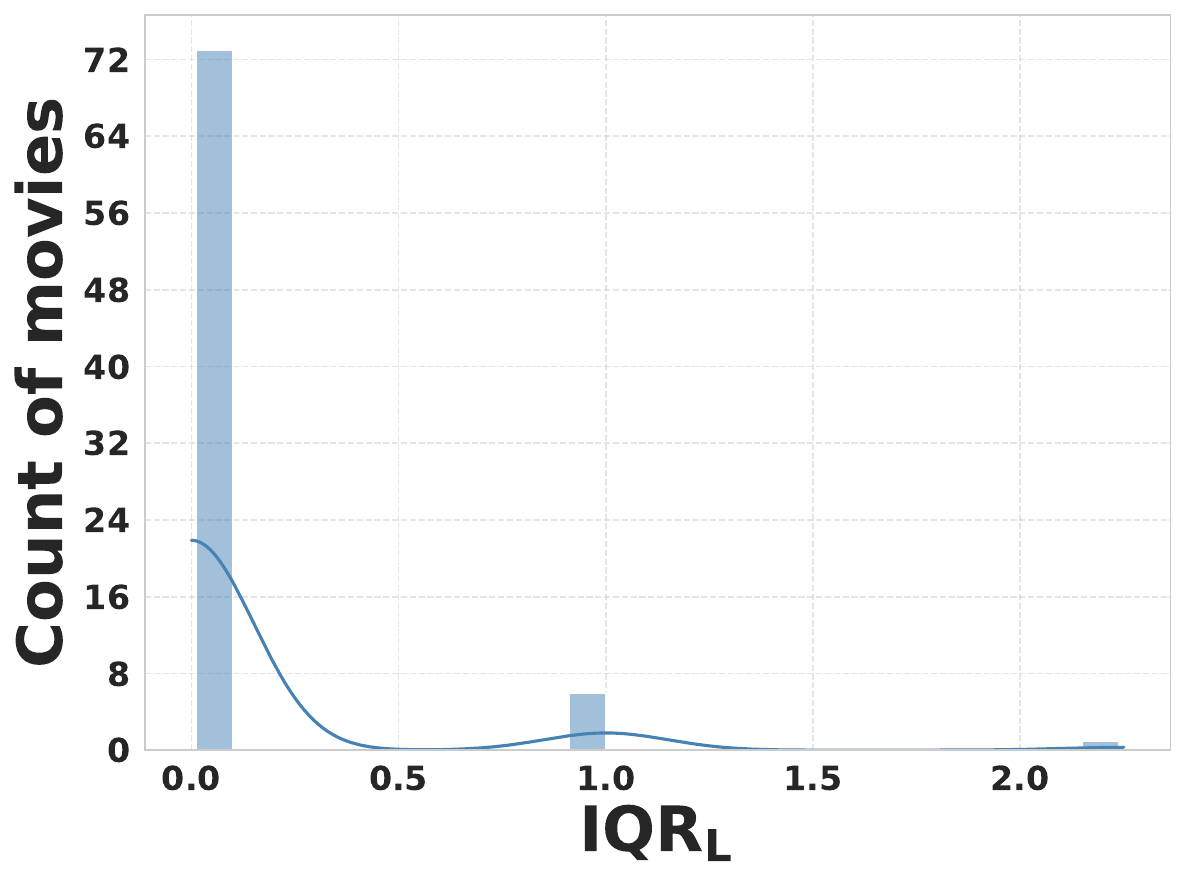}
        \caption{\small IQR Distribution}
        \label{fig:Distributions of Mistral-8B-Instruct-2410 opwh d}
    \end{subfigure}
    \caption{Results of SoS metrics for Mistral-8b-instruct in Scenario II (History Only).}
    \label{fig:Distributions of Mistral-8B-Instruct-2410 opwh}
\end{figure*}

\begin{figure*}[t]
    \centering
    \begin{subfigure}[t]{0.24\textwidth}
        \centering
        \includegraphics[width=\linewidth]{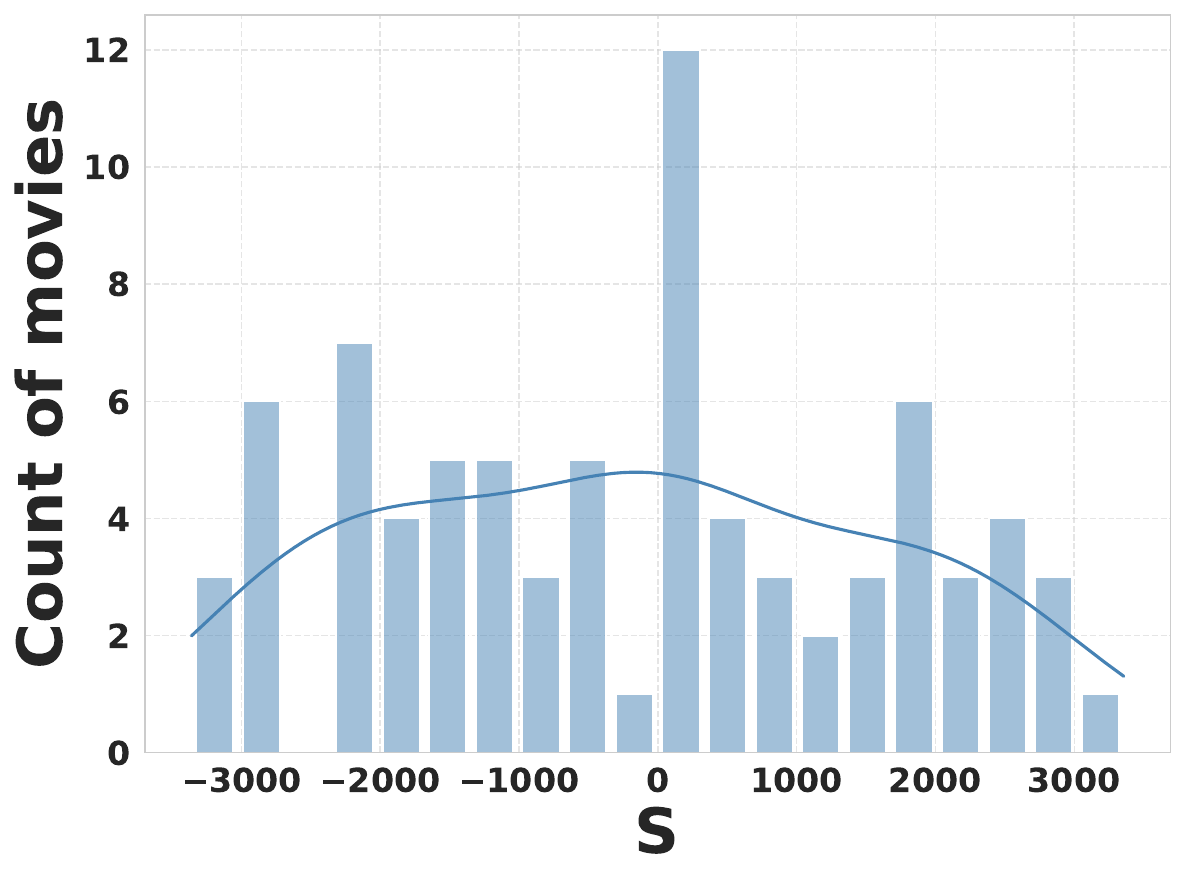}
        \caption{\small Mann–Kendall Distribution}
        \label{fig:Distributions of GPT-4o-mini wpoh a}
    \end{subfigure}
    \hfill 
    \begin{subfigure}[t]{0.24\textwidth}
        \centering
        \includegraphics[width=\linewidth]{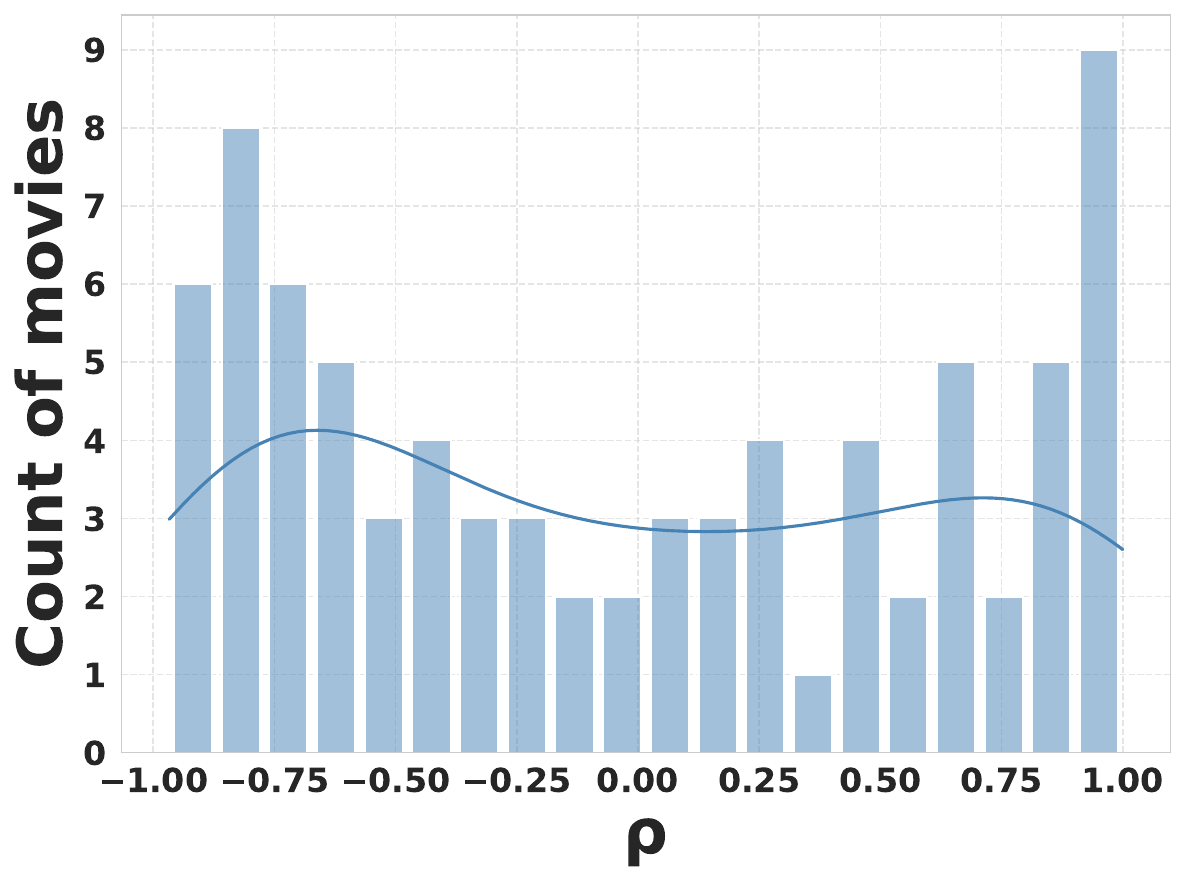}
        \caption{\small Spearman Distribution}
        \label{fig:Distributions of GPT-4o-mini wpoh b}
    \end{subfigure}
    \hfill 
    \begin{subfigure}[t]{0.24\textwidth}
        \centering
        \includegraphics[width=\linewidth]{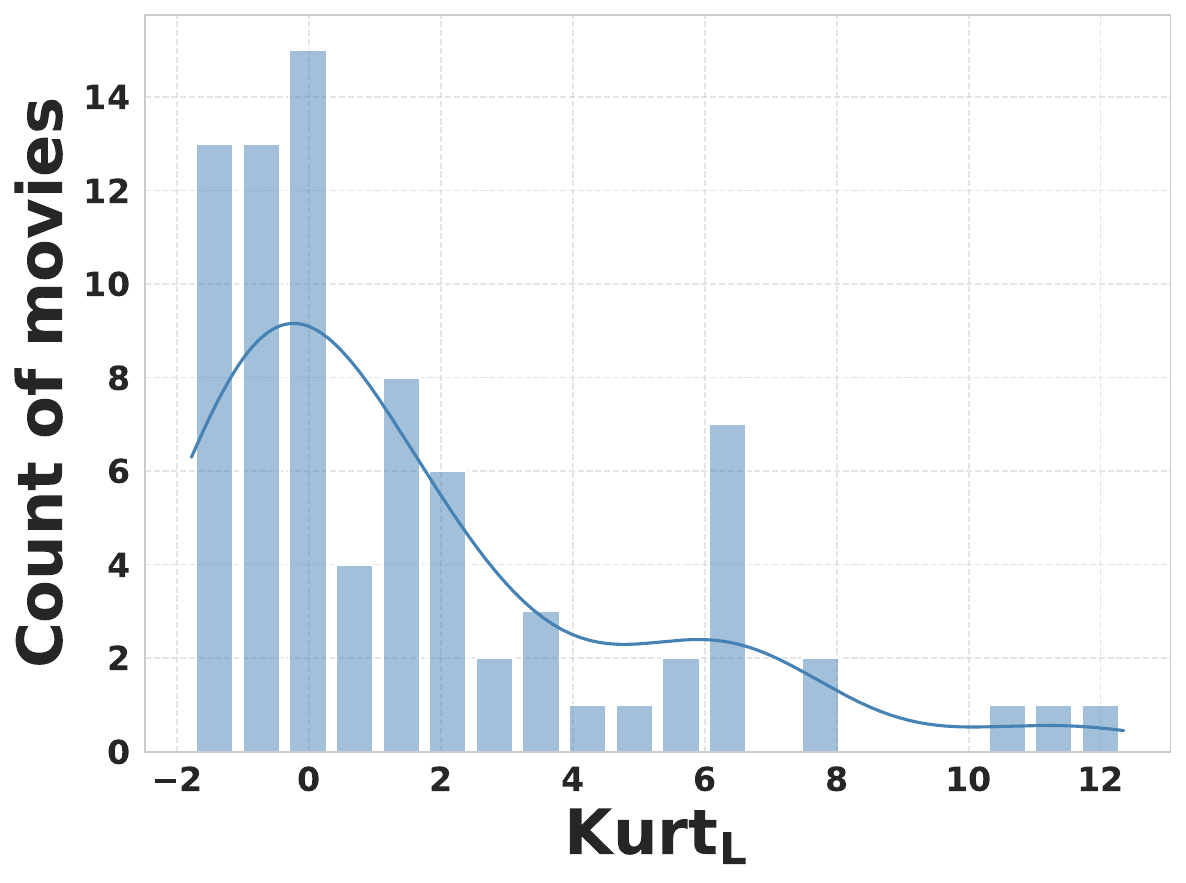}
        \caption{\small Kurtosis Distribution}
        \label{fig:Distributions of GPT-4o-mini wpoh c}
    \end{subfigure}
    \hfill 
    \begin{subfigure}[t]{0.24\textwidth}
        \centering
        \includegraphics[width=\linewidth]{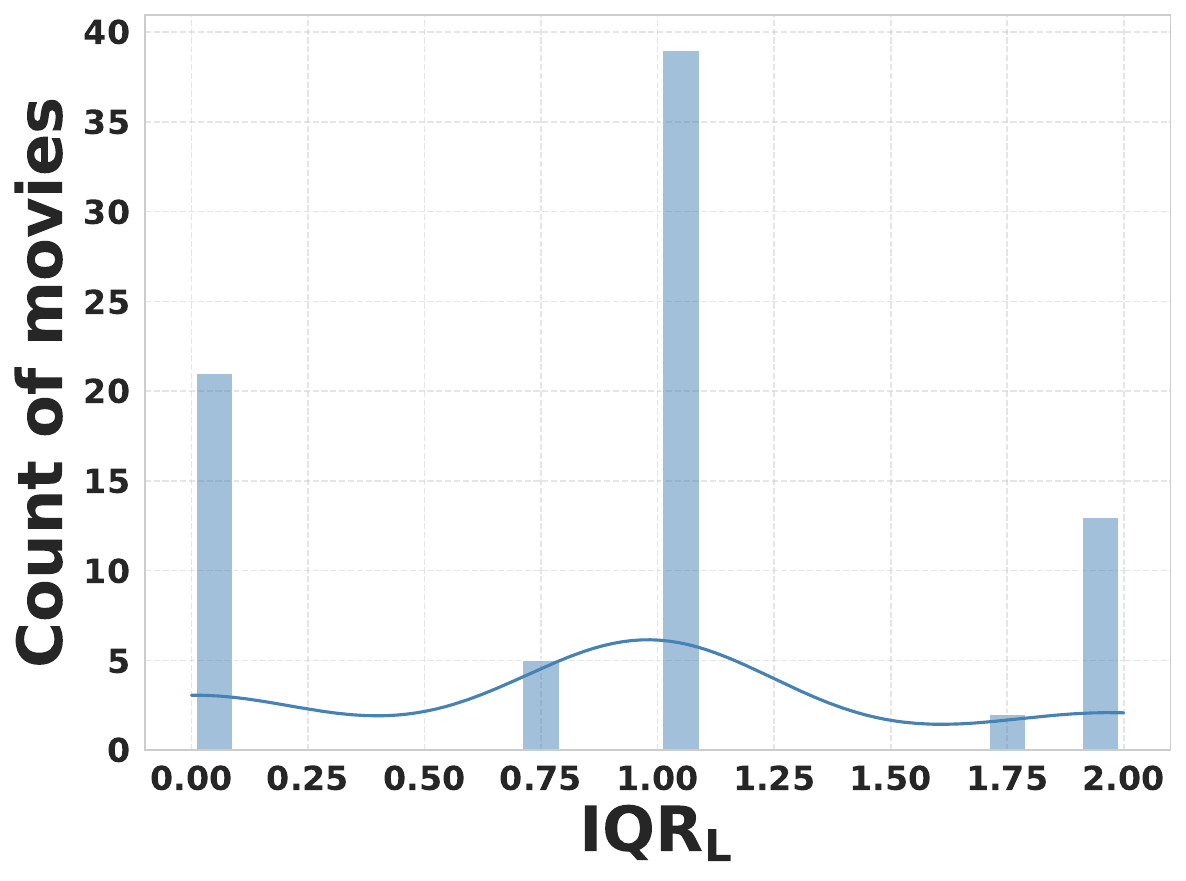}
        \caption{\small IQR Distribution}
        \label{fig:Distributions of GPT-4o-mini wpoh d}
    \end{subfigure}
    \caption{Results of SoS metrics for GPT-4o-mini in Scenario III (Persona Only).}
    \label{fig:Distributions of GPT-4o-mini wpoh}
\end{figure*}

\begin{figure*}[t]
    \centering
    \begin{subfigure}[t]{0.24\textwidth}
        \centering
        \includegraphics[width=\linewidth]{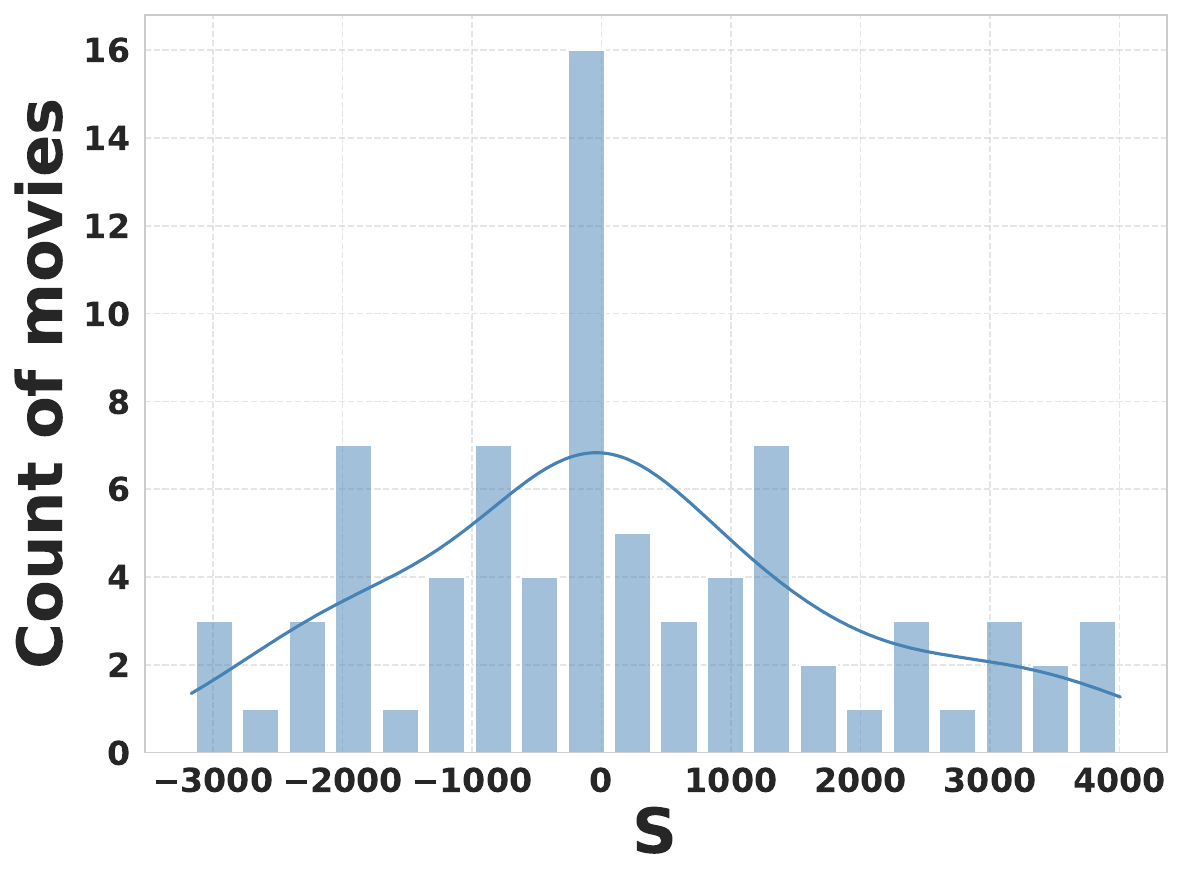}
        \caption{\small Mann–Kendall Distribution}
        \label{fig:Distributions of Mistral-8B-Instruct-2410 wpoh a}
    \end{subfigure}
    \hfill 
    \begin{subfigure}[t]{0.24\textwidth}
        \centering
        \includegraphics[width=\linewidth]{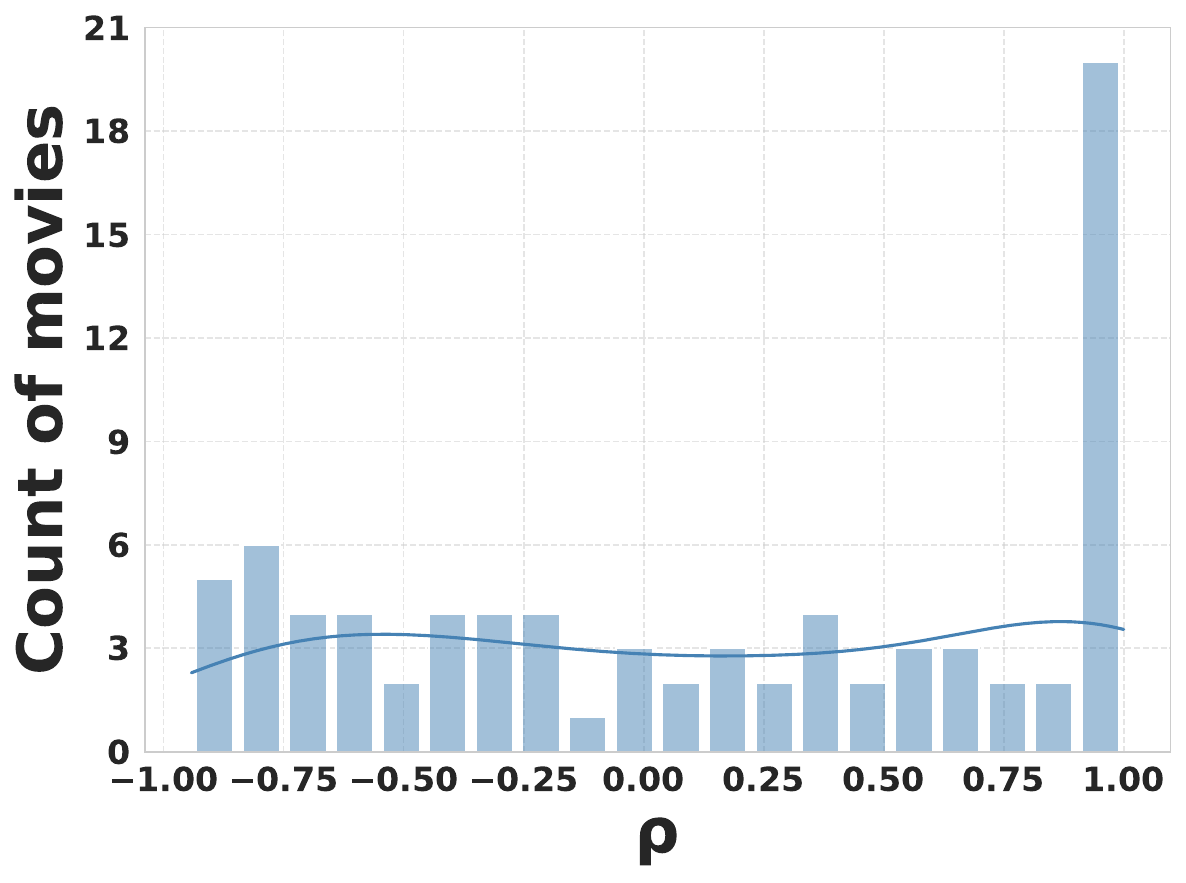}
        \caption{\small Spearman Distribution}
        \label{fig:Distributions of Mistral-8B-Instruct-2410 wpoh b}
    \end{subfigure}
    \hfill 
    \begin{subfigure}[t]{0.24\textwidth}
        \centering
        \includegraphics[width=\linewidth]{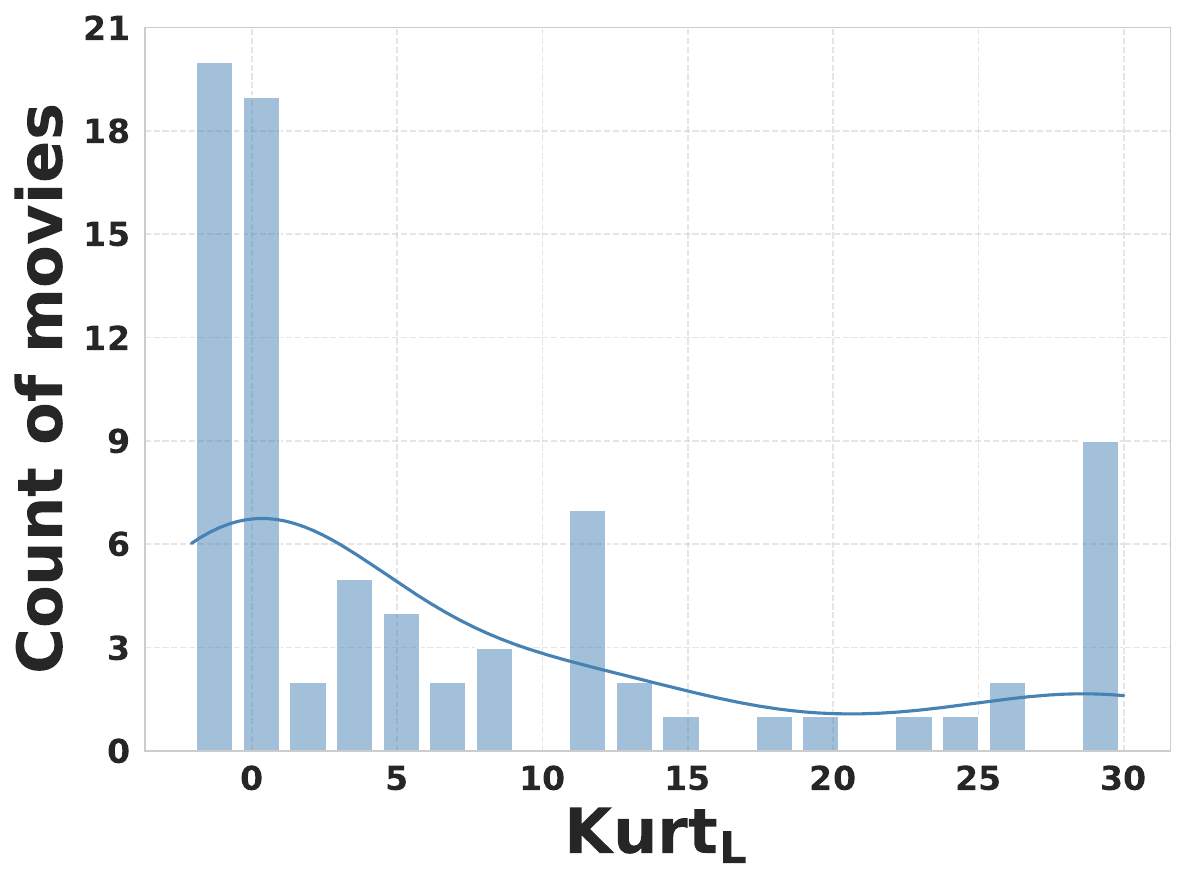}
        \caption{\small Kurtosis Distribution}
        \label{fig:Distributions of Mistral-8B-Instruct-2410 wpoh c}
    \end{subfigure}
    \hfill 
    \begin{subfigure}[t]{0.24\textwidth}
        \centering
        \includegraphics[width=\linewidth]{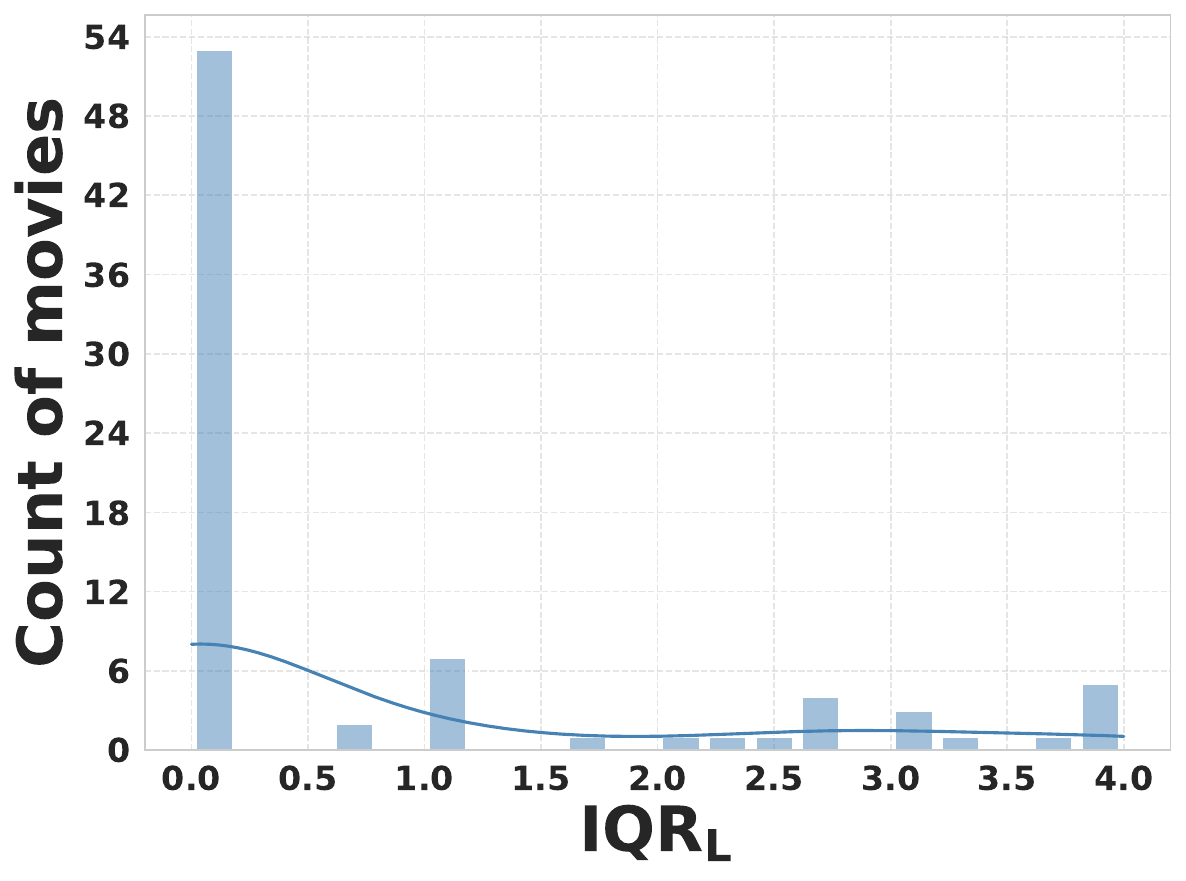}
        \caption{\small IQR Distribution}
        \label{fig:Distributions of Mistral-8B-Instruct-2410 wpoh d}
    \end{subfigure}
    \caption{Results of SoS metrics for Mistral-8b-instruct in Scenario III (Persona Only).}
    \label{fig:Distributions of Mistral-8b-instruct wpoh}
\end{figure*}

\paragraph{Scenario I (History + Persona):} 
\textbf{SoS emerges when both history and persona signals are present.}  
The metric distributions for GPT-4o-mini in Figure~\ref{fig:Distributions of GPT-4o-mini wpwh} and for Mistral-8B-Instruct in Figure~\ref{fig:Distributions of Ministral-8B-Instruct-2410 wpwh} show a clear progression from early disagreement to final consensus when both history and persona signals are present. This pattern provides strong evidence for the SoS effect, reflected in two key dimensions:  

\begin{itemize}[noitemsep, topsep=0pt, leftmargin=*]
    \item \textbf{Opinion Trend:} 
    Both models exhibit a positive monotonic trend. As shown in Figure \ref{fig:Distributions of GPT-4o-mini wpwh a}, GPT-4o-mini's Mann-Kendall ($S$) statistic tends to be positively skewed, and its Spearman's rank correlation ($\rho$) has a strong peak near 1.0 as shown in Figure \ref{fig:Distributions of GPT-4o-mini wpwh b}. In Figures \ref{fig:Distributions of Ministral-8B-Instruct-2410 wpwh a} and \ref{fig:Distributions of Ministral-8B-Instruct-2410 wpwh b}, Mistral-8B-Instruct shows an even more pronounced trend, with its ($\rho$) distribution concentrated at 1.0. These results indicate that majority opinions undergo a gradually reinforcing and monotonic convergence process, consistent with the SoS effect.
    \item \textbf{Rating Concentration:}
    Both models’ ratings indicate a high degree of final consensus. As shown in Figures~\ref{fig:Distributions of GPT-4o-mini wpwh c} and~\ref{fig:Distributions of GPT-4o-mini wpwh d}, GPT-4o-mini exhibits a positively skewed kurtosis distribution and an IQR distribution with distinct peaks near 0 and 1. Mistral-8B-Instruct displays a similar pattern, as illustrated in Figures~\ref{fig:Distributions of Ministral-8B-Instruct-2410 wpwh c} and~\ref{fig:Distributions of Ministral-8B-Instruct-2410 wpwh d}, with kurtosis skewed toward higher values and IQR values heavily concentrated at low values. These distributions suggest a substantial reduction in opinion diversity at the final stage.
\end{itemize}

% \begin{figure*}[t]
% \centering
% \includegraphics[width=\linewidth]{figure/gpt-4o-mini/gpt-4o-mini_ratings_with_history_distribution.pdf}
% \caption{Distributions of Metrics for All Movie Rating Sequences on GPT-4o-mini with ``w/ History \& w/ Persona''.}
% \label{fig:Distributions of GPT-4o-mini wpwh}
% \end{figure*}

% \begin{figure*}[t]
% \centering
% \includegraphics[width=\linewidth]{figure/mistral-8b/Mistral-8B-Instruct-2410_ratings_with_history_distribution.pdf}
% \caption{Distributions of Metrics for All Movie Rating Sequences on Mistral-8B-Instruct with ``w/ History \& w/ Persona''.}
% \label{fig:Distributions of Mistral-8B-Instruct-2410 wpwh}
% \end{figure*}

\paragraph{Scenario II (History only): Anchoring collective opinion appears when only history signals are present.}
When only the history signal is available 
% (w/ History \& w/o Persona)
, the rating distributions for GPT-4o-mini and Mistral-8B-Instruct, as shown in Figures~\ref{fig:Distributions of GPT-4o-mini opwh} and~\ref{fig:Distributions of Mistral-8B-Instruct-2410 opwh}, exhibit a clear anchoring effect. This can be observed from two aspects.

\begin{itemize}[noitemsep, topsep=0pt, leftmargin=*]
    \item \textbf{Opinion Trend:} 
    The lack of dynamic opinion evolution is evident in both models. The Mann-Kendall ($S$) distributions, as shown in Figures \ref{fig:Distributions of GPT-4o-mini opwh a} and \ref{fig:Distributions of Mistral-8B-Instruct-2410 opwh a}, show an extremely sharp peak around 0, indicating a lack of any monotonic trend. The Spearman values ($\rho$) are heavily concentrated around 1.0, as shown in Figures \ref{fig:Distributions of GPT-4o-mini opwh b} and \ref{fig:Distributions of Mistral-8B-Instruct-2410 opwh b}. These patterns reflects a strong anchoring effect, where initial opinions constrain later ratings, leading to minimal variation throughout the sequence.
    \item \textbf{Rating Concentration:} The concentration metrics clearly illustrate the outcome of anchoring effect. As shown in Figures~\ref{fig:Distributions of GPT-4o-mini opwh c} and~\ref{fig:Distributions of GPT-4o-mini opwh d} for GPT-4o-mini, and Figures~\ref{fig:Distributions of Mistral-8B-Instruct-2410 opwh c} and~\ref{fig:Distributions of Mistral-8B-Instruct-2410 opwh d} for Mistral-8B-Instruct, both models converge to a state of extreme and stable consensus. The IQR values are heavily concentrated at 0, indicating minimal variation across agents, while the strongly positive kurtosis values reflect peaked, tightly clustered rating distributions. These patterns confirm that once early opinions are established, later responses remain fixed, resulting in highly uniform outcomes.
\end{itemize}

% \begin{figure*}[t]
% \centering
% \includegraphics[width=\linewidth]{figure/gpt-4o-mini/gpt-4o-mini_ratings_no_persona_with_history_distribution.pdf}
% \caption{Distributions of Metrics for All Movie Rating Sequences on GPT-4o-mini with ``w/ History \& w/o Persona''.}
% \label{fig:Distributions of GPT-4o-mini opwh}
% \end{figure*}

\begin{figure*}[t]
    \centering
    \begin{subfigure}[t]{0.24\textwidth}
        \centering
        \includegraphics[width=\linewidth]{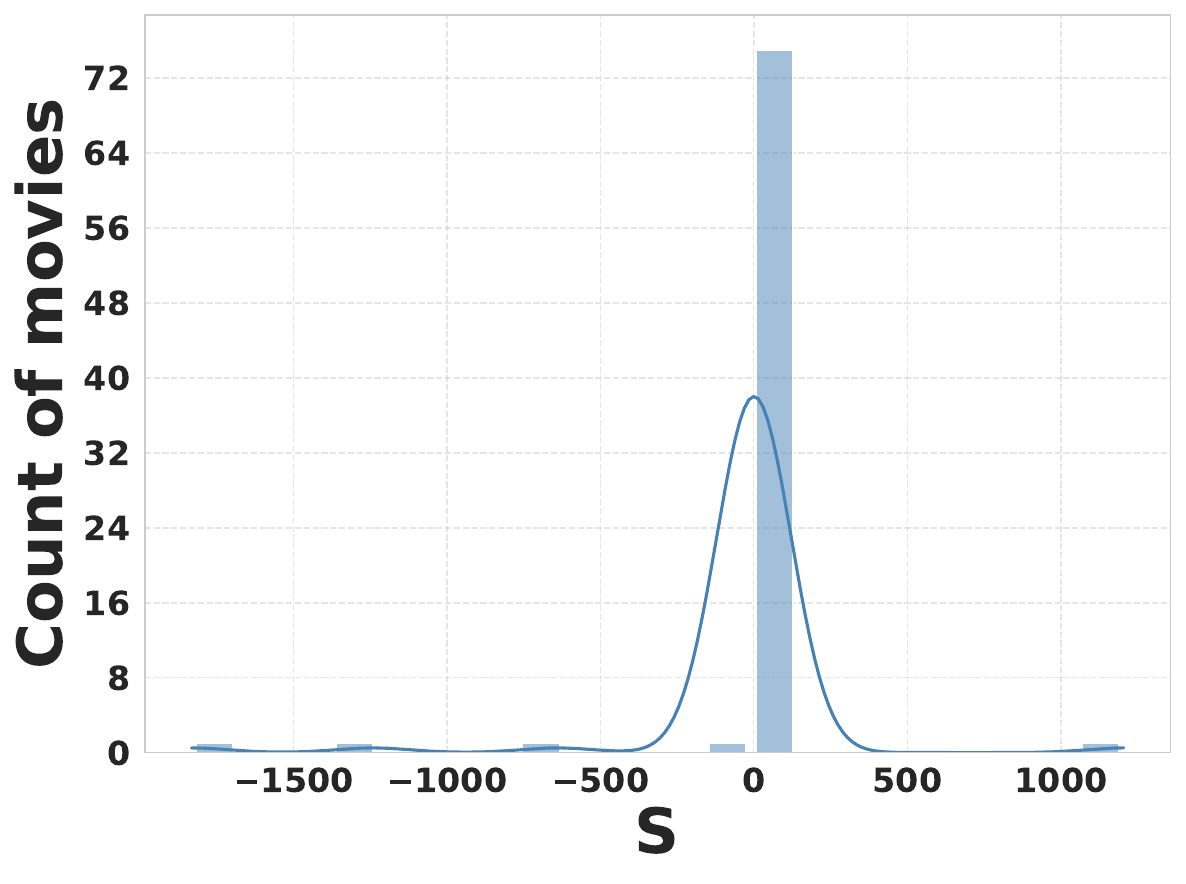}
        \caption{\small Mann–Kendall Distribution}
        \label{fig:Distributions of GPT-4o-mini opoh a}
    \end{subfigure}
    \hfill 
    \begin{subfigure}[t]{0.24\textwidth}
        \centering
        \includegraphics[width=\linewidth]{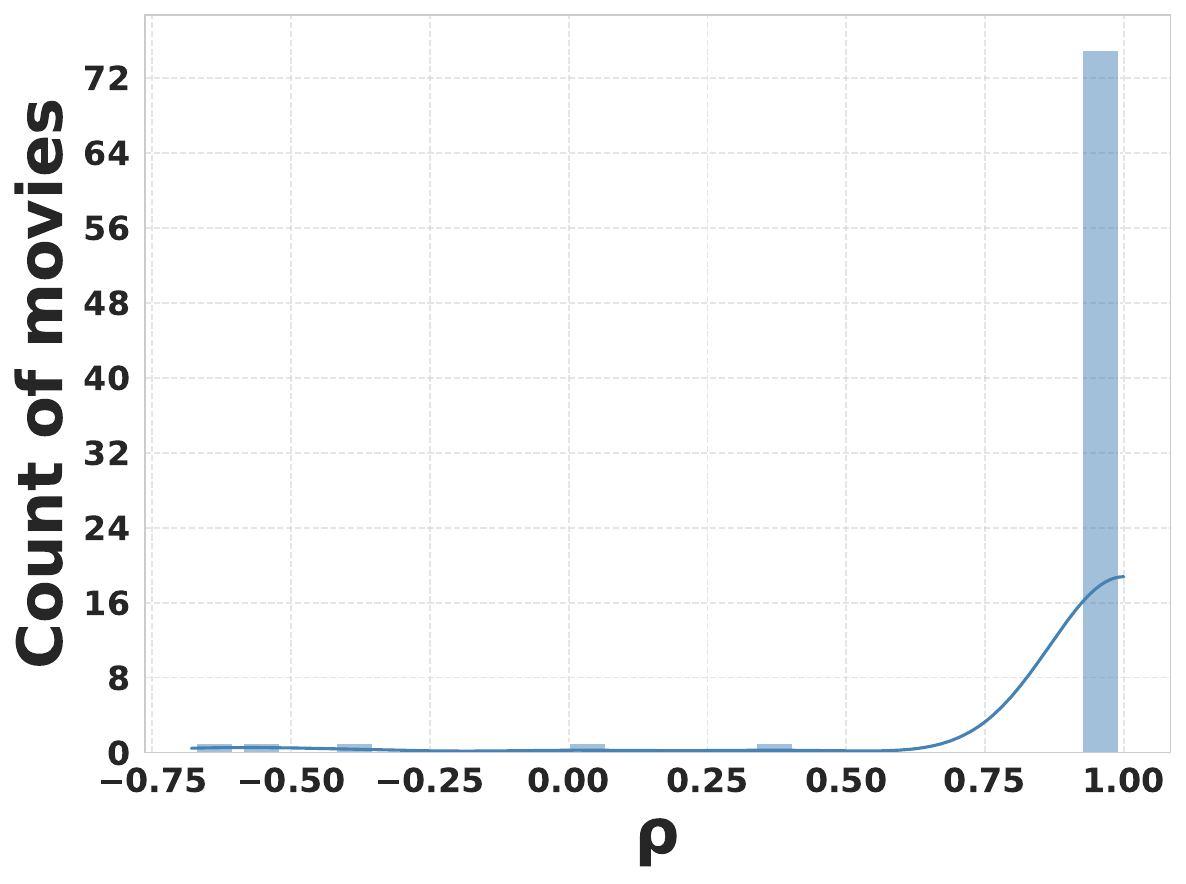}
        \caption{\small Spearman Distribution}
        \label{fig:Distributions of GPT-4o-mini opoh wpoh b}
    \end{subfigure}
    \hfill 
    \begin{subfigure}[t]{0.24\textwidth}
        \centering
        \includegraphics[width=\linewidth]{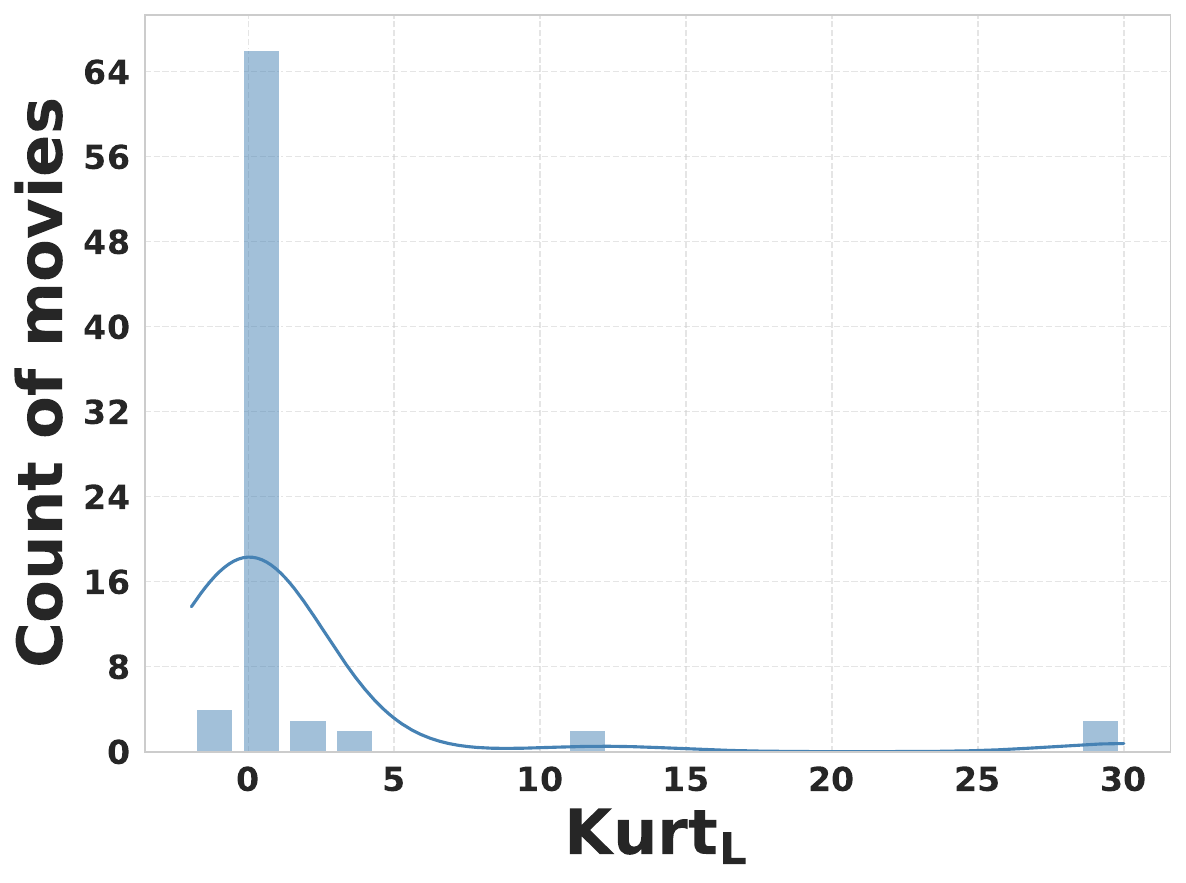}
        \caption{\small Kurtosis Distribution}
        \label{fig:Distributions of GPT-4o-mini opoh c}
    \end{subfigure}
    \hfill 
    \begin{subfigure}[t]{0.24\textwidth}
        \centering
        \includegraphics[width=\linewidth]{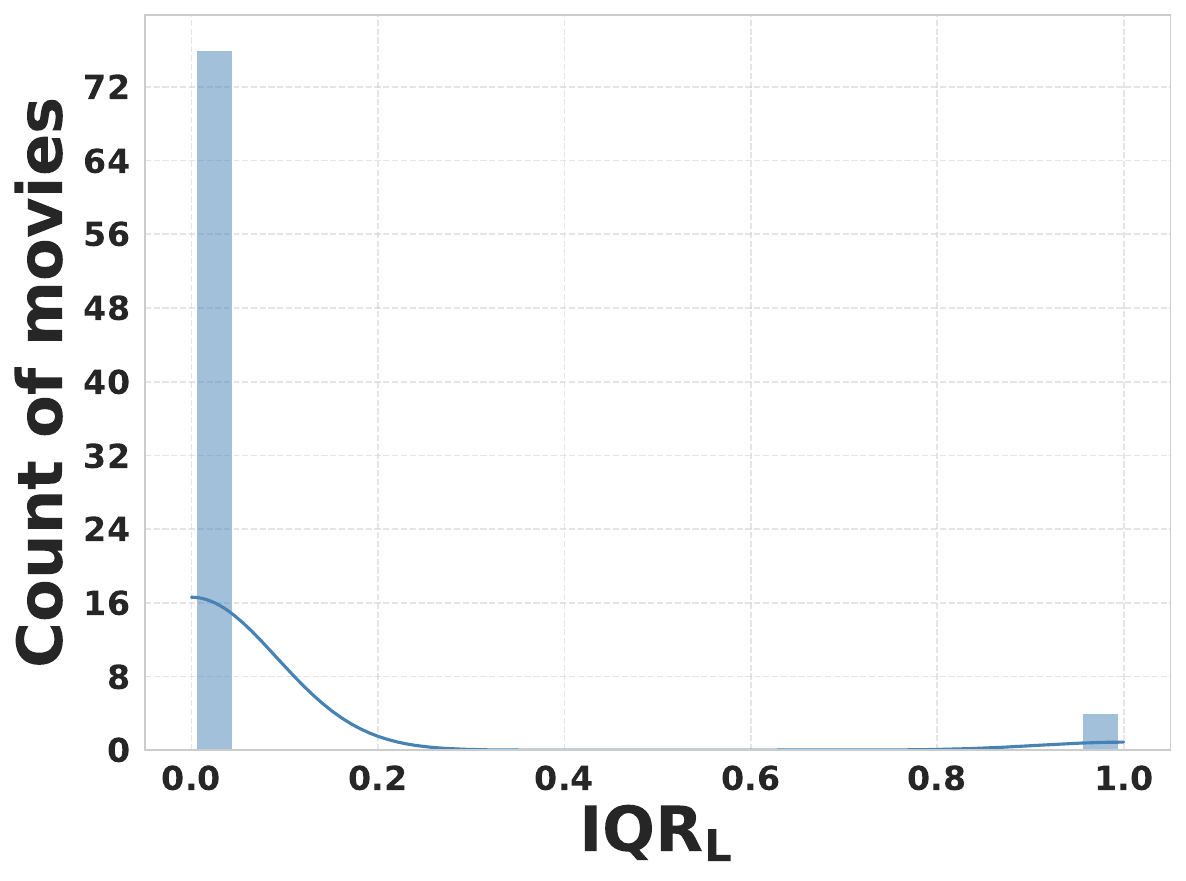}
        \caption{\small IQR Distribution}
        \label{fig:Distributions of GPT-4o-mini opoh d}
    \end{subfigure}
    \caption{Results of SoS metrics for GPT-4o-mini in Scenario IV (No History, No Persona).}
    \label{fig:Distributions of GPT-4o-mini opoh}
\end{figure*}

% \begin{figure*}[t]
% \centering
% \includegraphics[width=\linewidth]{figure/mistral-8b/Mistral-8B-Instruct-2410_ratings_no_persona_no_history_distribution.pdf}
% \caption{Distributions of Metrics for All Movie Rating Sequences on Mistral-8B-Instruct-2410 with ``w/o History \& w/o Persona''.}
% \label{fig:Distributions of Mistral-8B-Instruct-2410 opoh}
% \end{figure*}

\begin{figure*}[t]
    \centering
    \begin{subfigure}[t]{0.24\textwidth}
        \centering
        \includegraphics[width=\linewidth]{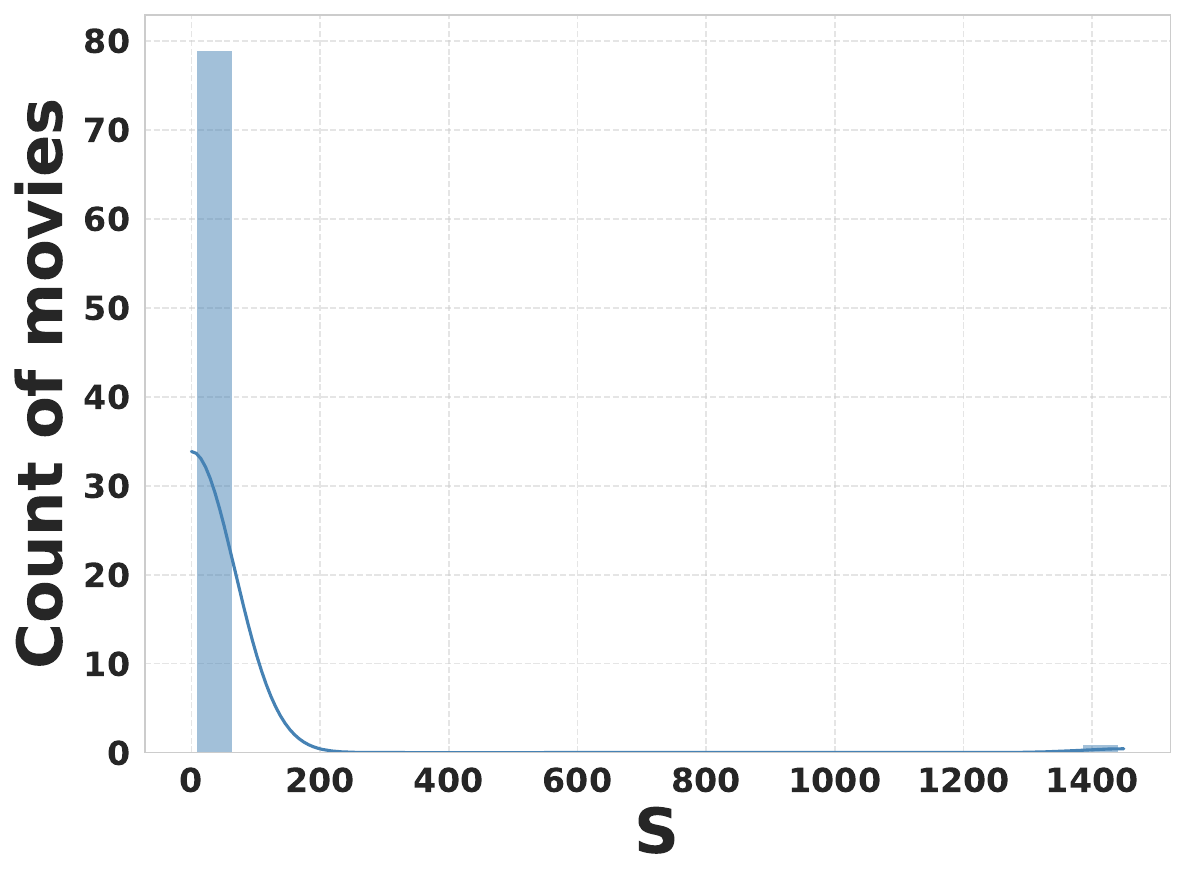}
        \caption{\small Mann–Kendall Distribution}
        \label{fig:Distributions of Mistral-8B-Instruct-2410 opoh a}
    \end{subfigure}
    \hfill 
    \begin{subfigure}[t]{0.24\textwidth}
        \centering
        \includegraphics[width=\linewidth]{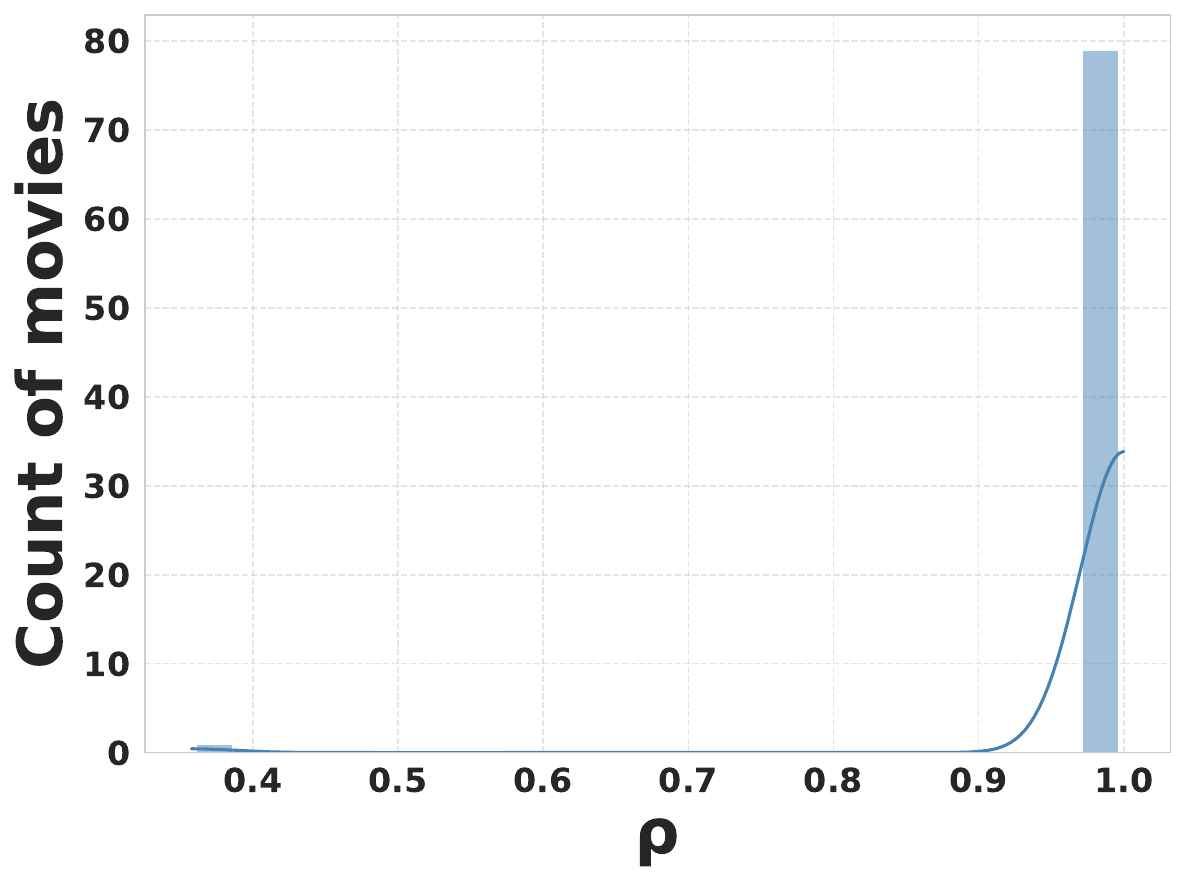}
        \caption{\small Spearman Distribution}
        \label{fig:Distributions of Mistral-8B-Instruct-2410 opoh b}
    \end{subfigure}
    \hfill 
    \begin{subfigure}[t]{0.24\textwidth}
        \centering
        \includegraphics[width=\linewidth]{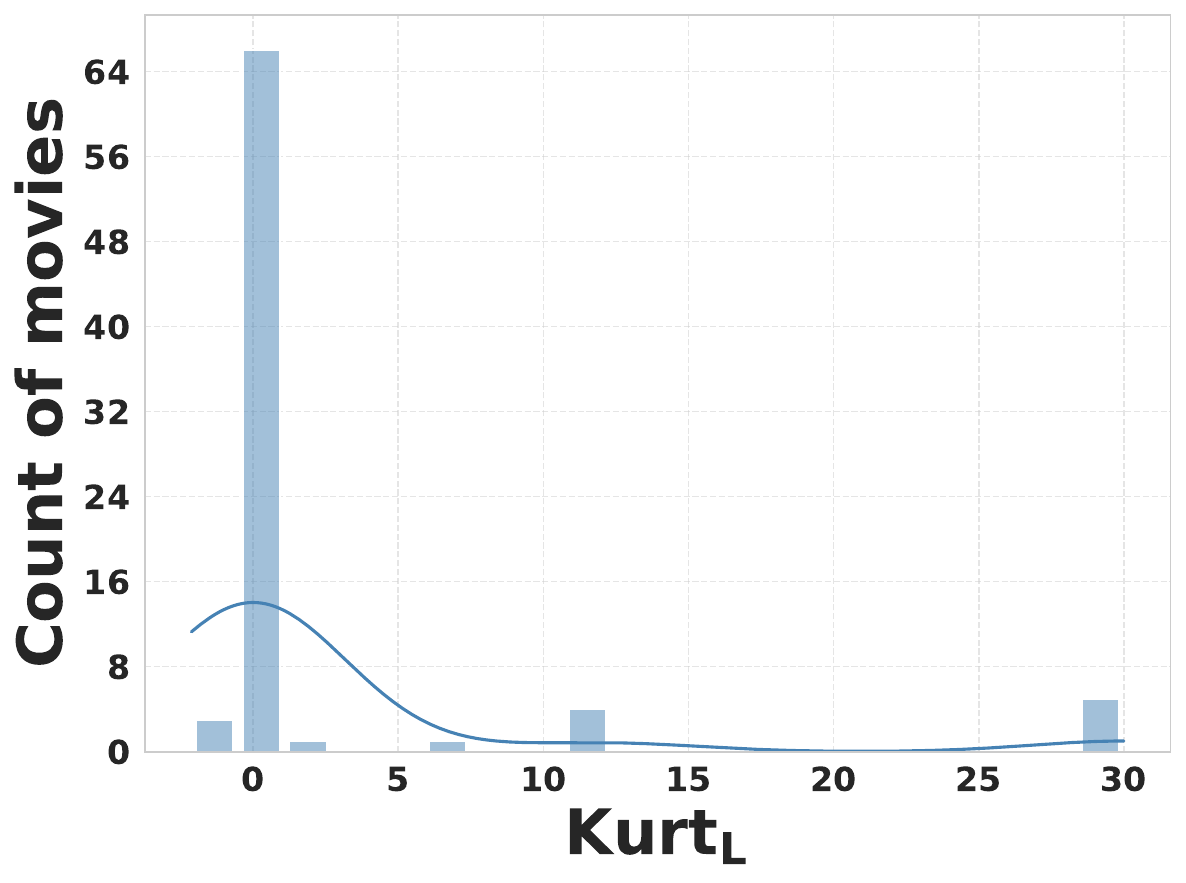}
        \caption{\small Kurtosis Distribution}
        \label{fig:Distributions of Mistral-8B-Instruct-2410 opoh c}
    \end{subfigure}
    \hfill 
    \begin{subfigure}[t]{0.24\textwidth}
        \centering
        \includegraphics[width=\linewidth]{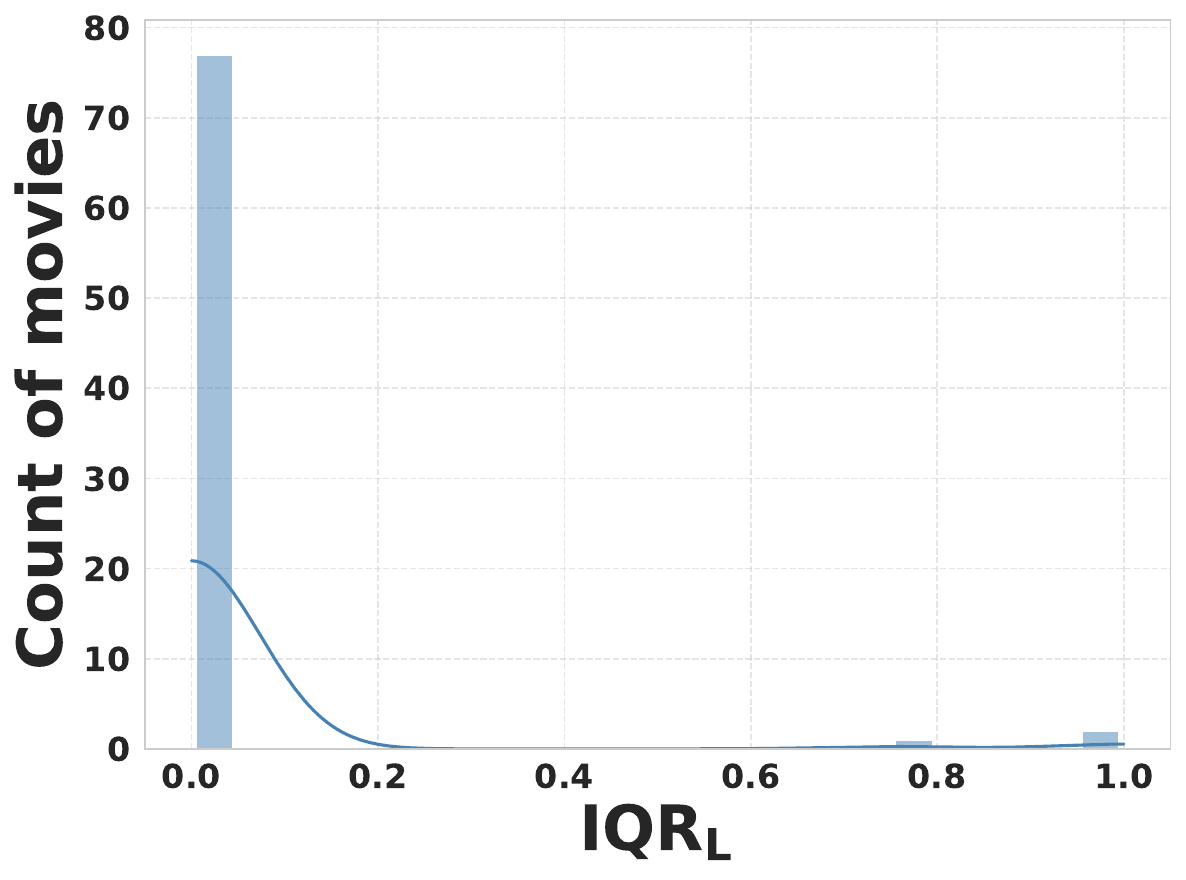}
        \caption{\small IQR Distribution}
        \label{fig:Distributions of Mistral-8B-Instruct-2410 opoh d}
    \end{subfigure}
    \caption{Results of SoS metrics for Mistral-8b-instruct in Scenario IV (No History, No Persona).}
    \label{fig:Distributions of Mistral-8B-Instruct-2410 opoh}
\end{figure*}

\paragraph{Scenario III (Persona only): Promoting opinion diversity appears when only persona signals are present.}

With only persona signals 
% (w/o History \& w/ Persona)
, the distributions for GPT-4o-mini and Mistral-8B-Instruct, as shown in Figure \ref{fig:Distributions of GPT-4o-mini wpoh} and in Figure  \ref{fig:Distributions of Mistral-8b-instruct wpoh}, show that both models foster opinion diversity without exhibiting any clear convergence.

\begin{itemize}[noitemsep, topsep=0pt, leftmargin=*]
    \item \textbf{Opinion Trend:} As shown in Figures \ref{fig:Distributions of GPT-4o-mini wpoh a} and \ref{fig:Distributions of GPT-4o-mini wpoh b}, and Figures  \ref{fig:Distributions of Mistral-8B-Instruct-2410 wpoh a} and \ref{fig:Distributions of Mistral-8B-Instruct-2410 wpoh b}, the distributions of the Mann–Kendall ($S$) and Spearman's rank correlation ($\rho$) statistics for both models show no clear trend. The values are broadly spread and centered around 0, indicating that opinions vary across agents and do not follow a consistent directional pattern.
    \item \textbf{Rating Concentration:} Figures \ref{fig:Distributions of GPT-4o-mini wpoh c} and \ref{fig:Distributions of GPT-4o-mini wpoh d}, and  \ref{fig:Distributions of Mistral-8B-Instruct-2410 wpoh c} and \ref{fig:Distributions of Mistral-8B-Instruct-2410 wpoh d} show that the concentration metrics support the maintenance of opinion diversity. For both models, the Kurtosis distribution is centered near 0 and includes negative values, while the IQR distributions are dispersed across a range of values, indicating that opinions remain heterogeneous.
\end{itemize}

\paragraph{Scenario IV (No History, No Persona): Revealing inherent biases under baseline, namely, when history and persona signals are not present.}
This scenario excludes both persona and history signals 
% (w/o History \& w/o Persona)
, serving as a baseline to capture the models' intrinsic tendencies. In this setting, the distributions for GPT-4o-mini and Mistral-8B-Instruct, as shown in Figure \ref{fig:Distributions of GPT-4o-mini opoh} and Figure \ref{fig:Distributions of Mistral-8B-Instruct-2410 opoh}, reveal an inherent bias.

\begin{itemize}[noitemsep, topsep=0pt, leftmargin=*]
    \item \textbf{Trend Metrics:} Both models display static and highly uniform trends. As shown in Figures~\ref{fig:Distributions of GPT-4o-mini opoh a} and~\ref{fig:Distributions of GPT-4o-mini opoh wpoh b}, and Figures~\ref{fig:Distributions of Mistral-8B-Instruct-2410 opoh a} and~\ref{fig:Distributions of Mistral-8B-Instruct-2410 opoh b}, the Spearman's rank correlation ($\rho$) is sharply peaked at 1.0, while the Mann–Kendall ($S$) values are narrowly centered around 0. 
    This suggests that the model defaults to a consistent opinion regardless of the movie, reflecting an internal bias rather than a context-driven response.
    \item \textbf{Concentration Metrics:} As shown in Figures~\ref{fig:Distributions of GPT-4o-mini opoh c} and~\ref{fig:Distributions of GPT-4o-mini opoh d}, and Figures~\ref{fig:Distributions of Mistral-8B-Instruct-2410 opoh c} and~\ref{fig:Distributions of Mistral-8B-Instruct-2410 opoh d}, both models display extreme rating concentration. The interquartile range (IQR) values are sharply peaked at 0, and the kurtosis values are strongly positive, indicating a highly uniform and narrowly distributed set of ratings across agents.
\end{itemize}

% \begin{figure*}[t]
% \centering
% \includegraphics[width=\linewidth]{figure/gpt-4o-mini/gpt-4o-mini_ratings_no_persona_no_history_distribution.pdf}
% \caption{Distributions of Metrics for All Movie Rating Sequences on GPT-4o-mini with ``w/o History \& w/o Persona''.}
% \label{fig:Distributions of GPT-4o-mini opoh}
% \end{figure*}

\subsection{Case Study} 

%\zj{
This section visualizes the evolution of positive and negative opinion proportions over time for a single movie under each experimental condition. We focus on GPT-4o-mini, which serves as a representative example consistent with the patterns observed in other models.
%}
% \mz{To complement the analysis, we present typical cases from GPT-4o-mini's simulations to visualize the opinion dynamics under each scenario. For brevity, we focus on GPT-4o-mini as it serves as a representative example observed in the other models.}

\begin{figure}[H]
\centering
\includegraphics[width=0.75\linewidth]{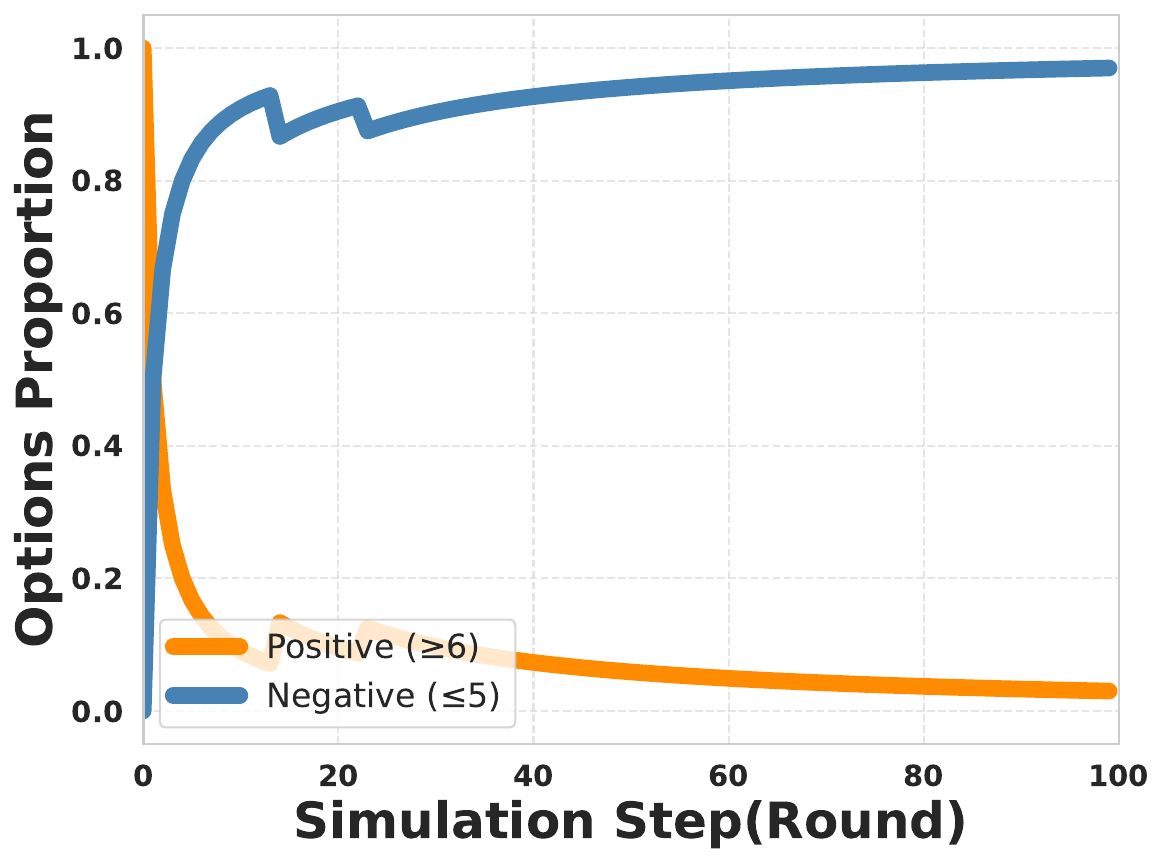}
\caption{A case study of the movie ``Norma'' with GPT-4o-mini in Scenario I (History + Persona).}
\label{fig:A Case of GPT-4o-mini wpwh}
\end{figure}

\paragraph{Scenario I.}%: Opinion convergence under full condition.} 
%\mz{
A typical case in the history + persona 
% (w/ History \& w/ Persona)
scenario is shown in Figure \ref{fig:A Case of GPT-4o-mini wpwh}. Initially, opinions are diverse due to the different personas. However, once one side gains a slight advantage in the early rounds, its dominance rapidly strengthens, while the minority opinion is quickly suppressed and ultimately silenced. This perfectly replicates the self-reinforcing dynamic central to SoS theory.
%}

\begin{figure}[H]
    \centering
    \begin{subfigure}{0.75\linewidth}
        \centering
        \includegraphics[width=\linewidth]{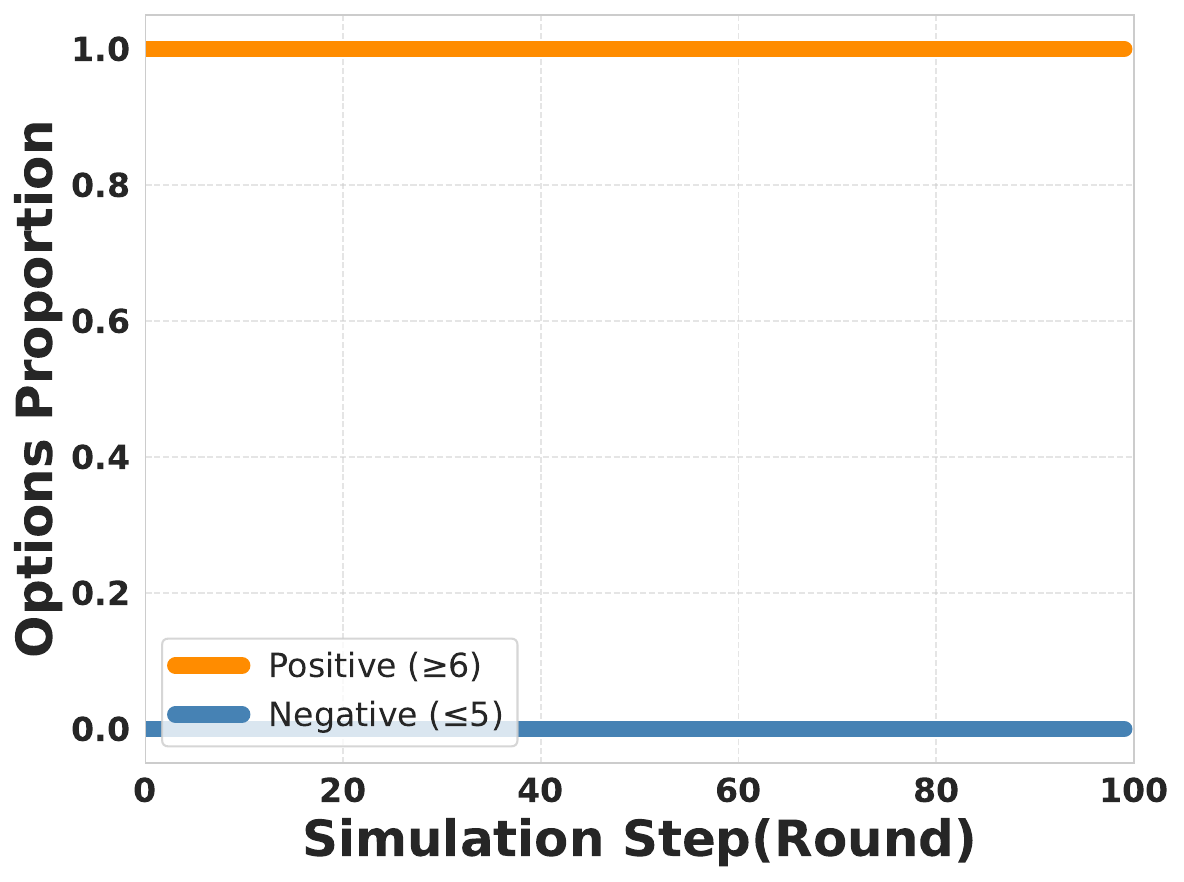}
        \caption{A Positive Case.}
        \label{fig:cases of GPT-4o-mini opwh positive}
    \end{subfigure}
    \vspace{1cm} 
    \begin{subfigure}{0.75\linewidth} 
        \centering
        \includegraphics[width=\linewidth]{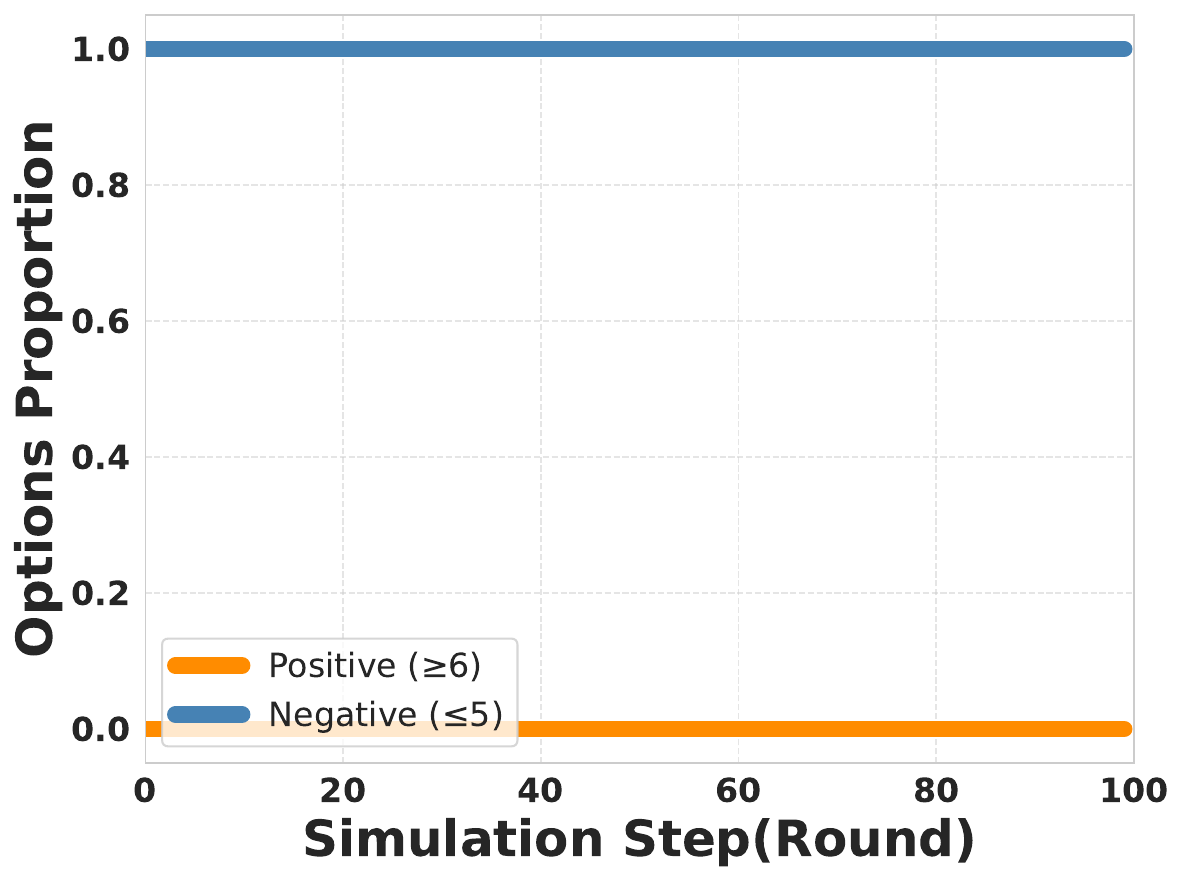}
        \caption{A Negative Case.}
        \label{fig:cases of GPT-4o-mini opwh negative}
    \end{subfigure}
    
    \vspace{-1cm}
    \caption{A positive case study of the
    movie ``Azzad'' and a negative case study of the movie ``Norma'' with GPT-4o-mini in Scenario II (History Only).}
    \label{fig:cases of GPT-4o-mini opwh}
\end{figure}

\paragraph{Scenario II.}%: Anchoring effect under history condition.}
%\mz{
When only history signal is present, we observe a strong anchoring effect rather than a self-reinforcing spiral. As shown in Figure \ref{fig:cases of GPT-4o-mini opwh positive} and Figure \ref{fig:cases of GPT-4o-mini opwh negative}, the collective opinion is randomly dominated by either positive or negative opinions at the start and then remains almost completely stable for the entire process. The opinion proportion remains unchanged over time, showing no signs of dynamic evolution.
%}

\vspace{-0.2cm} 

% \begin{figure}[H]
%     \centering
%     \begin{subfigure}{0.48\linewidth}
%         \centering
%         \includegraphics[width=\linewidth]{figure/gpt-4o-mini/gpt-4o-mini_ratings_positive_case.pdf}
%         \caption{A Positive Case.}
%         \label{fig:cases of GPT-4o-mini opwh positive}
%     \end{subfigure}%
%     \begin{subfigure}{0.48\linewidth}
%         \centering
%         \includegraphics[width=\linewidth]{figure/gpt-4o-mini/gpt-4o-mini_ratings_negative_case.pdf}
%         \caption{A Negative Case.}
%         \label{fig:cases of GPT-4o-mini opwh negative}
%     \end{subfigure}
%     \caption{A positive case study of the
%     movie ``Azzad'' and a negative case study of the movie ``Norma'' with GPT-4o-mini in Scenario II (History Only).}
%     \label{fig:cases of GPT-4o-mini opwh}
% \end{figure}

\begin{figure}[H]
\centering
\includegraphics[width=0.75\linewidth]{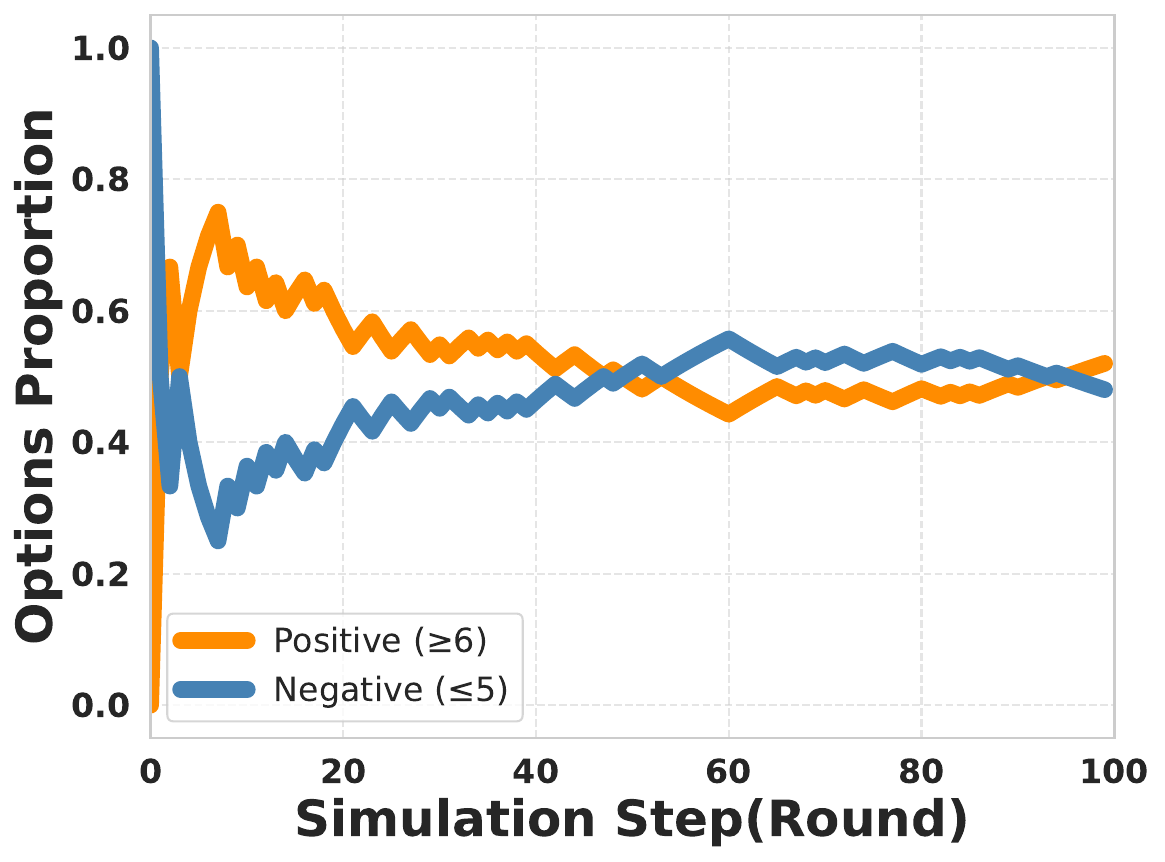}
\caption{A case study of the movie ``Norma'' with GPT-4o-mini in Scenario III (Persona Only).}
\label{fig:A Case of GPT-4o-mini with wpoh}
\end{figure}

\paragraph{Scenario III.}%: Opinion fluctuation under persona condition.}
%\mz{
When only persona signal is present 
% (w/o History \& w/ Persona)
, the opinion distribution exhibits a distinct dynamic. As shown in Figure \ref{fig:A Case of GPT-4o-mini with wpoh} positive and negative opinions fluctuate and compete throughout the rating process, with neither side achieving a lasting advantage.
%}

\begin{figure}[H]
\centering
\includegraphics[width=0.75\linewidth]{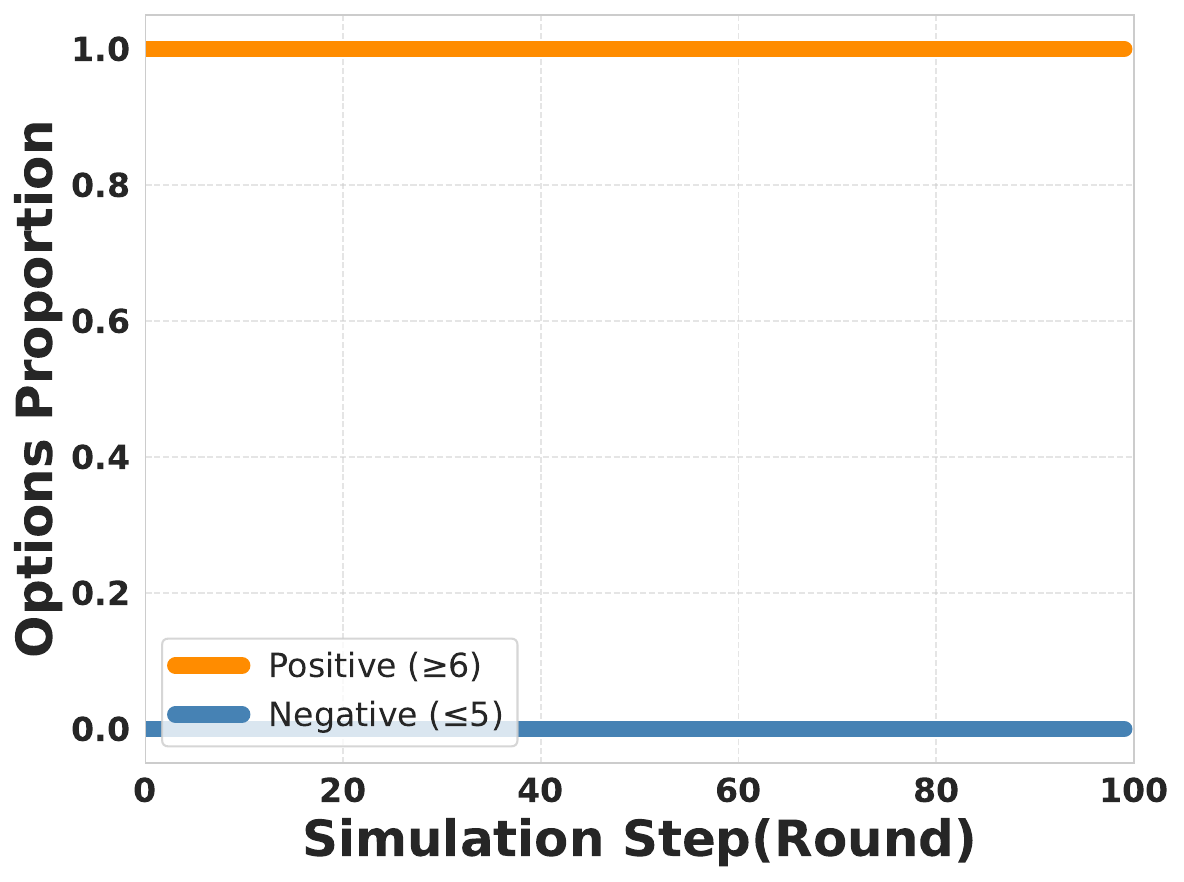}
\caption{A case study of the movie ``Azzad'' with GPT-4o-mini in Scenario IV (No History, No Persona).}
\label{fig:Positive Cases of GPT-4o-mini wpoh}
\end{figure}

\paragraph{Scenario IV.}%: Opinion fixation under baseline condition.}
%\mz{
In this scenario, when neither history nor persona signals are present, the model reveals its inherent bias. 
% This bias likely stems from the emotional tone of its training data. If the training corpora are generally positive in emotional expression, the model will more likely adopt this positivity as its default response. 
As shown in Figure~\ref{fig:Positive Cases of GPT-4o-mini wpoh}, the proportion of positive opinions remains fixed at 1.0 from the very beginning, with negative opinions entirely absent throughout the process. This static pattern is not the result of opinion dynamics but rather reflects a built-in positivity prior, reflecting the model’s default tendency to favor positive ratings when no contextual signals are available. This positivity bias aligns with prior LLM-as-a-judge evidence showing leniency toward higher scores and sycophancy effects that inflate positive ratings.
%}

% \begin{figure}[H]
%     \centering
%     \begin{minipage}[t]{0.48\linewidth}
%      \vspace{0pt}
%         \centering
%         \includegraphics[width=\linewidth]{figure/gpt-4o-mini/gpt-4o-mini_ratings_no_history_case.pdf}
%         \captionof{figure}{A case study of the movie ``Norma'' with GPT-4o-mini in Scenario III (Persona Only).}
%         \label{fig:A Case of GPT-4o-mini with wpoh}
%     \end{minipage}%
%     \hfill 
%     \begin{minipage}[t]{0.48\linewidth}
%      \vspace{0pt}
%         \centering
%         \includegraphics[width=\linewidth]{figure/gpt-4o-mini/gpt-4o-mini_ratings_positive_case.pdf}
%         \captionof{figure}{A case study of the movie ``Azzad'' with GPT-4o-mini in Scenario IV (No History, No Persona).}
%         \label{fig:Positive Cases of GPT-4o-mini wpoh}
%     \end{minipage}
% \end{figure}

% \begin{figure}[h]
%   \centering
%     \centering
%     \includegraphics[width=\linewidth]{figure/gpt-4o-mini/gpt-4o-mini_ratings_positive_case.pdf}
% \caption{A Positive Case of GPT-4o-mini with ``w/o History \& w/ Persona''.}
%   \label{fig:Positive Cases of GPT-4o-mini wpoh}
% \end{figure}

\subsection{Persona-Context Consistency}

We further explore whether an agent is more likely to conform to the collective opinion when its assigned persona aligns closely with the movie’s content.
To quantify this alignment, we introduce semantic match score and rating distance. The semantic match score quantifies the similarity between an agent's persona description and the movie's overview using TF-IDF–based cosine similarity. The rating distance is defined as the absolute difference between an agent’s individual rating and the historical average of all prior ratings, capturing the degree of deviation from collective opinion:

\begin{equation}
    \label{eq:rating distance}
    \operatorname{Dist}(r_{i,j,k}) \;=\; \bigl|\,r_{i,j,k} - \mathcal{F}(\mathcal{H}_{j,k})\bigr|.
\end{equation}

\begin{figure}[h]
  \centering
  \includegraphics[width=\linewidth]{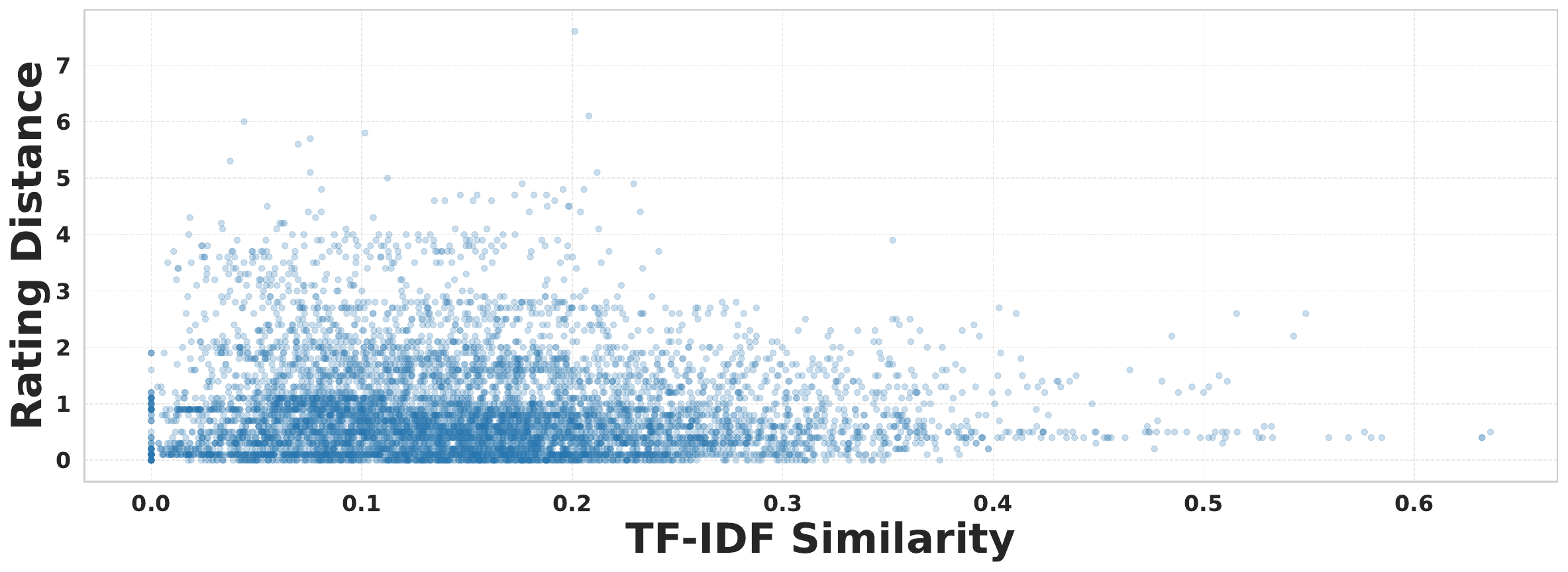}
  \caption{The relation between \textbf{Semantic Match vs. Rating Distance}.}
  \label{fig:Semantic Similarity}
\end{figure}

The results, as shown in Figure~\ref{fig:Semantic Similarity}, reveal a clear negative correlation between semantic match score and rating distance. When the alignment between an agent’s persona and the movie overview is low, ratings are more widely dispersed and frequently deviate from the collective opinion. In contrast, high semantic similarity corresponds to tightly clustered ratings with minimal variance. These findings suggest that consistency between persona and movie context plays a crucial role in shaping conformity. Agents whose personas are more closely aligned with the content are more likely to follow collective opinion. This pattern likely reflects greater confidence in its own assessment, resulting in more stable and less variable ratings.

% \mz{To further validate the effectiveness of the persona mechanism, we analyze the relationship between semantic match and rating distance. We intentionally selected TF-IDF over the model's own embeddings to use an external, interpretable metric, and avoid circular reasoning.}

% \mz{The results, as shown in Figure \ref{fig:Semantic Similarity}, reveal a clear negative correlation between them. When the semantic similarity between the persona and the movie is low, the distribution of rating distances is highly dispersed, with a large number of "dissenting" ratings that deviate from the collective opinion. However, when similarity is high, large rating distances become extremely rare and ratings cluster tightly around the collective opinion. This finding indicates that persona-context consistency is a key factor that regulates an agent's tendency to conform. When an agent's "identity" is highly relevant to the task, it is more inclined to follow collective norms; conversely, it is more likely to express dissenting opinions.}

\section{Related Work}

%\mz{
The SoS theory, introduced by \citet{noelle1974spiral}, posits that individuals suppress minority opinions due to fear of social isolation. Subsequent work has tested these ideas online. For example, a Pew survey \citep{hampton2014social} shows that social media users are much more likely to voice opinions when they believe that their network agrees with them, and \citet{porten2015spiral} shows that anonymity and low-effort feedback significantly increase willingness to express unpopular views on online forums. More recently, researchers have explored how LLM agents can simulate such social dynamics \citep{chuang2023simulating}. \citet{park2023generative} use generative agents in a simulated town; these agents exhibit emergent social behaviors. Similarly, \citet{nasim2025simulating} presents Gensim, a general social simulation platform with LLM agents, and \citet{light2023avalonbench, shi2023cooperation} studies a community of LLMs playing the social deduction game Avalon. \citet{akata2025playing} use behavioral game theory to let LLM agents play finitely repeated games, finding that models develop consistent cooperative or defection strategies. \citet{sarkadi2019modelling}  introduces the Traitors framework for LLMs to study trust and deceit. \citet{leng2023llm} analyzes LLM responses in canonical economics games using a probabilistic SUVA framework and reports that the decisions of most models reflect social welfare and reciprocity considerations rather than pure self-interest. 
%}

Other recent work focuses on how LLMs represent the majority opinions \citep{weng2025we}. For example, \citet{ye2024justice} systematically quantifies biases in LLM-as-a-Judge and identifies a strong bandwagon effect. To explore the dynamics of opinion, \citet{nasim2025simulating} proposes a simulator that embeds LLM-based agents into networked opinion spread models. By integrating classic theories of social influence \citep{kelman1958compliance, munroe2013social} with LLM communication, their framework allows researchers to study how LLM agents propagate influence. Likewise, \citep{yang2024oasis} presents OASIS, an open-scale social media simulator with up to a million LLM agents, and shows that larger simulated populations produce richer group dynamics and greater opinion diversity, and \citet{zhao2024electoral} shows the diversity of LLM agents. These LLM-based platforms connect directly to prior work on collective behavior: classical agent-based models by \citep{deffuant2002can, rainer2002opinion, friedkin2011social} demonstrate how repeated local interactions can produce global consensus or polarization.

\section{Conclusion}

%\mz{
Our study shows that LLM-based movie rating agents exhibit a clear positivity bias by default, yet develop more diverse opinions when given distinct personas and increasingly conform to prior context when a historical collective opinion is provided. By crossing binary signals design (persona × history), we isolate the influence of each signal: persona alone induces opinion diversity, history alone imposes anchoring consistency, and only their combination triggers a pronounced SoS. These results highlight that a SoS can spontaneously emerge in LLM agents without any emotional drive: purely from the interplay between internalized statistical biases and externally presented collective signals. This insight underscores the power of social context in shaping AI behavior and reminds us to remain alert to the social biases embedded in LLMs that can influence such simulations.
%}

% !TEX root = acl_latex.tex
\section{Limitations and Potential Risks}

%\mz{
\paragraph{Limitations.} Our study is subject to several practical constraints. First, due to available computational resources, our experiments focus on lightweight and midsized open-source models, rather than very large-scale models, which may exhibit different emergent dynamics. Second, our simulation of social feedback adopts a simplified agent, that is, providing agents only with the historical average rating as a substitute for social influence. Although this abstraction enables controlled investigation of majority dynamics, it does not capture the full range of factors shaping opinion formation in real-world societies, such as emotion, network structure, or identity effects. However, we believe that these design choices allow us to isolate and systematically analyze the core mechanisms of the emergence of the SoS in collectives of LLM-based agents, and we leave more complex extensions for future work.
%}

\paragraph{Potential Risks.} Our study shows that purely algorithmic LLM agents can reproduce SoS effect. Although this advances scientific understanding, it also entails several risks: Malicious actors could adapt our protocol to build large-scale manipulative campaigns or persuasive LLM-based chatbots that systematically nudge users toward the perceived majority, thus suppressing dissenting voices; If the initial prompt or training data carry demographic, political, or cultural biases, the SoS mechanism may magnify those biases and further marginalize minority opinions.

\paragraph{Licenses.} All models and tools used in this study are released under open-source or research licenses.

% !TEX root = acl_latex.tex
\section{Acknowledgements}

We thank the anonymous reviewers for their insightful comments and constructive feedback.
%, which greatly improved the quality of this paper. 
%We also acknowledge the support of our colleagues and collaborators for valuable discussions during the development of this work.

\bibliography{reference}

% !TEX root = acl_latex.tex
\newpage
\onecolumn
\appendix

\section{Appendix}
\addcontentsline{toc}{section}{Appendix}
\label{sec:appendix}

% \subsection{A Persona Example}

% \begin{figure}[h]
% \centering
% \begin{quote}
% \itshape
% A computer enthusiast who is interested in optimizing the performance of their system, particularly the CPU, GPU, and RAM. They are looking for software tools that can help them monitor and control the performance of their system, and they are willing to invest time in learning how to use these tools effectively. They are not necessarily looking for a professional-grade software tool, but rather a user-friendly and easy-to-use software that can provide comprehensive information about their system's performance and help them optimize it. They are also interested in software tools that can help them monitor the stability of their system after overclocking, as they want to avoid damaging their system.
% \end{quote}
% \caption{A Persona Example.}
% \label{fig:persona}
% \end{figure}

\subsection{LLM Agent Rating Prompts}
\label{sec:prompt}

We present the four distinct prompts employed in our experiments. Each prompt corresponds to one of the four experimental settings, differing by the presence or absence of historical average rating information, referred to as \textit{History}, and character profile information, referred to as \textit{Persona}. These prompts were designed to isolate and evaluate the specific contributions of social influence and persona-based individual differences to the emergence of the SoS effect.

\begin{tcolorbox}[colback=gray!5!white,colframe=black!75!white,title=Prompt: History + Persona]
\small
\begin{verbatim}
Please provide your rating for the movie.

# Your Character Profile
You are [persona]

# Movie Information
Title: [Movie Title]
Genres: [Genres]
Overview: [Movie Overview]
Movie average rating: [Historical Average] (1-10)

# Rating Principle
Rate the above movie on an integer scale from 1 to 10, where:

- 1 = Awful/Abysmal (unwatchable)
- 5 = Mediocre/Unsure (forgettable)
- 10 = Perfect/Masterpiece (flawless)

# Output Principle
Provide only a single integer (1-10) without extra text.
\end{verbatim}
\end{tcolorbox}

\begin{tcolorbox}[colback=gray!5!white,colframe=black!75!white,title=Prompt: History only]
\small
\begin{verbatim}
Please provide your rating for the movie.

# Movie Information
Title: [Movie Title]
Genres: [Genres]
Overview: [Movie Overview]
Movie average rating: [Historical Average] (1-10)

# Rating Principle
Rate the above movie on an integer scale from 1 to 10, where:

- 1 = Awful/Abysmal (unwatchable)
- 5 = Mediocre/Unsure (forgettable)
- 10 = Perfect/Masterpiece (flawless)

# Output Principle
Provide only a single integer (1-10) without extra text.
\end{verbatim}
\end{tcolorbox}

\begin{tcolorbox}[colback=gray!5!white,colframe=black!75!white,title=Prompt: Persona only]
\small
\begin{verbatim}
Please provide your rating for the movie.

# Your Character Profile
You are [persona]

# Movie Information
Title: [Movie Title]
Genres: [Genres]
Overview: [Movie Overview]

# Rating Principle
Rate the above movie on an integer scale from 1 to 10, where:

- 1 = Awful/Abysmal (unwatchable)
- 5 = Mediocre/Unsure (forgettable)
- 10 = Perfect/Masterpiece (flawless)

# Output Principle
Provide only a single integer (1-10) without extra text.
\end{verbatim}
\end{tcolorbox}

\begin{tcolorbox}[colback=gray!5!white,colframe=black!75!white,title={Prompt: No History, No Persona}]
\small
\begin{verbatim}
Please provide your rating for the movie.

# Movie Information
Title: [Movie Title]
Genres: [Genres]
Overview: [Movie Overview]


# Rating Principle
Rate the above movie on an integer scale from 1 to 10, where:

- 1 = Awful/Abysmal (unwatchable)
- 5 = Mediocre/Unsure (forgettable)
- 10 = Perfect/Masterpiece (flawless)

# Output Principle
Provide only a single integer (1-10) without extra text.
\end{verbatim}
\end{tcolorbox}

\subsection{Persona Example}

\label{sec:persona}

\begin{figure}[h]
  \centering
\begin{quote}\ttfamily\small
A computer enthusiast who is interested in optimizing the performance of their system, particularly the CPU, GPU, and RAM. They are looking for software tools that can help them monitor and control the performance of their system, and they are willing to invest time in learning how to use these tools effectively. They are not necessarily looking for a professional-grade software tool, but rather a user-friendly and easy-to-use software that can provide comprehensive information about their system's performance and help them optimize it. They are also interested in software tools that can help them monitor the stability of their system after overclocking, as they want to avoid damaging their system.
  \end{quote}
  \caption{A persona example}
\end{figure}

\subsection{Metrics Distributions for Other Models}
\label{sec:appendix-other-model-metrics}

In this section, we present the distribution of statistical metrics for additional models including DeepSeek V2 Lite Chat and the Qwen 2.5 series (1.5B, 3B, and 7B) under each of the four experimental scenarios. For each model, we visualize and summarize four key statistics across all movies: the Mann–Kendall statistic ($S$), Spearman's rank correlation coefficient ($\rho$), late-stage interquartile range (IQR), and late-stage kurtosis.

\begin{figure*}[htbp]
  \centering
  \begin{subfigure}[b]{0.48\linewidth}
    \centering
    \includegraphics[width=\linewidth]{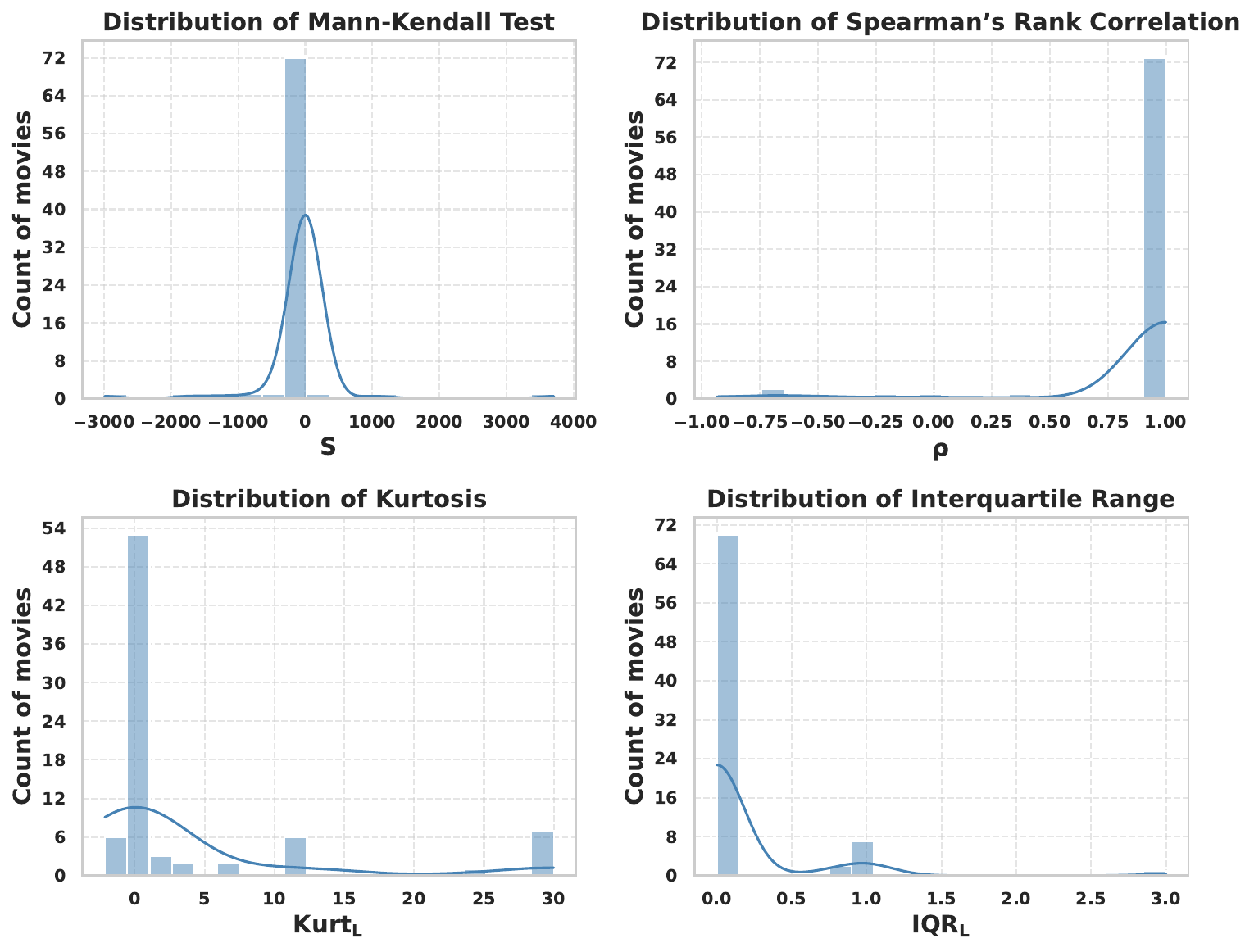}
    \caption{``w/o Persona \& w/o History''}
  \end{subfigure}
  \begin{subfigure}[b]{0.48\linewidth}
    \centering
    \includegraphics[width=\linewidth]{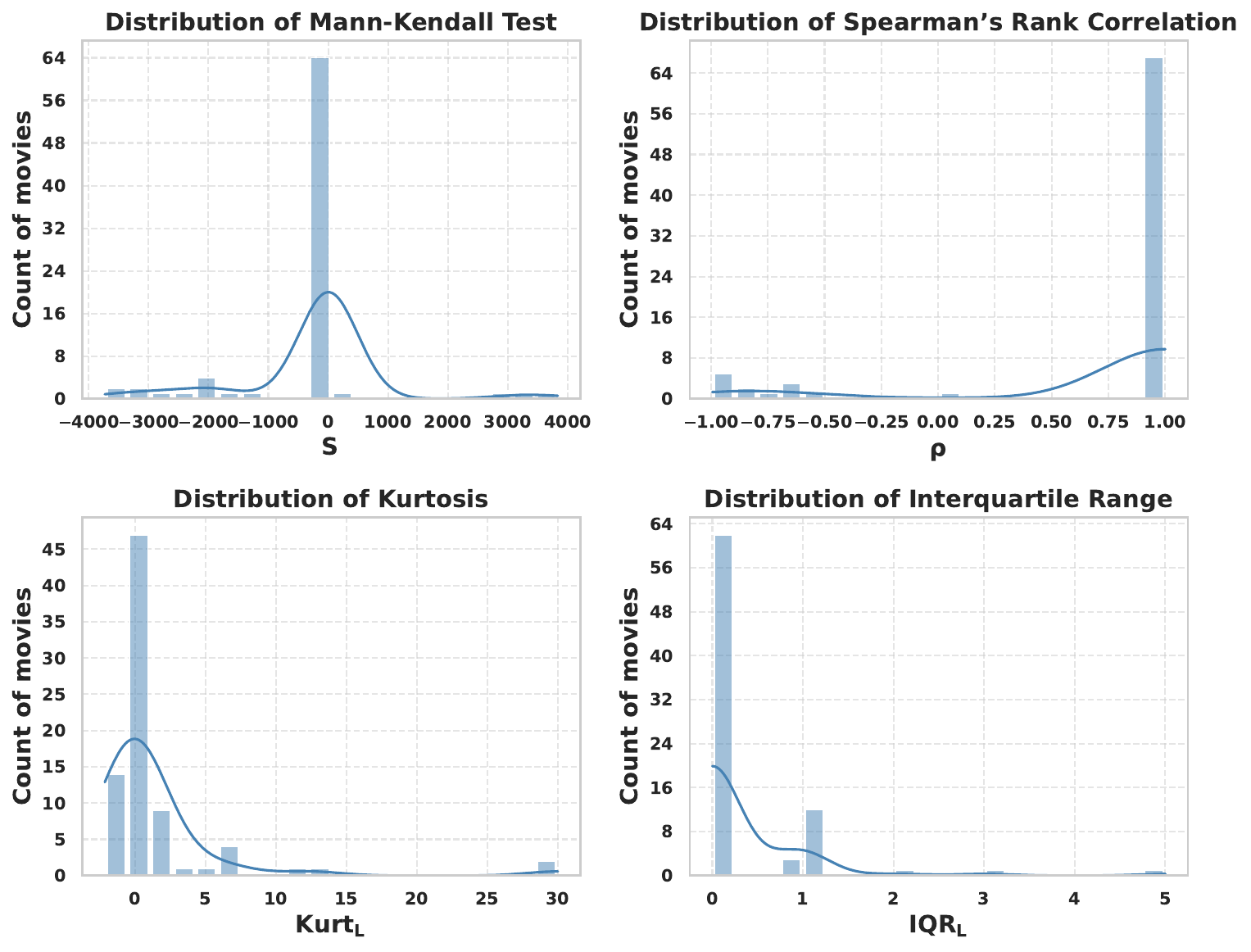}
    \caption{``w/o Persona \& w/ History''}
  \end{subfigure}\\
  \begin{subfigure}[b]{0.48\linewidth}
    \centering
   \includegraphics[width=\linewidth]{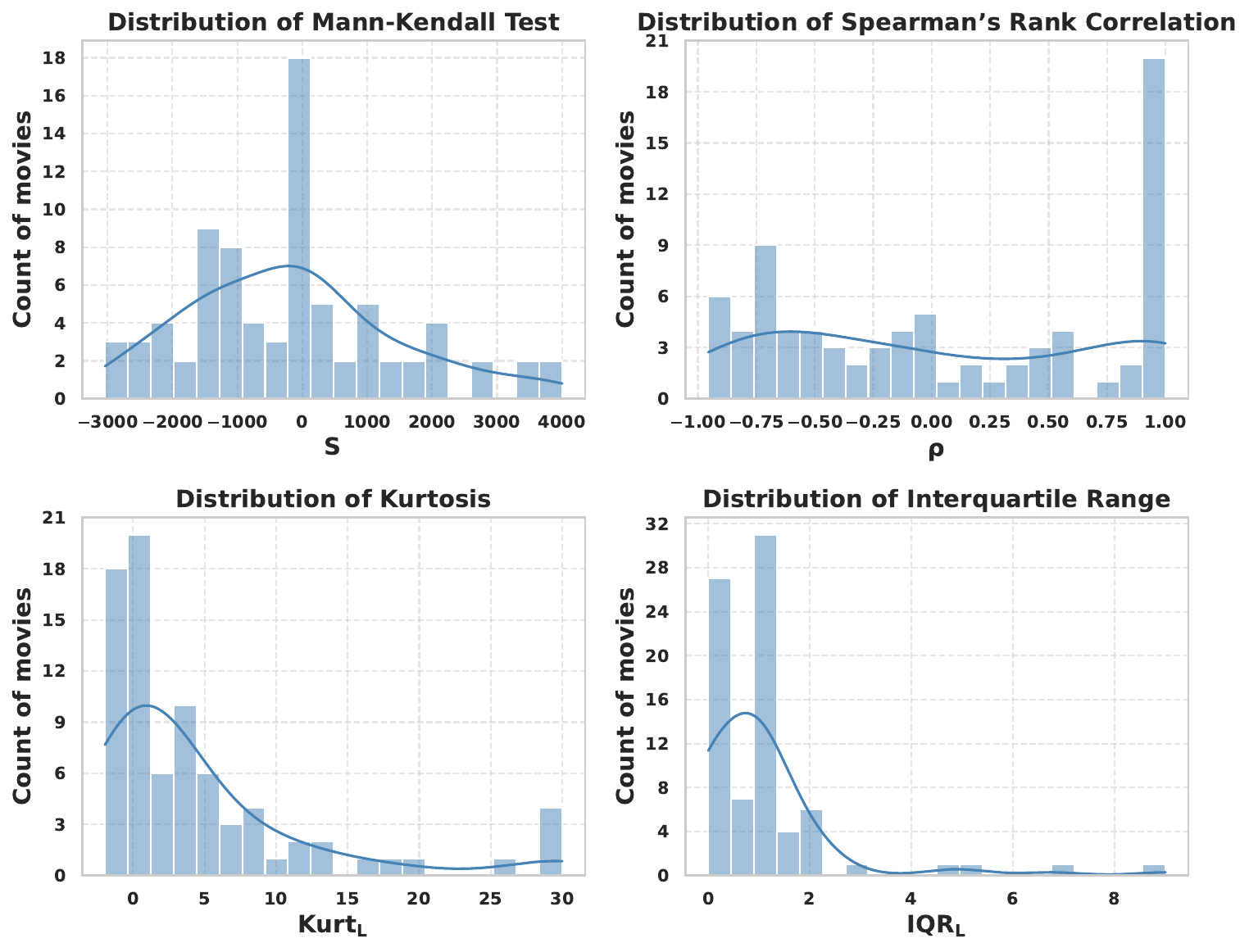}
    \caption{``w/ Persona \& w/o History''}
  \end{subfigure}
  \begin{subfigure}[b]{0.48\linewidth}
    \centering
    \includegraphics[width=\linewidth]{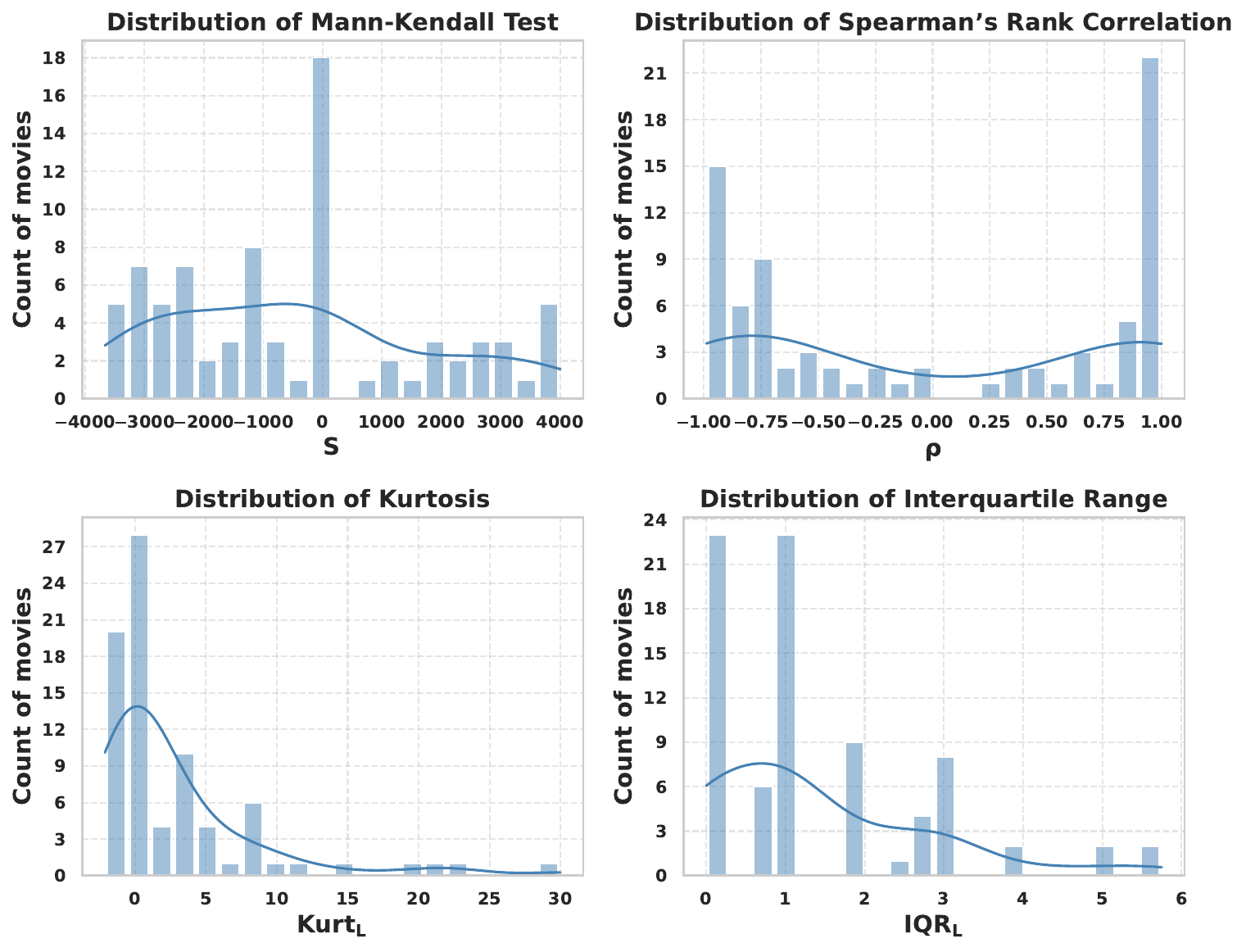}
    \caption{``w/ Persona \& w/ History''}
  \end{subfigure}
  \caption{Distributions of Mann–Kendall Statistic, Spearman Rank Correlation, Kurtosis, Inter-quartile Range for All Movie Rating Sequences on DeepSeek-V2-Lite-Chat.}
  \label{fig:Distributions of Mann–Kendall Statistic, Spearman Rank Correlation, Kurtosis, Inter-quartile Range for All Movie Rating Sequences on DeepSeek-V2-Lite-Chat}
\end{figure*}

\begin{figure*}[htbp]
  \centering
  \begin{subfigure}[b]{0.48\linewidth}
    \centering
    \includegraphics[width=\linewidth]{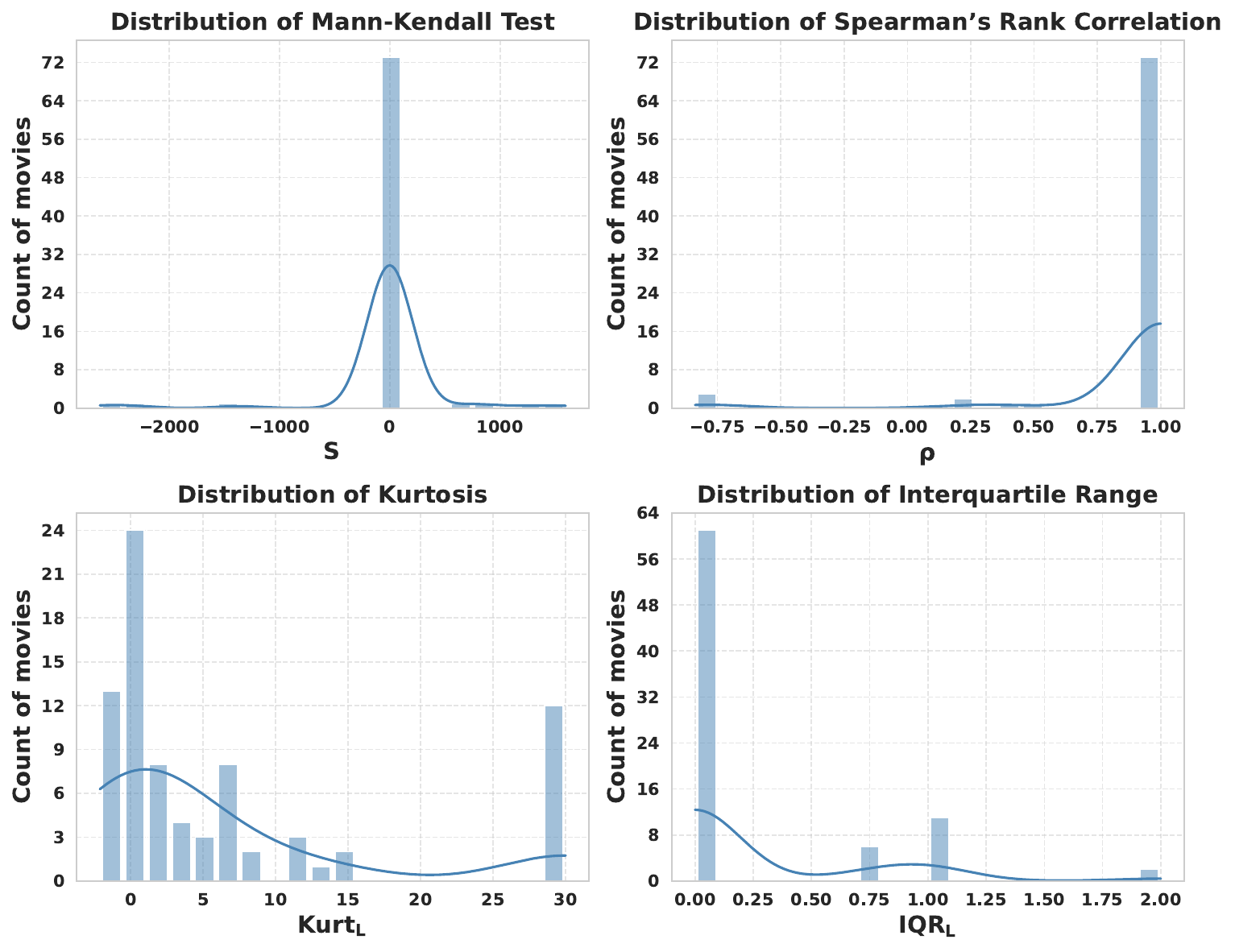}
    \caption{``w/o Persona \& w/o History''}
  \end{subfigure}
  \begin{subfigure}[b]{0.48\linewidth}
    \centering
    \includegraphics[width=\linewidth]{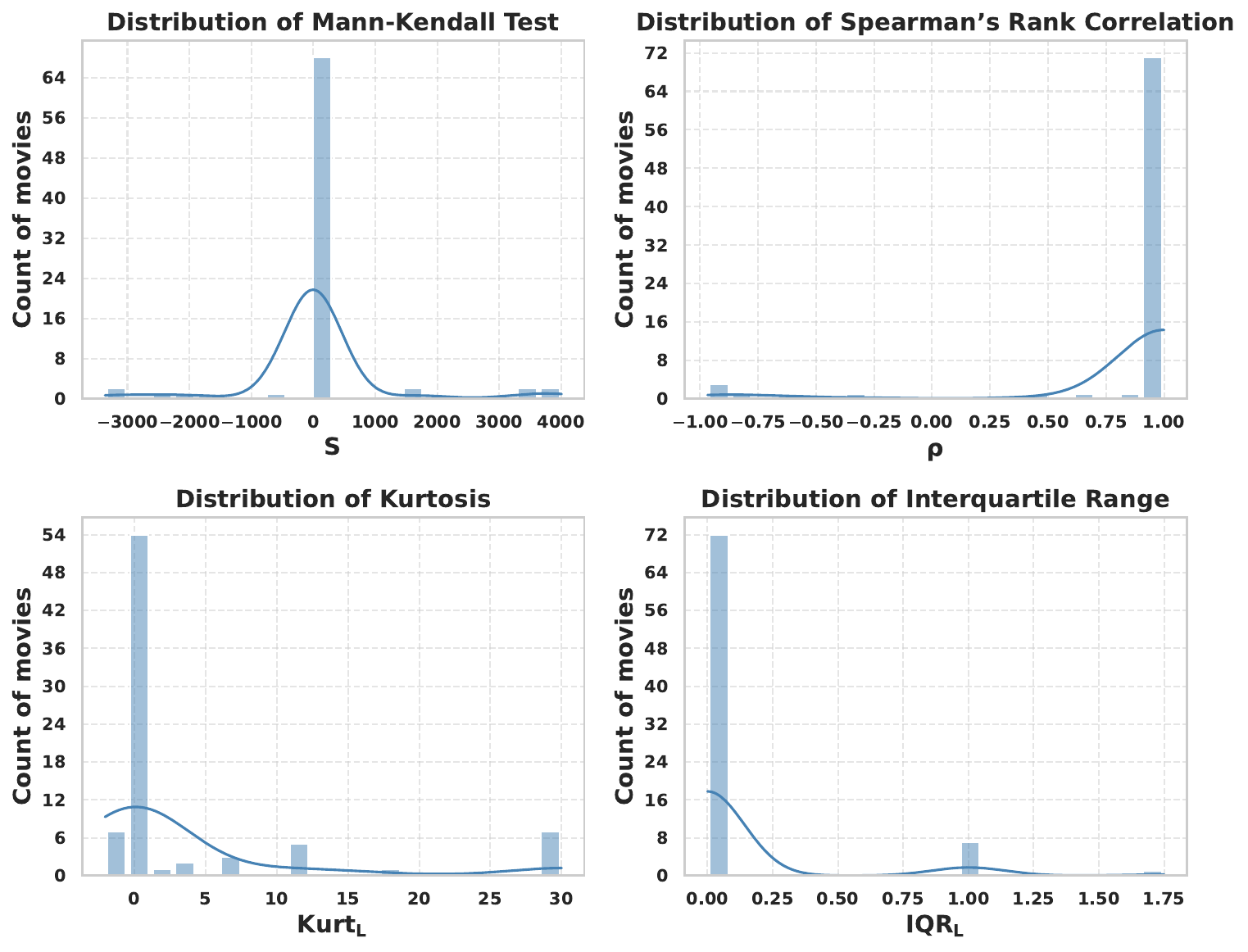}
    \caption{``w/o Persona \& w/ History''}
  \end{subfigure}\\
  \begin{subfigure}[b]{0.48\linewidth}
    \centering
   \includegraphics[width=\linewidth]{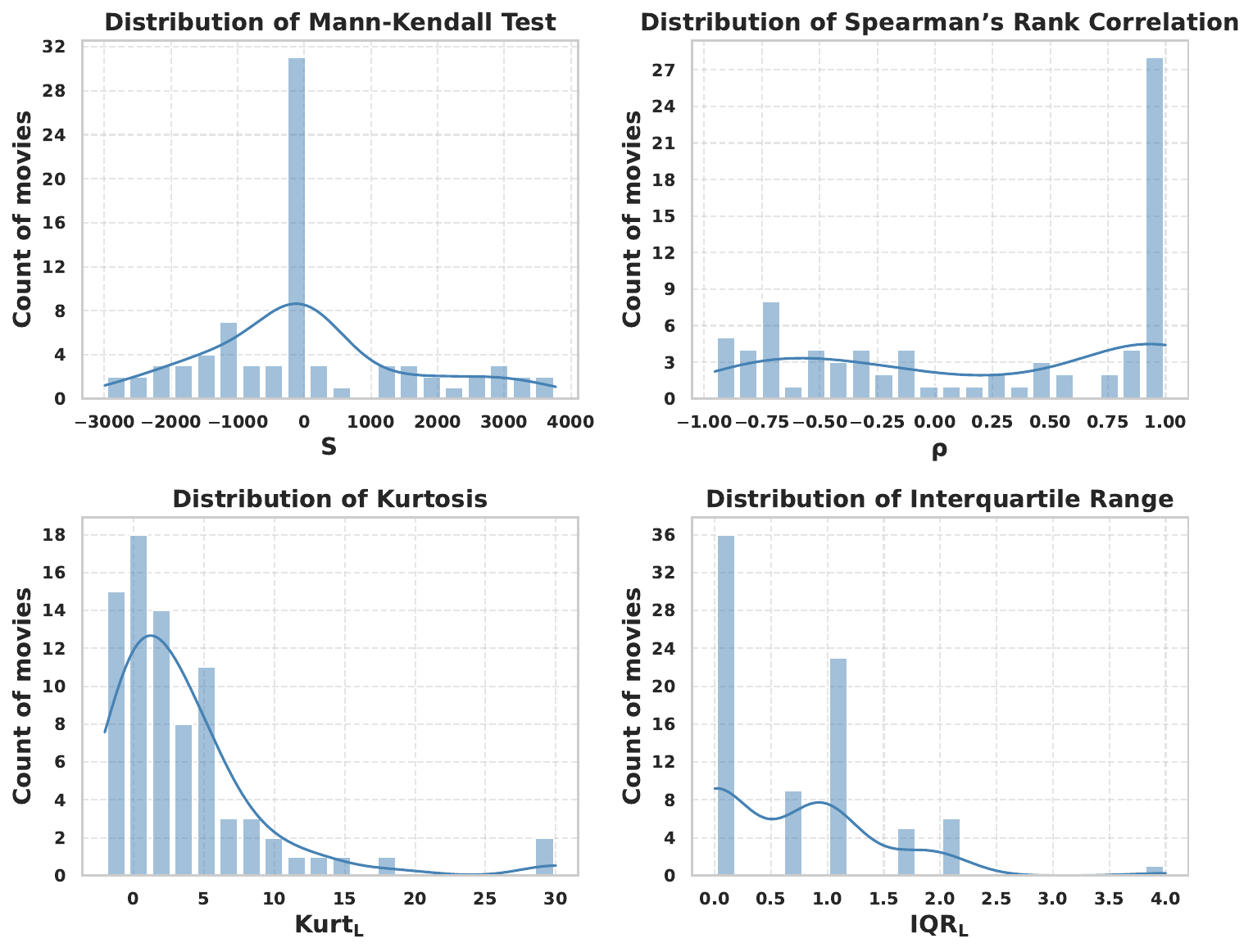}
    \caption{``w/ Persona \& w/o History''}
  \end{subfigure}
  \begin{subfigure}[b]{0.48\linewidth}
    \centering
    \includegraphics[width=\linewidth]{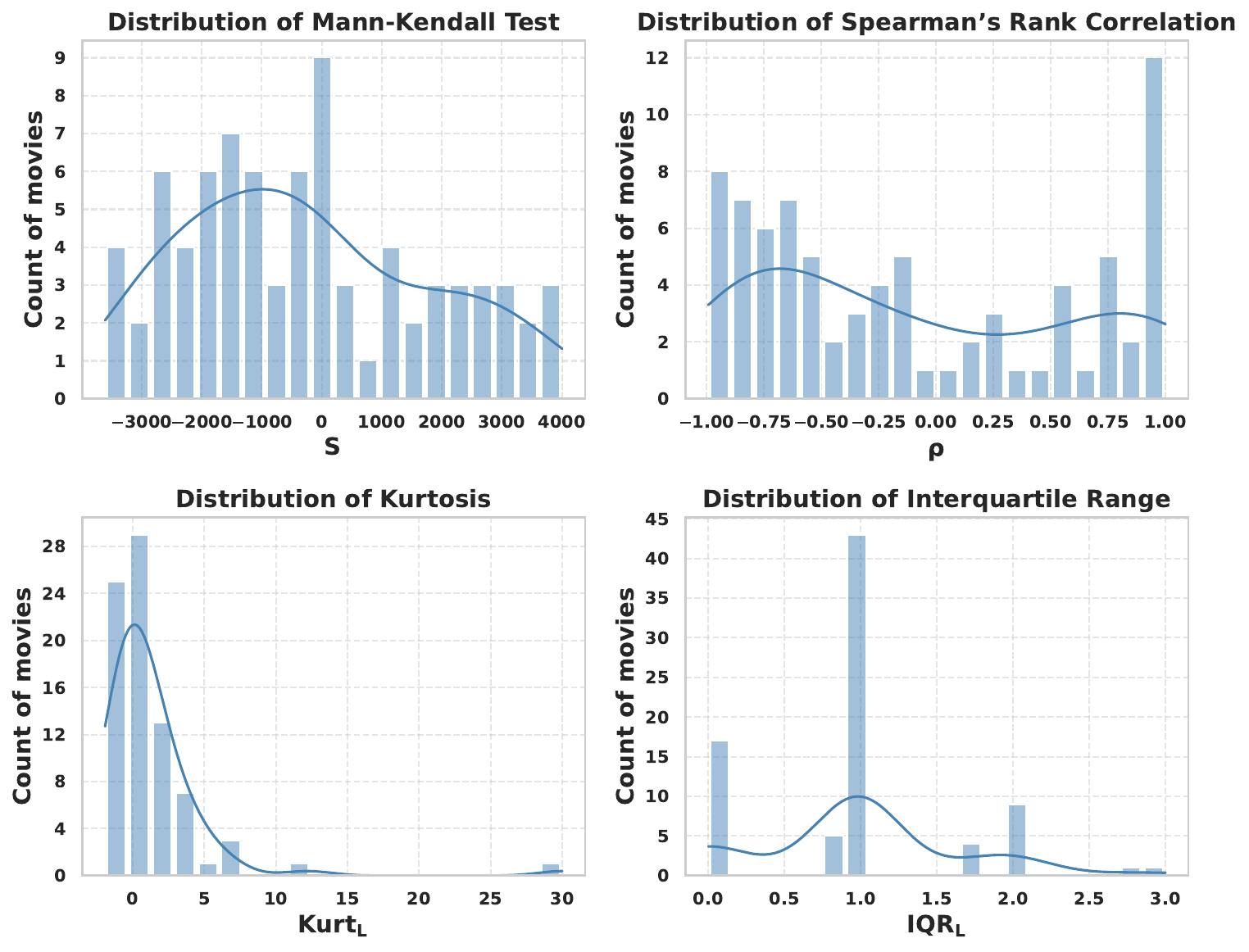}
    \caption{``w/ Persona \& w/ History''}
  \end{subfigure}
  \caption{Distributions of Mann–Kendall Statistic, Spearman Rank Correlation, Kurtosis, Inter-quartile Range for All Movie Rating Sequences on Qwen2.5-1.5B-Instruct.}
  \label{fig:Distributions of Mann–Kendall Statistic, Spearman Rank Correlation, Kurtosis, Inter-quartile Range for All Movie Rating Sequences on Qwen2.5-1.5B-Instruct}
\end{figure*}

\begin{figure*}[htbp]
  \centering
  \begin{subfigure}[b]{0.48\linewidth}
    \centering
    \includegraphics[width=\linewidth]{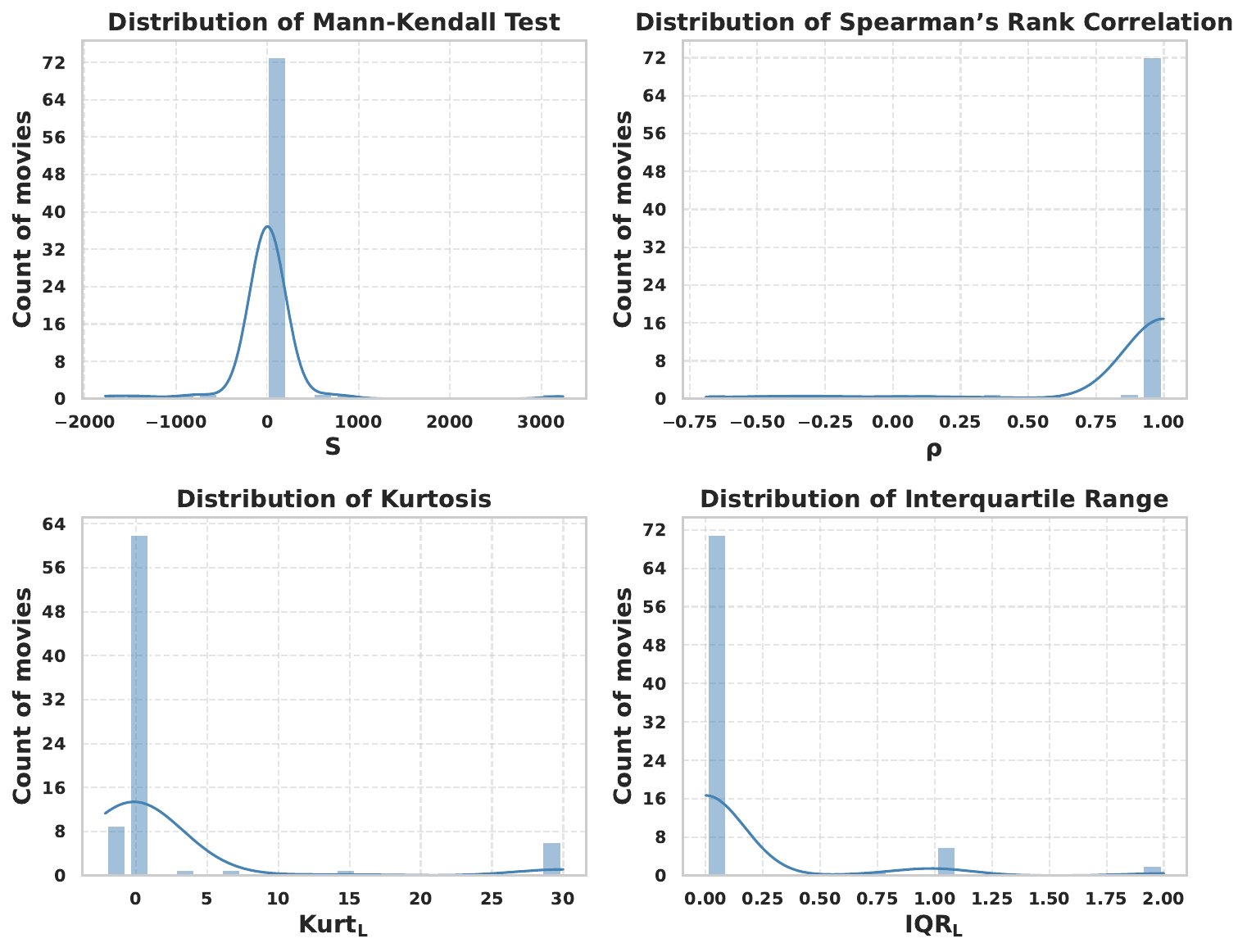}
    \caption{``w/o Persona \& w/o History''}
  \end{subfigure}
  \begin{subfigure}[b]{0.48\linewidth}
    \centering
    \includegraphics[width=\linewidth]{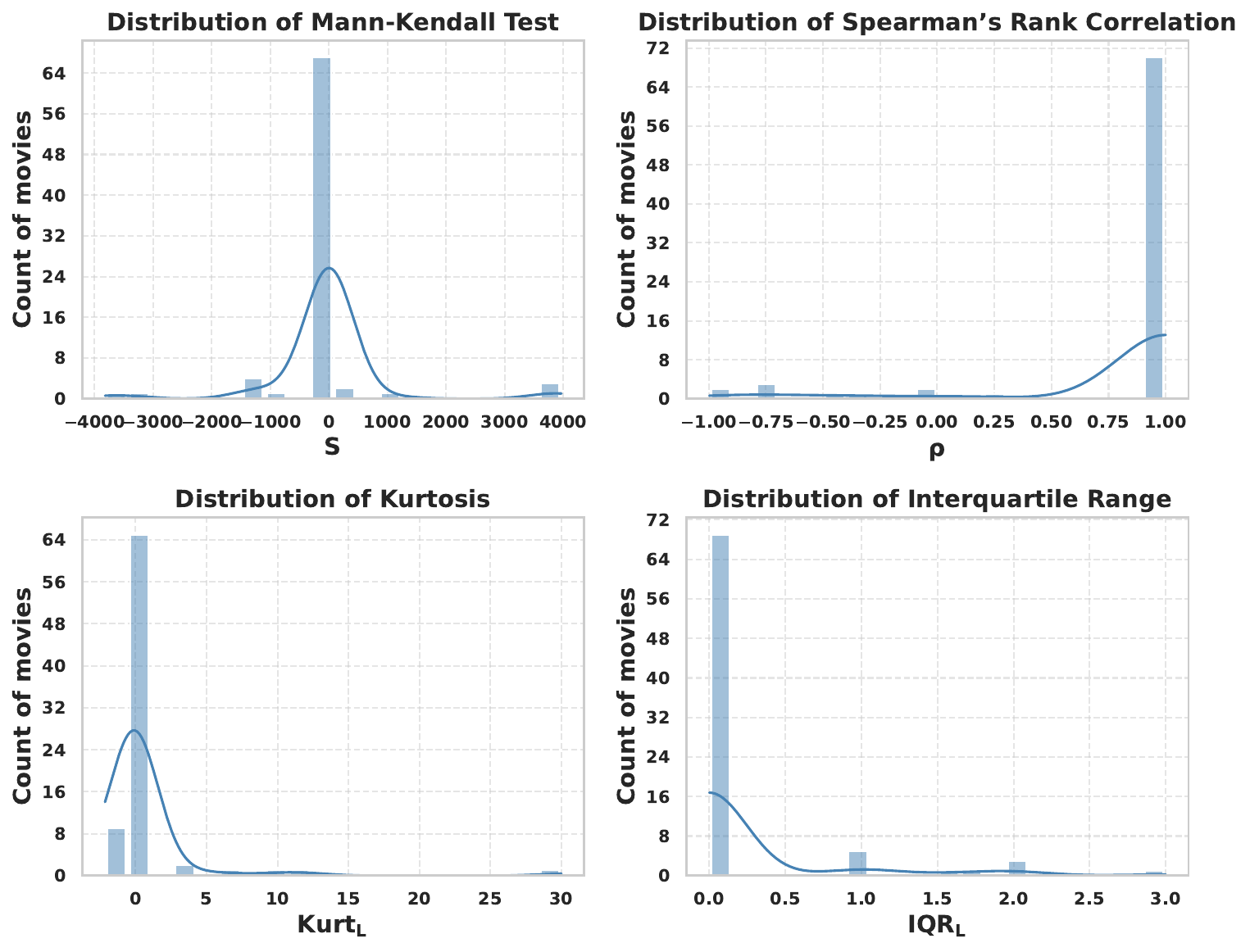}
    \caption{``w/o Persona \& w/ History''}
  \end{subfigure}\\
  \begin{subfigure}[b]{0.48\linewidth}
    \centering
   \includegraphics[width=\linewidth]{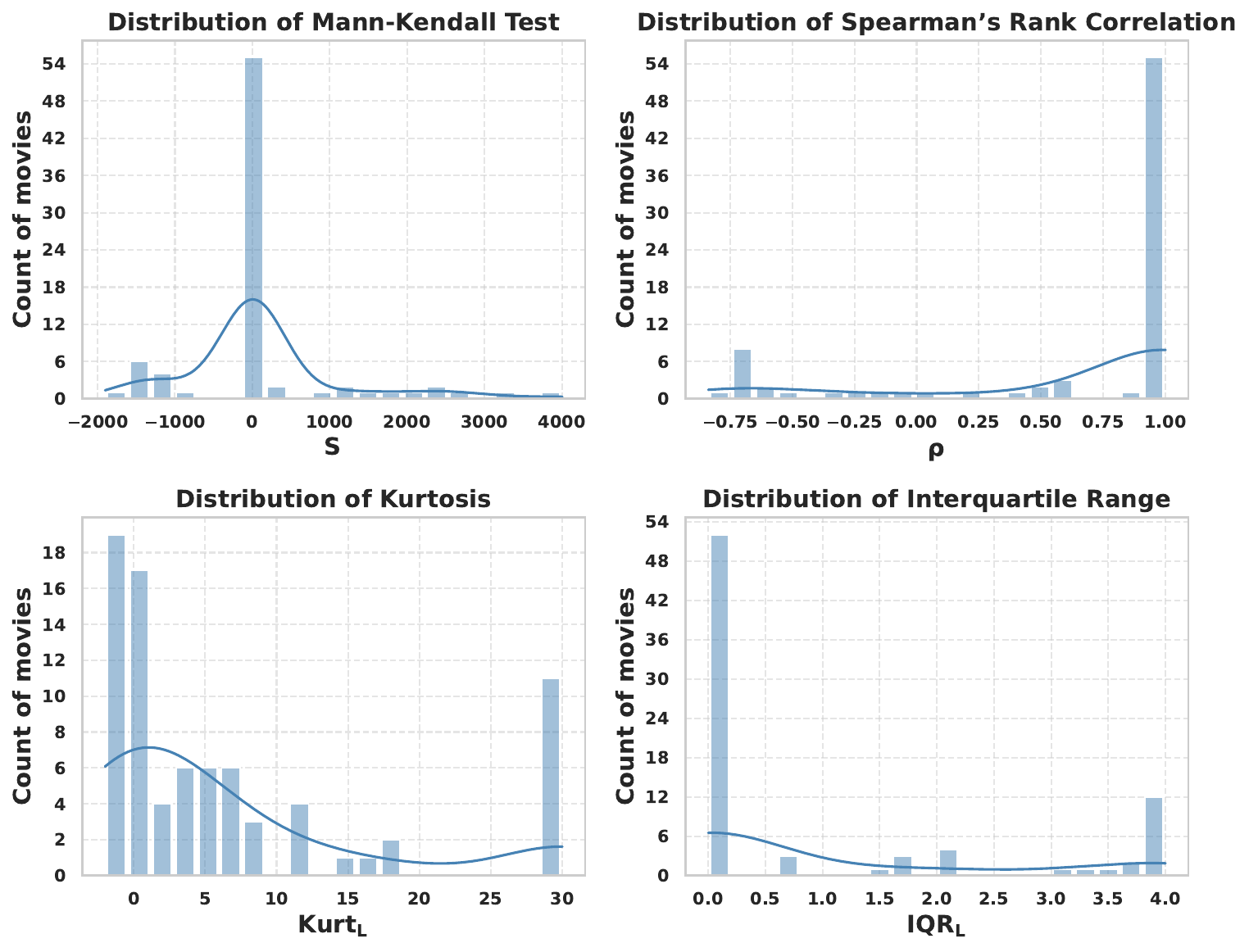}
    \caption{``w/ Persona \& w/o History''}
  \end{subfigure}
  \begin{subfigure}[b]{0.48\linewidth}
    \centering
    \includegraphics[width=\linewidth]{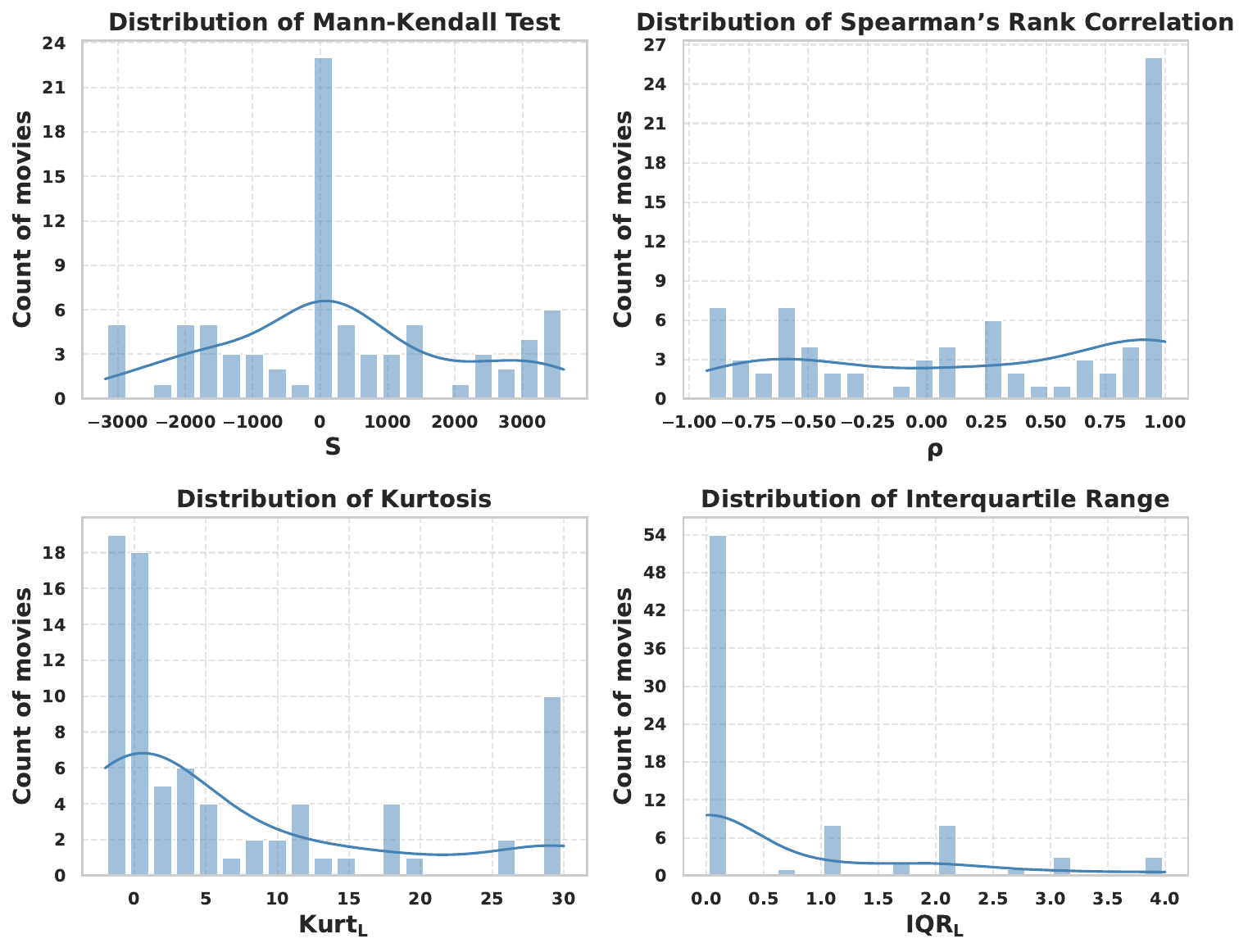}
    \caption{``w/ Persona \& w/ History''}
  \end{subfigure}
  \caption{Distributions of Mann–Kendall Statistic, Spearman Rank Correlation, Kurtosis, Inter-quartile Range for All Movie Rating Sequences on Qwen2.5-3B-Instruct.}
  \label{fig:Distributions of Mann–Kendall Statistic, Spearman Rank Correlation, Kurtosis, Inter-quartile Range for All Movie Rating Sequences on Qwen2.5-3B-Instruct}
\end{figure*}

\begin{figure*}[t]
  \centering
  \begin{subfigure}[b]{0.48\linewidth}
    \centering
    \includegraphics[width=\linewidth]{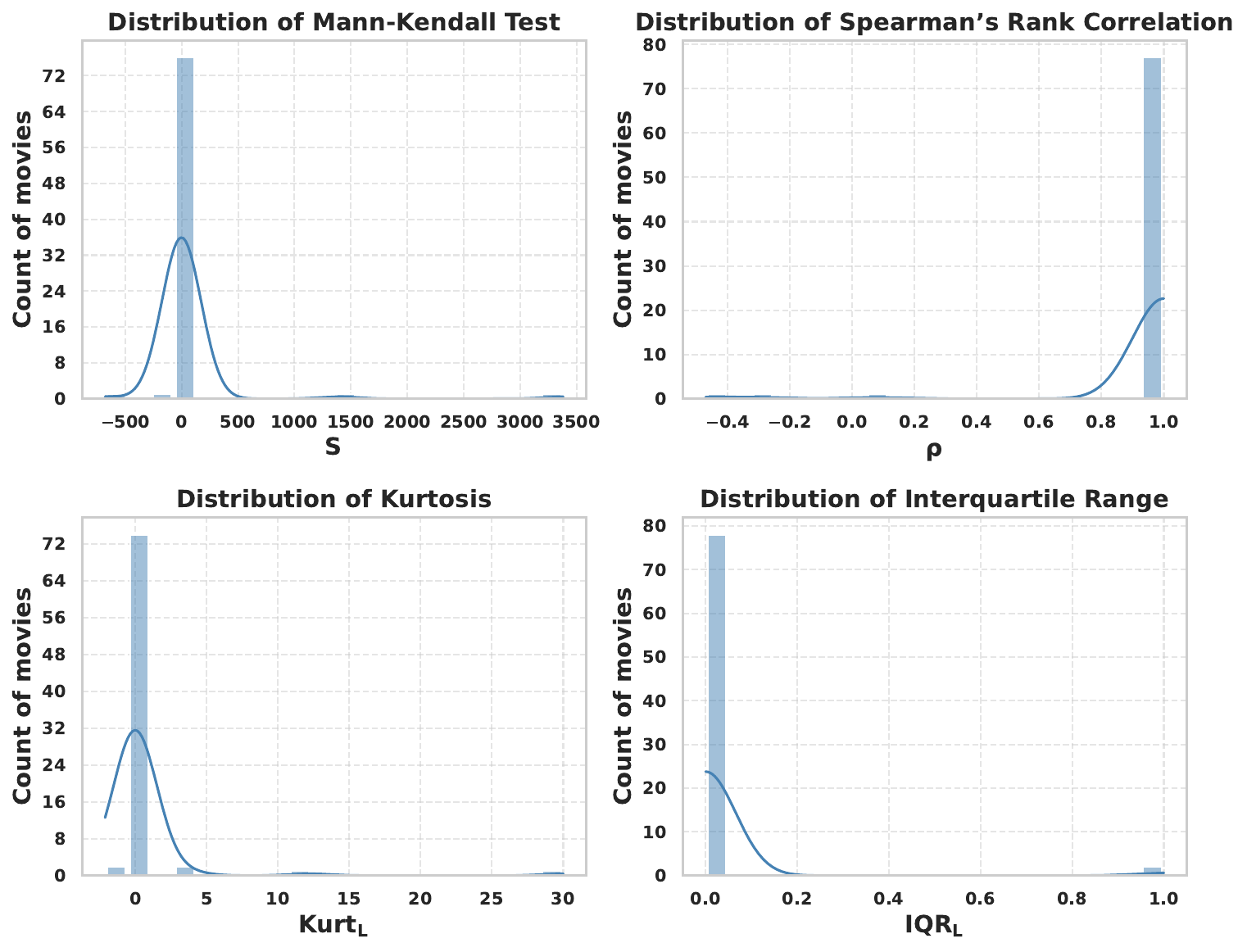}
    \caption{``w/o Persona \& w/o History''}
  \end{subfigure}
  \begin{subfigure}[b]{0.48\linewidth}
    \centering
    \includegraphics[width=\linewidth]{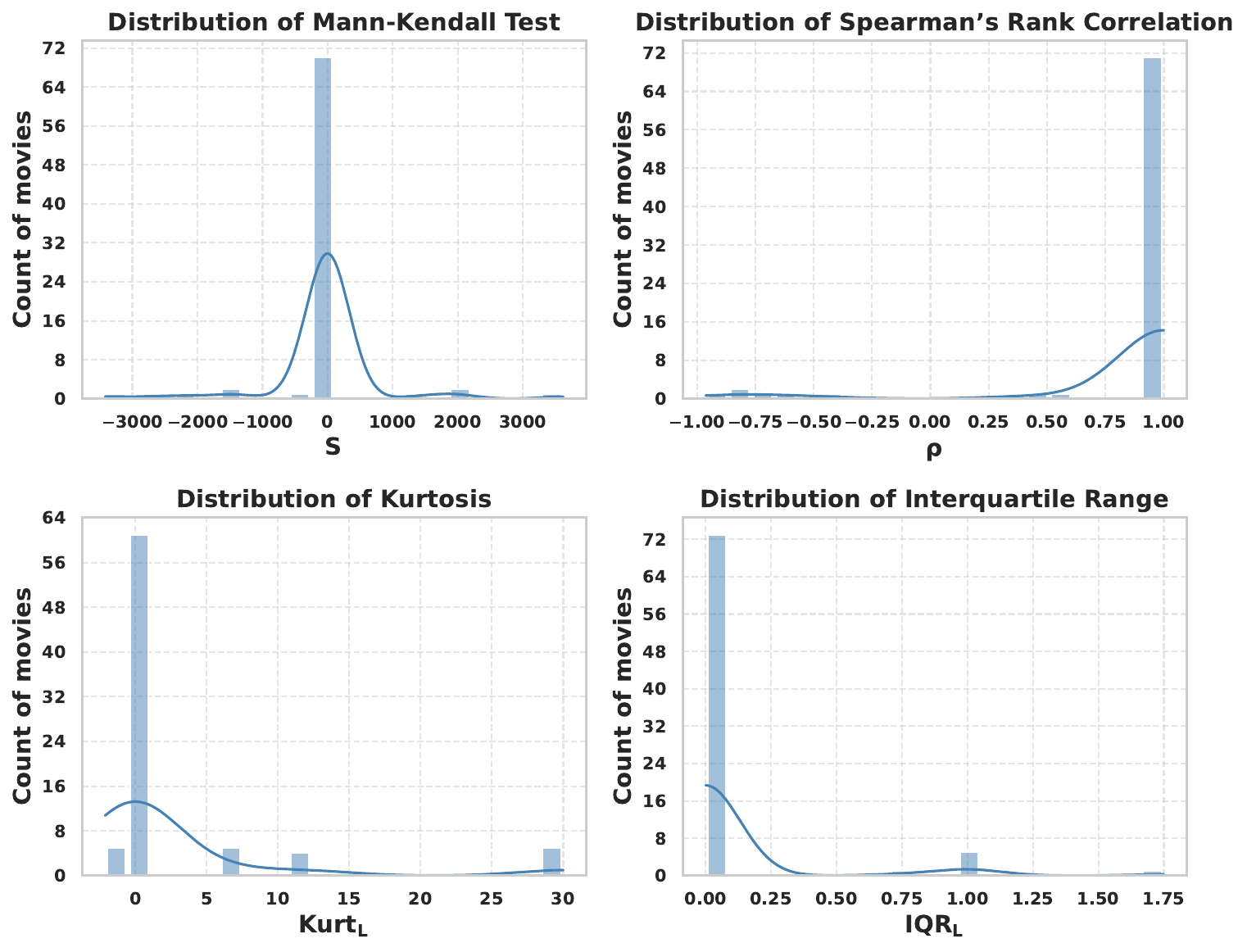}
    \caption{``w/o Persona \& w/ History''}
  \end{subfigure}\\
  \begin{subfigure}[b]{0.48\linewidth}
    \centering
   \includegraphics[width=\linewidth]{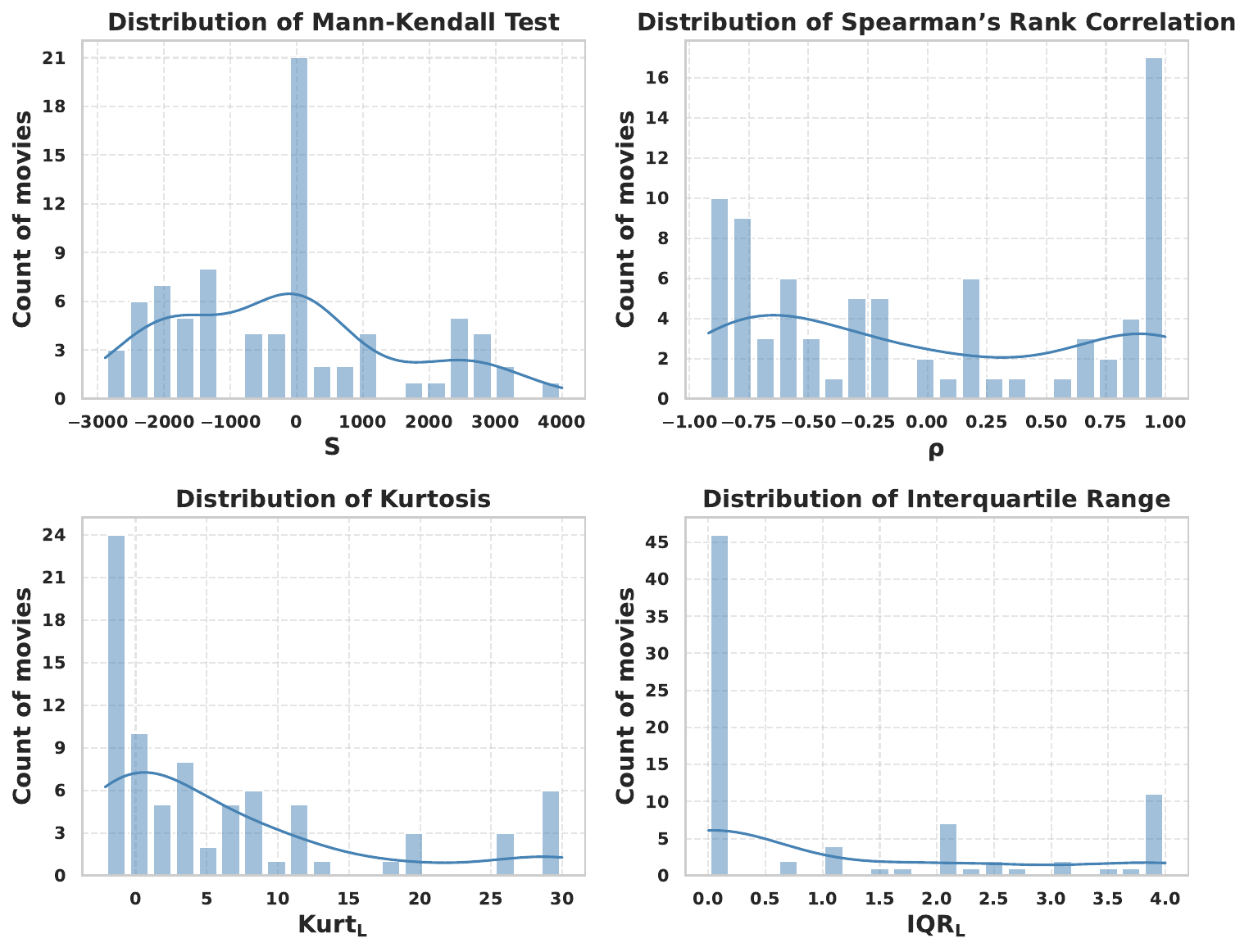}
    \caption{``w/ Persona \& w/o History''}
  \end{subfigure}
  \begin{subfigure}[b]{0.48\linewidth}
    \centering
    \includegraphics[width=\linewidth]{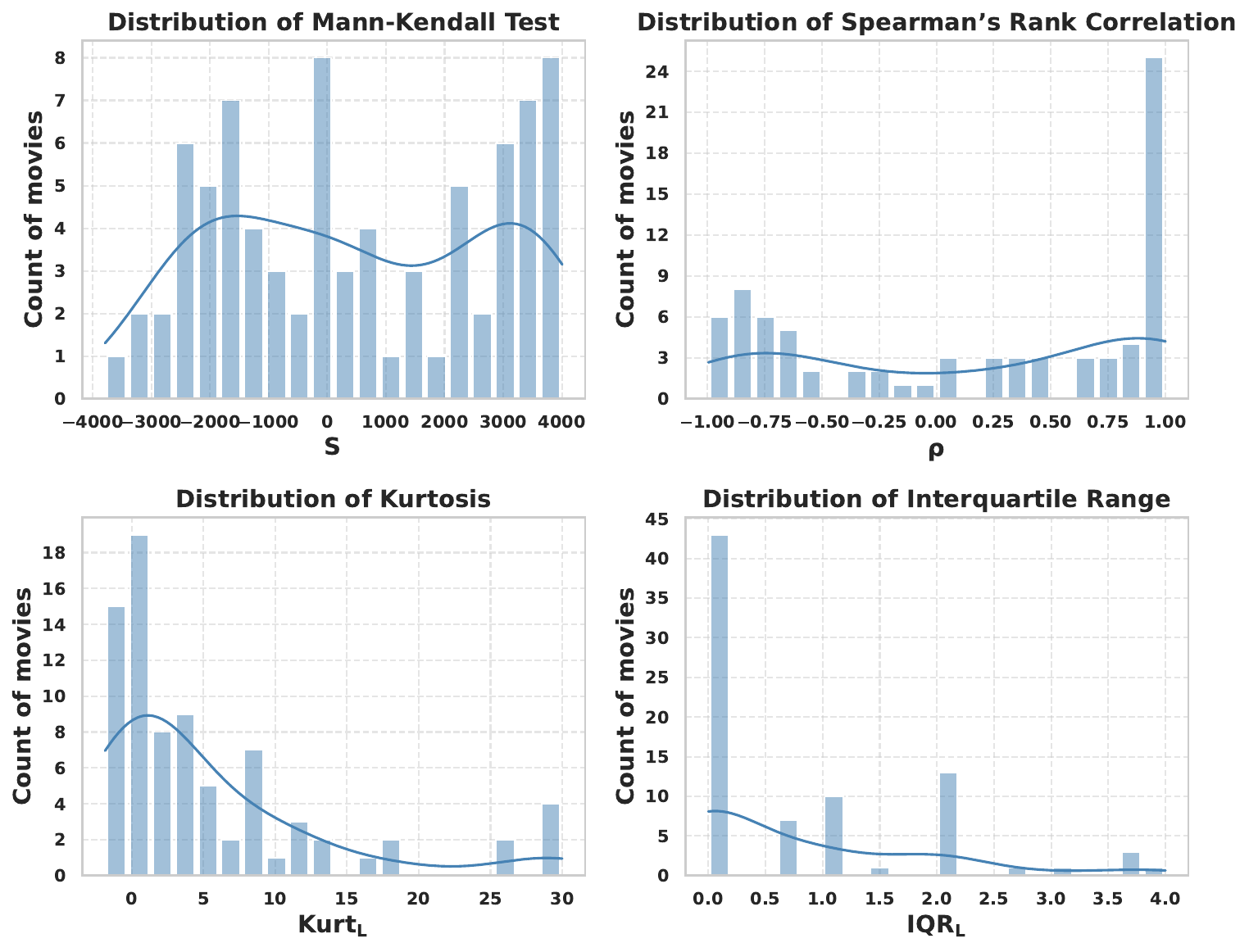}
    \caption{``w/ Persona \& w/ History''}
  \end{subfigure}
  \caption{Distributions of Mann–Kendall Statistic, Spearman Rank Correlation, Kurtosis, Inter-quartile Range for All Movie Rating Sequences on Qwen2.5-7B-Instruct.}
  \label{fig:Distributions of Mann–Kendall Statistic, Spearman Rank Correlation, Kurtosis, Inter-quartile Range for All Movie Rating Sequences on Qwen2.5-7B-Instruct}
\end{figure*}

\end{document}